\documentclass[twocolumn,a4paper]{svjour3} 

\smartqed  
\makeatletter
\expandafter\let\csname opt@amsmath.sty\endcsname\relax
\makeatother

\usepackage{mathtools}
\newcommand*\phantomrel[1]{\mathrel{\phantom{#1}}}

\usepackage[english]{babel}
\usepackage[autostyle, english = american]{csquotes}
\MakeOuterQuote{"}

\usepackage{graphicx}
\usepackage{flushend}
\usepackage{subcaption} 
\usepackage{algorithm,algpseudocode}
\usepackage{tabularx}
\usepackage{tabularray}
\usepackage{amsfonts} 
\usepackage[numbers]{natbib}

\usepackage{hyperref}
\hypersetup{
colorlinks=true,
linkcolor=blue,
citecolor=blue,
urlcolor=blue
}

\usepackage{esvect}
\usepackage{graphicx}
\usepackage{amsmath}

\usepackage{xcolor}

\usepackage{pdflscape} 
\usepackage{afterpage}
\usepackage{longtable}

\usepackage{hvfloat}
\usepackage{rotfloat}
\usepackage{varwidth}

\usepackage{multirow} 

 \journalname{Preprint submitted to Computer Methods and Programs in Biomedicine}
\begin{document}
\raggedbottom

\title{Real-time respiratory motion forecasting with online learning of recurrent neural networks for accurate targeting in externally guided radiotherapy}
\titlerunning{Real-time respiratory motion forecasting with online learning of RNNs for accurate targeting in radiotherapy}        

\author{Michel Pohl         \and
        Mitsuru Uesaka \and
        Hiroyuki Takahashi \and
        Kazuyuki Demachi \and
        Ritu Bhusal Chhatkuli
}

\institute{Michel Pohl \at
              The University of Tokyo, 113-8654 Tokyo, Japan\\
              \email{michel.pohl@centrale-marseille.fr}
           \and
           Mitsuru Uesaka \at
              Japan Atomic Energy Commission, 100-8914 Tokyo, Japan
           \and
           Hiroyuki Takahashi \and Kazuyuki Demachi \at
              The University of Tokyo, 113-8654 Tokyo, Japan
           \and
           Ritu Bhusal Chhatkuli \at
              National Institutes for Quantum and Radiological Science and Technology, 263-8555 Chiba, Japan              
}

\date{ }

\maketitle

\begin{abstract}
\setlength{\parindent}{0pt}

\emph{Background and Objective}: In lung radiotherapy, infrared cameras can track reflective objects on the chest to estimate tumor motion due to breathing. However, treatment system latencies hinder radiation beam precision. Real-time recurrent learning (RTRL), the conventional online learning approach for training recurrent neural networks (RNNs), is a potential solution that can learn patterns within non-stationary respiratory data but has high complexity. This research assesses the capabilities of resource-efficient online algorithms for RNNs---unbiased online recurrent optimization (UORO), sparse one-step approximation (SnAp-1), and decoupled neural interfaces (DNI)---to forecast respiratory motion during radiotherapy accurately. 

\emph{Methods}: We use nine time series lasting from 73s to 320s, each containing the three-dimensional (3D) locations of three external markers on the chest of healthy subjects. We propose efficient implementations for SnAp-1 and DNI that compress the influence and immediate Jacobian matrices and accurately update the linear coefficients used in credit assignment estimation, respectively. Data was originally sampled at 10Hz; we resampled it at 3.33Hz and 30Hz to analyze the effect of the sampling rate on performance. We use UORO, SnAp-1, and DNI to forecast each marker's 3D position with horizons $h \leq 2.1\text{s}$ (the time interval in advance for which predictions are made) and compare them with RTRL, least mean squares, kernel support vector regression, and linear regression. 

\emph{Results}: RNNs trained online achieved similar or better accuracy than most previous works using larger training databases and deep learning, although we used only the first minute of each sequence to predict motion within that exact sequence. SnAp-1 had the lowest normalized root-mean-square errors (nRMSEs) averaged over the horizon values considered, equal to 0.335 and 0.157, at 3.33Hz and 10Hz, respectively. Similarly, UORO had the lowest nRMSE at 30Hz, equal to 0.086. Linear regression was effective at low horizons, attaining an nRMSE of 0.098 for $h = 100\text{ms}$ at 10Hz. DNI's inference time (6.8ms per time step at 30Hz, Intel Core i7-13700 CPU) was the lowest among the RNN methods; it was 5 times lower than that of RTRL. 

\emph{Conclusions}: UORO, SnAp-1, and DNI can accurately forecast respiratory movements using little data, which will help improve radiotherapy safety.

\end{abstract}

\section{Introduction}

\subsection{Background on respiratory motion management} 
\label{subs:external markers}  

Machine learning applications to radiotherapy take various forms, including motion compensation during treatment \cite{huynh2020artificial}. Such compensation is needed because healthy tissue adjacent to the tumor, unfortunately, also receives irradiation due to inherent organ displacements during beam delivery. The main component of these displacements is breathing, but they are also partly comprised of other modes of deformation caused by cardiac or digestive activity that add noise to recorded chest trajectories. Chest tumor motion is primarily cyclic and has an extent in the superior-inferior (SI) direction that can reach beyond 5cm \cite{sarudis2017systematic}. Nonetheless, it is affected by phase shifts and fluctuations in amplitudes and frequencies \cite{verma2010survey, ehrhardt20134d}. Amplitude shifts designate steep and intermittent variations of the average tumor location, while the term "drift" encompasses more steady changes occurring within a single treatment session. Baseline intrafractional drifts of 1.65 $\pm$ 5.95mm, 1.50 $\pm$ 2.54mm, and 0.45 $\pm$ 2.23mm (mean $\pm$ standard deviation) in the SI, anterior-posterior, and left-right axes, respectively, have been highlighted in \cite{takao2016intrafractional}. Overall body movements associated with subject relaxation over time or subtle positional adjustments on the treatment couch also contribute to respiratory record variability. In addition, sudden changes or irregular patterns may result from yawning, hiccupping, sneezing, or coughing. One common approach to address these challenges involves recording the positions of external markers on the subject's abdomen and chest using infrared cameras. Subsequently, a mathematical correspondence model can be used to link the locations of these objects with that of the tumor \cite{ehrhardt20134d, mcclelland2013respiratory}. Systems like CyberKnife (Accuray) or Vero (BrainLab) utilize low-frequency kV imaging to update that correlation model in real time. In this context, AI techniques can help provide accurate estimates of the tumor position from the surrogate signals \cite{chen2018internal} and forecast the latter to compensate for the delay between target localization and treatment system response.

\subsection{Respiratory motion forecasting with artificial neural networks}
\label{section:intro pred in radiotherapy}

Radiotherapy treatment machines are affected by latencies intrinsic to data acquisition and processing, robotic control, and treatment beam delivery. Each system is characterized by its latency period: "for most radiation treatments, [it] will be more than 100ms, and can be up to two seconds" \cite{verma2010survey}. Not taking it into account can result in excessive damage to healthy tissue, which leads, in turn, to unwanted side effects such as radiation pneumonitis or pulmonary fibrosis. This is especially true in the cases of stereotactic radiosurgery and stereotactic body radiotherapy, where a high dose is delivered to the tumor in a few fractions, and narrow margins are required to spare normal tissue. Interstitial lung disease patients are particularly affected by this issue, as they are often deemed inoperable by anatomical surgical resection and are, therefore, usually treated with stereotactic ablative radiotherapy. Yet, they are at a higher risk of radiation-induced pulmonary toxicity \cite{goodman2020primer}. 

Various methods based on classical machine learning have been proposed to solve this problem \cite{verma2010survey, lee2014prediction, ehrhardt20134d}. Among these, artificial neural networks (ANNs) have generally been found effective at forecasting non-stationary and complex signals with a high horizon, also called response time or look-ahead time, which is the time interval in advance for which the prediction is made. The first studies about time-series forecasting in radiotherapy mainly involved ANNs with one hidden layer only, but deeper architectures are more common in recent works. The availability of larger datasets is one of the drivers of this transition, as deep learning networks often continue to improve as the dataset size increases. For instance, \citeauthor{lin2019towards} reported training long short-term memory (LSTM) networks using data comprising 1703 respiratory traces from 985 patients acquired at three clinical institutions \cite{lin2019towards}. 

Most previous studies used grid search to tune hyperparameters, such as the signal history length (SHL) or regularization strength. The latter are common to all algorithms; other hyperparameters specific to neural networks include the learning rate, the number of layers (in the case of deep ANNs), and the number of units per layer. It has been reported that an extensive search may not be clinically feasible due to high computational costs \cite{krauss2011comparative}. To address that challenge, \citeauthor{samadi2023respiratory} proposed a nonsequential-correlated hyperparameter optimization algorithm for deep recurrent neural networks (RNNs) to reduce the hyperparameter combinations that they explored from 700 million to just 30,000 \cite{samadi2023respiratory}. It was generally found that performance decreased as the horizon increased. Some studies addressed the robustness of respiratory prediction to unsteady patterns and breathing speed variations \cite{sun2017respiratory, sun2020adaptive, jeong2022clinical, liang2023real}. For instance, \citeauthor{jeong2022clinical} clustered irregular signals into three groups (irregular amplitude, irregular frequency, and both cases) based on a numerical variability metric and observed that those for which irregular amplitude patterns prevailed corresponded to a higher accuracy drop \cite{jeong2022clinical}. \citeauthor{liang2023real} experimentally observed that faster breathing also led to higher forecasting errors, as that scenario is equivalent to a lower signal sampling rate, $f$ \cite{liang2023real}. Indeed, it was observed that the root-mean-square error (RMSE) associated with a multilayer perceptron (MLP) with a single hidden layer (we refer to that structure as a one-layer MLP) predicting the position of an implanted marker increased from 2.5mm to 4.9mm and from 4.3mm to 6.0mm at $h=200\text{ms}$ and $h=1.0\text{s}$, respectively, when $f$ decreased from 30Hz to 1.0Hz \cite{sharp2004prediction}. 

Most previous works about respiratory motion forecasting focused on predicting one-dimensional (1D) respiratory signals. However, considering the correlation between time series corresponding to different moving points and directions will likely improve the accuracy of tumor position estimation. A straightforward approach consists of concatenating these components into a single vector fed into the network \cite{POHL2021101941, pohl2022prediction}; some studies employ a specialized module to capture inter-dimensional information, such as external attention \cite{zhang2023lgeanet}. It was reported in \cite{krauss2011comparative} that using principal components from successive 3D tumor centroid positions as input led to a higher forecasting accuracy than performing coordinate-wise prediction when $h \geq 0.4\text{s}$ with several classical machine learning algorithms.

Some works focused on the combined use of surrogate signal prediction and correspondence models \cite{wang2021real, chang2021real}. For instance, \citeauthor{wang2021real} compared support vector regression (SVR) and LSTMs to predict liver motion obtained with four-dimensional (4D) ultrasound imaging from light-emitting diodes (LEDs) fixed on the chest of volunteers (AccuTrack 250 system) and observed that LSTMs were more efficient both at correlating internal and external motion and forecasting markers on the chest surface \cite{wang2021real}. They also reported that continuously updating the correlation model enhanced accuracy. Similarly, \citeauthor{chang2021real} used temporal convolutional networks (TCNs) with residual connections to predict the positions of internal fiducial markers recorded via orthogonal X-ray imaging from luminous diodes on the abdomen and chest of cancer patients (CyberKnife Synchrony system). They found that using three external markers instead of one or two led to better overall forecasting performance \cite{chang2021real}.

Some recent studies apply time-series forecasting to surrogates from magnetic resonance (MR) images, as recent advances in MR-guided linear accelerator (LINAC) systems made it technically possible to visualize and track tumors in two-dimensional (2D) planes at frequencies of approximately 5Hz during treatment. For instance, \citeauthor{li2023online} compared the performance of linear regression and recurrent models forecasting the centroid position of lung tumors and the liver imaged with the MR scanner of the Unity system \cite{li2023online}. A recurrent model can also serve as a module in a larger architecture performing chest image prediction for MR-guided radiotherapy. For example, \citeauthor{romaguera2021probabilistic} integrated a sequence-to-sequence-inspired convolutional LSTM (convLSTM) model within an architecture performing 3D reconstruction from 2D navigator MR slices based on a convolutional variational autoencoder (cVAE) to forecast temporal image feature representations \cite{romaguera2021probabilistic}. They observed that the end-of-inhale phase was the hardest to predict, as it is subject to high variability among the different cycles.

Some more applied works focus on productizing forecasting algorithms within robotic treatment systems and evaluating their impact on dose delivery accuracy. For instance, \citeauthor{lee2021geometric} experimentally observed that LSTMs led to a higher gamma passing rate (the percentage of points for which the gamma index is lower than 1, indicating high local correlation between calculated and measured dose) than exponential smoothing or the absence of forecasting, under the 2\%/2mm and 3\%/3mm tolerance criteria \cite{lee2021geometric}. 

Advances in respiratory motion forecasting will also impact motion management in other areas of medicine. Indeed, methods based on ANNs have recently been proposed to predict the positions of arteries in X-ray angiographic imaging and help with navigation guidance in cardiac interventions \cite{azizmohammadi2023patient}, estimate future target trajectories in ultrasound image sequences to improve automated puncture systems in ablation surgery \cite{yao2022feature}, and forecast the locations of vertebrae to enhance the accuracy of pedicle screw placement in spinal surgery \cite{hanrespiratory}.

\subsection{RNNs and transformers for breathing motion prediction}
\label{section:intro RNN and transformers}

Recurrent connections within network architectures are prevalent in the recent research literature regarding respiratory motion forecasting for radiotherapy. Indeed, the feedback loop characterizing various types of RNNs behaves as a memory and allows information retention as time elapses. As a result, these networks can learn patterns and dependencies within sequential and time-series data efficiently. Some recent works demonstrated the potential of deep recurrent architectures based on LSTMs, bi-LSTMs, and bi-gated recurrent unit (bi-GRU) layers for respiratory motion prediction \cite{lin2019towards, wang2018feasibility, yu2020rapid, samadi2023respiratory}. It was reported, for instance, that bi-LSTMs had better performance than the adaptive-boosting MLP model \cite{wang2018feasibility}.

The recent development of attention-based architectures, including the transformer, also impacted research on respiratory motion prediction. Attention mechanisms were first introduced for natural language processing tasks; they calculate soft word embedding weights that can change during runtime. They address RNN weaknesses, such as slow processing and the fading of words appearing early in a text, by leveraging parallelism and providing all tokens equal access to any sentence part, respectively. When applied to time-series prediction, they help networks focus on time intervals that significantly impact accuracy by increasing corresponding weights. Despite initial works providing evidence that attention-based architectures can be more efficient than RNNs at respiratory motion forecasting \cite{yao2022feature, jeong2022clinical, romaguera2023conditional, shi2022respiratory} and the general high performance of transformers at many tasks due to parallel processing and the absence of a vanishing gradient, transformers "are impractical for training or inference in resource-constrained environments due to their computational and memory requirements" \cite{subramoney2023efficient}. Indeed, their complexity quadratically grows with the input window length, which hinders their ability to learn long-range dependencies \cite{li2019enhancing, dao2022flashattention}. For instance, it was observed in \cite{romaguera2023conditional} that transformers predicting breathing signal representations from chest cine-MR imaging led to an inference time approximately three times higher than convolutional GRUs. Furthermore, recent works integrating recurrent and attention-based modules in the same architecture demonstrated high performance in respiratory motion prediction \cite{tan2022lstformer, zhang2023lgeanet}. These findings suggest that RNN-based approaches are still relevant in this field.

\subsection{Irregular motion mitigation via parameter adaptation}
\label{section:intro parameter adaptation}

Regardless of the chosen architecture, adapting the prediction model as new training examples arrive can help cope with irregular breathing characteristics that may not have yet appeared in the training set. That can help mitigate the complexity of acquiring large datasets in the medical space (see, for instance, the following related works tackling data acquisition constraints in medical imaging, exploring supervised segmentation with scarce data and unsupervised domain adaptation: \cite{hong2022unsupervised, hong2022source, su2023attention, li2024source}). A simple strategy in time-series forecasting consists of retraining the model as new samples arrive using a sliding window, beyond which data is not used for training. Such an approach was proposed for classical machine learning algorithms (linear regression, kernel density estimation, and SVR) and one-layer MLPs to predict tumor centroid positions estimated from marker surrogates \cite{krauss2011comparative, teo2018feasibility}. A corresponding RMSE decrease of approximately 5\% when using an adaptive retraining scheme was reported in \cite{krauss2011comparative}. \citeauthor{yu2020rapid} were the first to apply such a sliding window approach to recurrent models, as they predicted 1D principal component analysis (PCA) respiratory traces from AccuTrack 250 external marker data with continually retrained bi-GRUs \cite{yu2020rapid}. In that study, the network weights were updated when the prediction error exceeded an arbitrary value. Later, it was shown that dynamically retrained LSTMs performed significantly better than LSTMs trained offline and adaptive linear filters when forecasting the tumor centroid SI position in cine-MR images at horizon values $h \geq 500\text{ms}$ \cite{lombardo2022offline}. In the latter work, the relatively low sampling rate of 4Hz allowed retraining the LSTM for 10 epochs at each time step. Although sliding window adaptation can improve performance, it has several downsides. First, it introduces more hyperparameters, such as the number of epochs and length of the window containing the data for dynamic retraining (e.g., an increasing window length is proposed in \cite{krauss2011comparative}), and necessitates arbitrary choices, such as the criterion to stop the retraining process and a heuristic determining when parameter update is appropriate (e.g., every $k$ time steps with $k$ to select or/and when the prediction error is too high). Second, when adapting to a new window, the algorithm gradually "forgets" the previously learned data characteristics beyond that window with successive training epochs. This phenomenon is analogous to catastrophic forgetting in the continual learning setting.

Concerning online learning with classical machine learning algorithms, SVRpred was used to adaptively predict simulated and real (CyberKnife) respiratory data without fitting the SVR model from scratch at regularly spaced intervals \cite{ernst2009forecasting}. In SVRpred, the support vector set and kernel matrix are incrementally updated in an efficient manner, which helps avoid solving the entire quadratic programming problem and recomputing kernel values repetitively, thereby reducing the computational complexity compared to full retraining \cite{ma2003accurate}. \citeauthor{ma2003accurate} found that SVRpred was more effective than its static SVR counterpart, which undergoes no updates after the initial training, for various time-series benchmarks. Regarding respiratory signal forecasting, \citeauthor{ernst2009forecasting} experimentally observed that SVRpred was more accurate than multi-step linear methods (MULIN) and wavelet-based multiscale autoregression (wLMS) at the inhalation peaks \cite{ernst2009forecasting}. Similarly, an architecture combining feature extraction with random convolution nodes (RCNs) governed by local receptive fields (LRFs) and extreme learning machines (ELMs), trained with an efficient online update rule, referred to as "online sequential forecasting RCN" (OS-fRCN) was proposed in \cite{wang2020fast}. Experiments with PCA-processed traces from 304 motion records revealed that OS-fRCN led to lower prediction errors than other ELM-based methods and a relevance vector machine (RVM) model at various horizons, except at $h=76\text{ms}$, where the RVM was more accurate. Additionally, OS-fRCN was compared to a deep LSTM and a deep CNN; while their accuracy was similar at low horizons, that of OS-fRCN was relatively higher for higher values of $h$.

\subsection{Online learning of recurrent neural networks}

In contrast to adaptive retraining with a sliding window, truly online algorithms for ANNs do not discard past information as the associated network update equations do not explicitly reference past activity, which prevents forgetting distant dependencies. Real-time recurrent learning (RTRL), the backbone of many developments in the field of online learning algorithms for RNNs, is based on the recursive exact update of the influence matrix (the total derivative of the hidden state with respect to the parameters), also called sensitivity matrix, which characterizes the network behavior, at every time step \cite{williams1989learning}. That algorithm was found relatively effective in the context of radiotherapy for predicting the positions of spherical markers implanted in the lung (SyncTraX system) \cite{jiang2019prediction}, chest and abdominal tumors recorded from the CyberKnife Synchrony system \cite{mafi2020real}, chest internal points tracked using deformable registration in 4D computed tomography (CT) and 4D cone-beam CT (4D-CBCT) images \cite{POHL2021101941}, and external markers on the chest and abdomen of healthy subjects (NDI Polaris) \cite{pohl2022prediction}. The main drawback of RTRL is its high computational complexity of $\mathcal{O}(q^4)$, where $q$ is the number of neurons. That makes inference practically unfeasible for even relatively moderate values of $q$. 

Various resource-efficient online training algorithms have been developed to address the slow processing time of RTRL and estimate the loss gradient without bias in order to strike a balance between short-term and long-term temporal dependencies (Table \ref{table:RNN online learning comparison}). This is something which truncated backpropagation through time \cite{jaeger2002tutorial}, the more conventional sliding window retraining approach for RNNs, cannot achieve. \citeauthor{marschall2020unified} compared several of those alternative algorithms and proposed a unified framework based on tensor structure and a distinction between past-facing and future-facing algorithms \cite{marschall2020unified}. The latter refers to whether the sum of past or future instantaneous losses is minimized. Past-facing algorithms try to compress the influence matrix. In contrast, future-facing algorithms must predict the credit assignment vector (also called error signal), which is the derivative of the total loss with respect to the hidden states. 

\begin{table}[htb!]
\setlength{\tabcolsep}{4pt}
\begin{center}
\begin{tabular}{lll}
\hline
Algorithm                                                   & \multicolumn{2}{c}{Complexity}  \\
                                                            & Memory & Time \\
\hline
Real-time recurrent learning \cite{williams1989learning}    & $\mathcal{O}(q^3)$  & $\mathcal{O}(q^4)$      \\
Truncated BPTT \cite{williams1990efficient}                 & $\mathcal{O}(T q)$  & $\mathcal{O}(T q^2)$    \\
Unbiased online recurrent                                   & $\mathcal{O}(q^2)$  & $\mathcal{O}(q^2)$      \\
optimization \cite{tallec2018unbiased} & & \\
Kronecker-factored RTRL \cite{mujika2018approximating}      & $\mathcal{O}(q^2)$  & $\mathcal{O}(q^3)$      \\
Kernel RNN learning \cite{roth2018kernel}                   & $\mathcal{O}(q^2)$  & $\mathcal{O}(q^2)$      \\
r-optimal Kronecker-sum                                     & $\mathcal{O}(rq^2)$ & $\mathcal{O}(rq^3)$     \\
approximation \cite{benzing2019optimal} & & \\
Random-feedback online learning \cite{murray2019local}      & $\mathcal{O}(q^2)$  & $\mathcal{O}(q^2)$      \\
Sparse one-step approximation \cite{menick2020practical}      & $\mathcal{O}(q^2)$  & $\mathcal{O}(q^2)$      \\
Reverse Kronecker-factored RTRL \cite{marschall2020unified} & $\mathcal{O}(q^2)$  & $\mathcal{O}(q^3)$      \\
Efficient BPTT \cite{marschall2020unified}                  & $\mathcal{O}(T q)$  & $\mathcal{O}(q^2)$      \\
Future-facing BPTT \cite{marschall2020unified}              & $\mathcal{O}(T q)$  & $\mathcal{O}(T q^2)$    \\
Decoupled neural interfaces \cite{jaderberg2017decoupled}   & $\mathcal{O}(q^2)$  & $\mathcal{O}(q^2)$      \\
\hline
\end{tabular}
\caption{Memory and time complexity of several online learning algorithms for RNNs. In the last two columns, $q$ and $T$ designate the number of hidden units of the RNN and the truncation length, respectively\protect\footnotemark.}
\label{table:RNN online learning comparison}
\end{center}
\end{table}

\footnotetext{Adapted from \citep{marschall2020unified} (open-access article), Copyright JMLR 2020.}

Unbiased online recurrent optimization (UORO) is a past-facing algorithm that attempts to estimate the influence matrix as the product of two random vectors recursively updated at each time step, based on the "rank-one trick" \cite{tallec2018unbiased}. This technique helps reduce the overall algorithm complexity to $\mathcal{O}(q^2)$ while maintaining a closed-form update at the expense of introducing stochasticity. Among online algorithms for RNNs, RTRL and UORO have strong theoretical backing regarding local convergence \cite{masse2020convergence}. It has been observed that UORO is practically more accurate than RTRL while maintaining an acceptable inference time when predicting the motion of external markers on the chest of breathing subjects \cite{pohl2022prediction}. The latter study also provided closed-form expressions for quantities appearing in the calculation of the loss gradient of vanilla RNNs to help implement UORO efficiently for that particular architecture. Among all the algorithms for online training of RNNs examined in \cite{marschall2020unified}, the lowest time complexity achieved was $\mathcal{O}(q^2)$ (Table \ref{table:RNN online learning comparison}). This is also the case of decoupled neural interfaces (DNI), a future-facing algorithm that relies on linear prediction of the credit assignment vector from the past state and the latest incoming data sample based on a "bootstrapping" technique. DNI was initially introduced as a broad framework also applicable to non-recurrent networks. It seeks to break the constraints of modules needing to wait for others to finish forward or backward computation before their own update \cite{jaderberg2017decoupled}. This is accomplished through learning a "synthetic gradient," a separate prediction of the loss gradient for every network layer. In contrast to UORO, DNI's updates are biased, deterministic, and numerical, as there is no straightforward formula to calculate the linear regression coefficients, and a gradient descent step is performed instead.

Some of the most recent approaches in online learning of RNNs involve small independent recurrent modules, where each module state does not affect the dynamics of others and for which exact RTRL is computationally cheap \cite{zucchet2023online, javed2023scalable}. \citeauthor{silver2021learning} remarked, "the directional derivative of a recurrent function along any arbitrary direction \emph{u} can be computed efficiently and then can be used to construct a descent direction" \cite{silver2021learning}. Following that observation, they proposed deep online directional gradient estimate (DODGE), whose particular case with multiple random directions generalizes RTRL. Another research direction consists of the improvement of RTRL performance through sparsity. \citeauthor{subramoney2023efficient} introduced combined activity and parameter sparsity for event-based GRUs (EGRUs) \cite{subramoney2023efficient}, whereas sparse-n step approximation (SnAp-n), proposed by \citeauthor{menick2020practical}, integrates parameter sparsity and influence matrix approximations \cite{menick2020practical}. In SnAp-n, only the influence of parameters on neurons affected by them within $n$ steps of the recurrent core are tracked; the update is biased but has a non-stochastic closed form. The case $n=1$ (SnAp-1) corresponds to a diagonal approximation of the influence matrix, applicable to any recurrent architecture, similar to the diagonal approximation of RTRL used in the original LSTM article \cite{hochreiter1997long}.

\subsection{Content of this study}

\begin{figure*}[htb!]
	\centering
	\includegraphics[width=\linewidth]{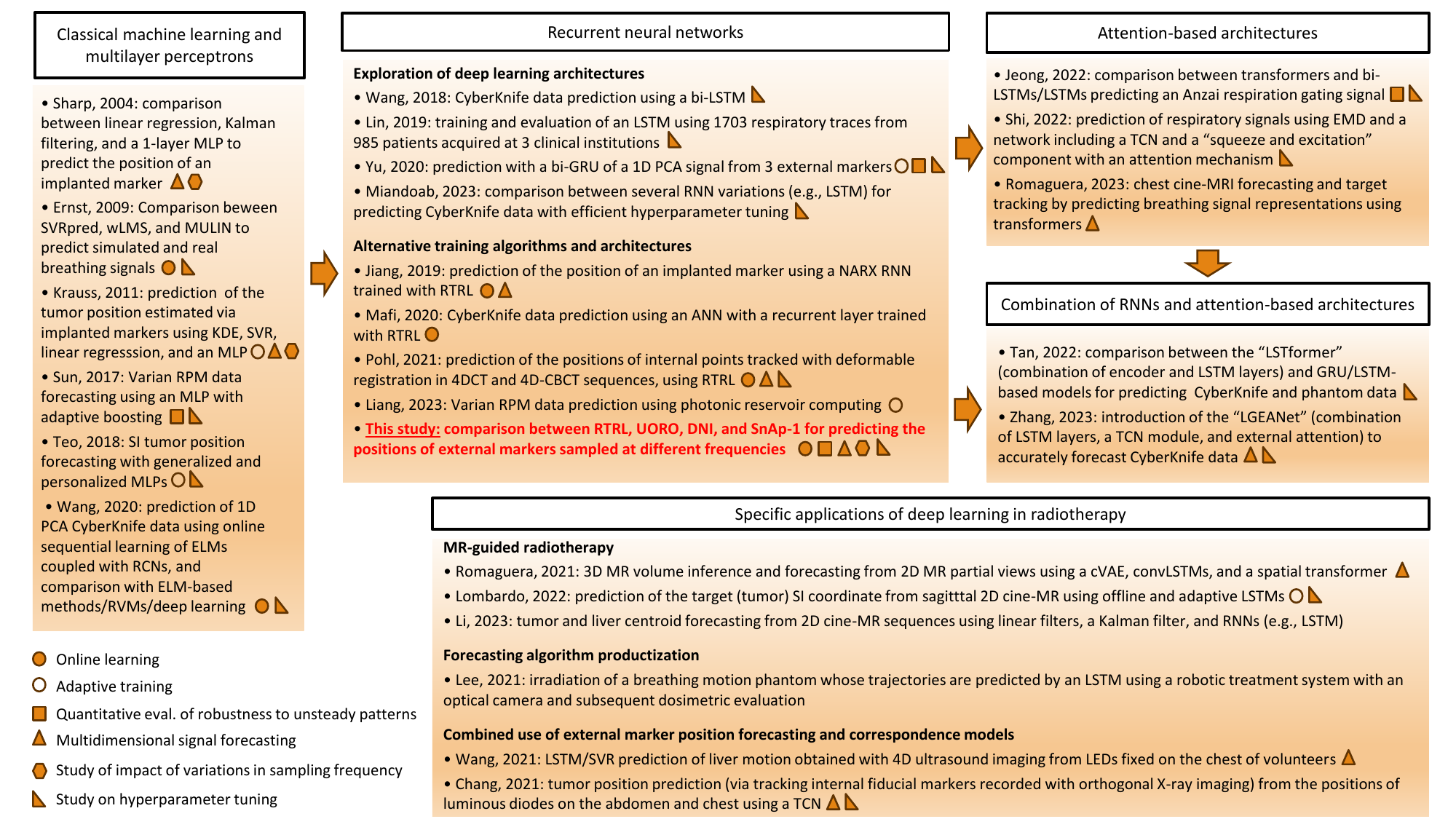}
	\caption{Roadmap illustrating the significance of this study within the broader context of respiratory motion forecasting for radiotherapy\protect\footnotemark. }
	\label{fig:time scope}
\end{figure*}

\footnotetext{In the figure, the acronyms "KDE," "RPM," and "NARX" respectively stand for "kernel density estimation," "real-time position management," and "nonlinear autoregressive model with exogenous inputs." This figure does not exhaustively represent the whole field; we selected studies arbitrarily based on their perceived diversity, importance, and relatedness to this article. It is worth mentioning that some studies could belong to multiple categories among those outlined (e.g., \cite{romaguera2023conditional} could also be classified as within the "MR-guided radiotherapy" group).}

Our research investigates the feasibility of forecasting breathing motion with fast online learning algorithms for RNNs. This is the first work analyzing the potential of RNNs trained with DNI and SnAp-1 to accurately predict the displacements of external markers on the chest and abdomen for safer externally guided radiotherapy (Fig. \ref{fig:time scope}). These two learning algorithms have high clinical potential as they can leverage the RNN memory structure, suitable for sequence processing, and bring adaptation capabilities without forgetting data while benefiting from low computational requirements (Table \ref{table:RNN online learning comparison}). Prior studies on respiratory motion forecasting tend to explore ANN architectures and propose generalized models. Instead, we focus on the training algorithm itself and build a patient-specific model as a complementary approach. We propose efficient implementations of DNI and SnAp-1 for vanilla RNNs, based respectively on a "compression" of the influence and Jacobian matrices into non-sparse matrices, lowering memory requirements, and an improved formulation of the updates of the linear coefficients involved in credit assignment estimation, with regards to that in \cite{marschall2020unified}. We compare these two methods with RTRL, UORO, least mean squares (LMS), SVR with a radial basis function (RBF) kernel, and linear regression for an extensive range of response time values $h$, spanning from $h_{min} = 0.1\text{s}$ to $h_{max} = 2.1\text{s}$, and sampling frequencies, $f$, from 3.33Hz to 30Hz. Investigating performance variation with $f$ addresses a knowledge gap as prior studies on the influence of $f$ are scarce, yet this can provide valuable insights into how the forecasting behavior of different prediction methods varies across diverse clinical systems. Notably, low frequencies are typical of radiotherapy guided by magnetic resonance imaging (MRI). In addition, to the best of our knowledge, this work is the first to quantify both the effects of $f$ and $h$ on RNN hyperparameter optimization in the context of respiratory motion forecasting. Unlike most prior studies that tackle univariate signal prediction, we perform three-dimensional (3D) breathing motion forecasting and leverage correlations between respiratory traces corresponding to each direction or signal, as this is likely to enhance accuracy and robustness to unsteadiness and noise. Moreover, this setting is more relevant clinically, as tumor motion is also three-dimensional. We analyzed the robustness of each algorithm to non-stationary patterns by splitting the records into two groups, namely regular and irregular breathing, and comparing the performance obtained with each group. Furthermore, we assessed how hyperparameter selection affected the accuracy of UORO, DNI, and SnAp-1 while considering variations in horizons and frequencies. We report the highest number of characterization metrics (mean average error [MAE], RMSE, normalized RMSE [nRMSE], maximum error, and jitter) among the previous works about breathing motion forecasting; this helps better describe the behavior of different algorithms.

\section{Material and Methods}

\subsection{Marker position data}

In our work, we consider nine time series, each corresponding to the 3D trajectories of three external markers placed on the abdomen and chest of three subjects (healthy males aged 20 to 40 years) breathing in a supine position. Markers 1, 2, and 3 were respectively located on the lower abdomen center, upper abdomen center, and upper chest center, except in sequences 6 and 7. In these two sequences, markers 2 and 3 were instead placed on the lower-right and upper-right sides of the abdomen, respectively. The respiratory traces were acquired via an infrared stereo camera (NDI Polaris). The raw time-dependent positions from the acquisition system (Rubedo Systems) were not equally spaced in time. Therefore, \citeauthor{krilavicius2016predicting} resampled these time series to 10Hz \cite{krilavicius2016predicting}; it is this resampled data that is used in our study. The motion extent in the craniocaudal, left-right, and dorsoventral directions is between 6mm and 40mm, 2mm and 10mm, and 18mm and 45mm, respectively. Each sequence lasts between 73s and 320s. Five traces are associated with regular breathing, while the remaining four were recorded as individuals were instructed to engage in various activities. Specifically, sequences 1 and 4, corresponding respectively to talking and "laughing and talking," are characterized by high fluctuations in amplitude. Such strong irregularities also appear in sequence 9, although the latter was labeled as "normal breathing" in \cite{krilavicius2016predicting}. Sequence 7, classified as "other" in the latter article, corresponds to slow and high-amplitude breathing motion. It is the shortest time series within the entire dataset and only features three full respiratory cycles. The breathing motion in sequence 3 was categorized as "normal and other" in \cite{krilavicius2016predicting}. Finally, one can observe a pronounced general drift of the positions of the markers throughout record 8. Further details about the dataset are available in \cite{krilavicius2016predicting}. In our study, we not only use the original data sampled at 10Hz, but we also downsample it to 3.33Hz by selecting one data point every three time steps and upsample it to 30Hz using cubic spline interpolation (Fig. \ref{fig:resampling process} in Appendix \ref{appendix:resampling process}). After the upsampling step, we add random additive noise following a normal distribution to the data points not originally in the 10Hz sequence to simulate noise related to sensor limitations and local respiratory motion unsteadiness\footnotemark. Finally, we set the precision of the upsampled signal to one decimal place, as in the original 10Hz signal, using truncation. 

\footnotetext{The noise standard deviation is chosen as $\sigma_{i,j} = \gamma \text{ max}_{k, l}|v_i^j(t_k) - v_i^j(t_l)|$, where $v_i^j(t)$ is the (non-normalized) displacement of marker $i \in \{1, 2, 3\}$ along coordinate $j \in \{x, y, z\}$ at time $t$, and $\gamma$ is a proportionality constant arbitrarily set to $1/150$.}

\subsection{Online training algorithms for RNNs}

\subsubsection{General framework for standard RNNs}\label{section: intro - standard RNNs}

In this study, an RNN with a single hidden layer is trained to forecast in real time the positions of three external markers as they move on the chest of each subject during breathing. We use the same general RNN equations as in \cite{pohl2022prediction}, which we recall in this section. We denote by $u_n \in \mathbb{R}^{m+1}$, $x_n \in \mathbb{R}^q$, $y_{n+1} \in \mathbb{R}^p$, and $\theta_n$ the input, state, output, and synaptic weight vectors of the RNN at time $t_n$, respectively. The state equation characterizes the update of the RNN's internal states given a new input and the previous state vector:
\begin{equation} \label{eq:RNN state eq}
x_{n+1} = F_{\text{st}}(x_n, u_n, \theta_n)
\end{equation}
Similarly, the measurement equation describes how to compute the RNN output given the updated state vector (calculated via Eq. \ref{eq:RNN state eq}):
\begin{equation} \label{eq:RNN measurement eq}
y_{n+1} = F_{\text{out}}(x_n, u_n, \theta_n)
\end{equation}

In the online learning setting, incoming data arrives in a streaming fashion, with training examples, $(u_n, y_{n+1})$, coming one after another, and the RNN synaptic weights are updated with each newly available example. This is why we denote the parameter vector by $\theta_n$ and not $\theta$. As follows, the instantaneous square loss $L_{n+1}$ is defined as the square of the instantaneous error $e_{n+1}$ between the prediction $y_{n+1}$, computed from the input $u_n$, and the ground truth $y_{n+1}^{*}$:
\begin{equation} \label{eq:loss function}
 e_{n+1} = y_{n+1}^* - y_{n+1} ,
\qquad
 L_{n+1} = \frac{1}{2} \|e_{n+1}\|_2^2
\end{equation}

In this work, we use a vanilla RNN structure, a network whose updated state $x_{n+1}$ results from applying a non-linear activation function $\Phi$ to a linear combination of the current state $x_n$ and input $u_n$ (Eq. \ref{eq:state_vanilla}) and whose output $y_{n+1}$ linearly depends on the updated state\footnotemark (Eq. \ref{eq:measurement_vanilla}). The parameter vector $\theta_n$ is defined as the concatenation of the flattened coefficient matrices $W_{a,n}$, $W_{b,n}$, and $W_{c,n}$, of respective sizes $q \times q$, $q \times (m+1)$, and $p \times q$, appearing in those two equations.
\begin{equation} \label{eq:state_vanilla}
F_{\text{st}}(x_n, u_n, \theta_n) = \Phi(z_n) \text{ with } z_n=W_{a,n} x_n + W_{b,n} u_n
\end{equation}
\begin{equation} \label{eq:measurement_vanilla}
F_{\text{out}}(x_n, u_n, \theta_n) = W_{c,n} F_{\text{st}}(x_n, u_n, \theta_n)
\end{equation}

\footnotetext{In this work, we use the coordinate-wise hyperbolic tangent function as the hidden layer activation: 
\begin{equation}
\Phi (z_1, ..., z_q) 
= [\phi(z_1), ..., \phi(z_q)]
= [\text{tanh}(z_1), ..., \text{tanh}(z_q)]
\end{equation}\label{eq:non_linearity} 
for $(z_1, ..., z_q) \in \mathbb{R}^q$.
}

\subsubsection{Past-facing algorithms: RTRL, UORO, and SnAp-1}

The impacts of alterations of $\theta_n$ on the state vector $x_{n+1}$ and instantaneous loss $L_{n+1}$ are characterized respectively by Eqs. \ref{eq:influence update} and \ref{eq:parameters_gradient}. The latter can be derived using the chain rule applied to the state and measurement equations (Eqs. \ref{eq:RNN state eq} and \ref{eq:RNN measurement eq}).
\begin{equation} \label{eq:influence update}
\frac{\partial{x_{n+1}}}{\partial \theta} = 
\frac{\partial{F_{\text{st}}}}{\partial x}(x_n, u_n, \theta_n) 
\frac{\partial{x_n}}{\partial \theta} + 
\frac{\partial{F_{\text{st}}}}{\partial \theta}(x_n, u_n, \theta_n)
\end{equation}
\begin{multline} \label{eq:parameters_gradient}
\frac{\partial{L_{n+1}}}{\partial \theta} = 
\frac{\partial{L_{n+1}}}{\partial y}(y_{n+1}) 
\left[ \frac{\partial{F_{\text{out}}}}{\partial x}(x_n, u_n, \theta_n)
\frac{\partial{x_n}}{\partial \theta} \right. \\
\left. + \frac{\partial{F_{\text{out}}}}{\partial \theta}(x_n, u_n, \theta_n) \right]
\end{multline}

The RTRL algorithm involves calculating the gradient of $L_{n+1}$ with respect to $\theta_n$ via Eq. \ref{eq:parameters_gradient} and recursively updating the influence matrix $\partial{x_n}/ \partial \theta$ via Eq. \ref{eq:influence update}. RTRL is computationally demanding due to the size of the latter matrix, which grows cubically with $q$. UORO alleviates that burden by introducing an unbiased rank-one estimator to approximate the influence matrix. Specifically, two random column vectors, $\tilde{x}_n$ and $\tilde{\theta}_n$, undergo recursive updates so that the relationship $\mathbb{E}(\tilde{x}_n \tilde{\theta}_n^T) = \partial{x_n}/ \partial \theta$ is satisfied at each time step. Details concerning UORO in general and its implementation in this study are available in \cite{tallec2018unbiased} and \cite{pohl2022prediction}, respectively.

In SnAp-1, the dynamic matrix $D_n = {\partial{F_{\text{st}}}}/{\partial x}$ is approximated by a diagonal matrix $\overline{D_n}$ whose elements are exactly its diagonal elements. Consequently, entries in ${\partial{x_n}}/{\partial \theta}$, which we initialize to the null matrix, are kept only if those at the same place in the immediate Jacobian matrix ${\partial{F_{\text{st}}}}/{\partial \theta}$ are non-zero, as the influence matrix update equation becomes:
\begin{equation} \label{eq:influence update SnAp-1}
\frac{\partial{x_{n+1}}}{\partial \theta} = 
\overline{D_n} \,
\frac{\partial{x_n}}{\partial \theta} + 
\frac{\partial{F_{\text{st}}}}{\partial \theta}(x_n, u_n, \theta_n)
\end{equation}

In the case of vanilla (dense) RNNs, defined by Eqs. \ref{eq:state_vanilla} and \ref{eq:measurement_vanilla}, one can demonstrate that the immediate Jacobian has at most one non-zero element per column at the same location for all steps $n$:
\begin{equation}\small
\frac{\partial F_{\text{st}}}{\partial \theta}
= \Big[ x_{n,1} \text{Diag}(\Phi'(z_n)), ..., u_{n,m+1} \text{Diag}(\Phi'(z_n)), 0_{q \times pq} \Big] \label{eq: standard RNN immediate Jacobian short version}
\end{equation}
Therefore, as we initialize ${\partial{x_n}}/{\partial \theta}$ to the null matrix, one can prove by recursion that it has also at most one non-zero element per column at the same location. In other words, when approximating $D_n$ by a diagonal matrix (SnAp-1 assumption) and using standard RNNs, the formula describing the recursive update of the influence matrix (Eq. \ref{eq:influence update SnAp-1}) involves only sparse matrices. Hence, performing multiplications using that formulation lacks efficiency. To mitigate that limitation and improve time performance, in this work, we introduce the compact immediate Jacobian 
\begin{equation}\label{eq:compact immediate Jacobian}
I_n = \Phi'(z_n) [x_n^T, u_n^T]
\end{equation}
and rewrite Eq. \ref{eq:influence update SnAp-1} as follows:
\begin{equation}\label{eq: reduced influence update}
J_{n+1} = \overline{D_n} J_n + I_n
\end{equation} 
In the latter equation, $J_n \in \mathbb{R}^q \times \mathbb{R}^{m+q+1}$  is the compressed influence matrix, whose terms are exactly the non-zero elements of ${\partial{x_n}}/{\partial \theta}$. Eq. \ref{eq: reduced influence update} reduces the algorithm memory requirement by a factor of $q$ and leads to a lower time complexity of $\mathcal{O}(q (m+p+q))$. The detailed implementation of SnAp-1 that we proposed and further explanations regarding the latter, including the proof of Eqs. \ref{eq: standard RNN immediate Jacobian short version} and \ref{eq: reduced influence update}, can be found respectively in Algorithm \ref{alg:RNN-SnAp-1} and Appendix \ref{appendix: SnAp-1}.

\subsubsection{DNI as a future-facing algorithm}
\label{section:material and methods - DNI}

RTRL, UORO, and SnAp-1 can be categorized as past-facing within the framework proposed in \cite{marschall2020unified} since the direction of the parameter update vector $\Delta \theta$ can be described using the sum of all the past instantaneous loss gradients instead of only the "current" one as we do here for simplicity. By contrast, in DNI, the gradient update $\Delta \theta$ is proportional to the sum of all future instantaneous losses:
\begin{align}
\Delta \theta &\propto \sum_{t=n}^{+ \infty} \frac{\partial L_{t+1}}{\partial \theta}(\theta_n) \\
&\approx \sum_{t=n}^{+ \infty} \frac{\partial L_{t+1}}{\partial x}(x_{n+1}) \frac{\partial F_{\text{st}}}{\partial \theta}(x_n, u_n, \theta_n) \label{eq:DNI approximation of influence matrix} \\ 
&= c_n \frac{\partial F_{\text{st}}}{\partial \theta}(x_n, u_n, \theta_n) \label{eq: DNI gradient loss}
\end{align} 
The line vector $c_n = \sum_{t=n}^{+ \infty} \frac{\partial L_{t+1}}{\partial x}(x_{n+1})$ in the expression above is called the credit assignment vector or error signal\footnote{One makes the approximation that the influence matrix ${\partial{x_n}}/{\partial \theta}$ is close to the immediate Jacobian ${\partial{F_{\text{st}}}}/{\partial \theta}$ in Eq. \ref{eq:DNI approximation of influence matrix}.}. It can be developed as follows\footnotemark:
\begin{align}
c_n &= \frac{\partial L_{n+1}}{\partial x}(x_{n+1}) + \sum_{t=n+1}^{+ \infty} \frac{\partial L_{t+1}}{\partial x}(x_{n+2}) \frac{\partial{F_{\text{st}}}}{\partial x}(x_{n+1}) \\
&= \nabla_x L_{n+1}^T + c_{n+1} D_{n+1} \label{eq: credit assignment development}
\end{align} 

\footnotetext{$\nabla_x L_{n+1}$, appearing in Eq. \ref{eq: credit assignment development}, is the column vector obtained by transposing $\partial L_{n+1} / \partial x$. More generally, in this article, the gradient notation $\nabla$ denotes column vectors (whereas the partial derivative of a scalar with respect to a vector is a row vector).}

In DNI, one assumes that there exists a coefficient matrix $A$ of size $(p+q+1, q)$ such that:
\begin{equation}\label{eq: credit assignment proportionality relationship}
c_n \approx \tilde{x}_n A 
\end{equation}
where $\tilde{x}_n$ is the line vector defined as the concatenation of the state and ground-truth output vectors at time index $n$, plus a unit bias component:
\begin{equation}\label{eq: DNI feature vector}
\tilde{x}_n = [x_n^T, y^{*T}_n, 1]
\end{equation}

At each time step, $A$ is estimated by fitting the synthetic gradient $\tilde{x}_n A$ to the true gradient $c_n$, that is, by minimizing the $l^2$ norm of the following difference:
\begin{align}
\tilde{x}_n A  - c_n
&= \tilde{x}_n A  - \nabla_x L_{n+1}^T - c_{n+1} D_{n+1} \\
&\approx \tilde{x}_n A  - \nabla_x L_{n+1}^T - \tilde{x}_{n+1} A  D_{n+1} \\
&\approx f(A) \label{eq: non-varying dynamic matrix}
\end{align}
where we define:
\begin{equation}\label{eq:function whose square norm minimization gives A in DNI}
f(A) = \tilde{x}_n A  - \nabla_x L_{n+1}^T - \tilde{x}_{n+1} A  D_n
\end{equation}

In the equations above, we successively replaced $c_n$ and $c_{n+1}$ with their expressions in Eqs. \ref{eq: credit assignment development} and \ref{eq: credit assignment proportionality relationship}, respectively. We also substituted $D_{n+1}$ with $D_n$ in Eq. \ref{eq: non-varying dynamic matrix}, assuming that these two quantities are approximately equal\footnote{This approximation is necessary because using $D_{n+1}$ would require having access to future information.}. Instead of minimizing $\| f(A) \|$ from scratch at every time step, we obtain $A$ via a single gradient descent step, using its estimate from the previous time step, $n$, to keep computation time low. The error signal and loss gradient direction are then successively derived via Eqs. \ref{eq: credit assignment proportionality relationship} and \ref{eq: DNI gradient loss}, respectively. Our main contribution to the DNI algorithm is showing that the gradient of $\| f(A) \| ^2$ can be expressed as follows (proof in Appendix \ref{appendix: derivation of the gradient of norm(f(A)) squared in DNI}):
\begin{equation}\label{eq:final expression of grad(norm(f(A)))}
\frac{1}{2}\frac{\partial \| f(A) \|^2}{\partial A}
= \tilde{x}_n^T f(A) - \tilde{x}_{n+1}^T f(A) D_n^T
\end{equation} 
The latter formula extends the corresponding expression in \cite{marschall2020unified} by incorporating the previously neglected term $\tilde{x}_{n+1}^T f(A) D_n^T$. The detailed implementation of DNI in our work and further related elements can be found in Algorithm \ref{alg:RNN-DNI} and Appendix \ref{appendix:DNI_alg}, respectively. 

\begin{figure*}[hbt!]
\centering
\begin{minipage}{.85\textwidth}
\begin{algorithm}[H]
\small
\caption{Sparse One-Step Approximation}
\label{alg:RNN-SnAp-1}
\begin{algorithmic}[1]
\State \textbf{Standard RNN parameters}
\State $L \in \mathbb{Z}_{>0} $: signal history length, $n_{\text{M}} = 3$: number of external markers considered
\State $m = 3 n_{\text{M}} L$, $q \in \mathbb{Z}_{>0}$, and $p = 3 n_{\text{M}}$: dimensions of the input, state, and output of the RNN
\State $\eta \in \mathbb{R}_{>0} $ and $\tau \in \mathbb{R}_{>0 }$: learning rate and gradient threshold
\State $\sigma_{\text{init}} \in \mathbb{R}_{>0} $: standard deviation of the Gaussian distribution of the initial weights
\State
\State \textbf{Standard RNN initialization}
\State $W_{a,n=1}$, $W_{b,n=1}$, $W_{c,n=1}$: synaptic weight matrices of respective sizes $q \times q$, $q \times (m+1)$, and $p \times q$, initialized following a Gaussian distribution with standard deviation $\sigma_{\text{init}}$
\State \textit{Notation :} $|W_{a}| = q^2$, $|W_{b}| = q(m+1)$, $|W_{c}| = p q$, and $|W| = q(m+p+q+1)$
\State $x_{n=1} := 0_{q \times 1}$: state vector
\State $\Delta\theta := 0_{1 \times |W|}$: gradient of the loss function with respect to the synaptic weights
\State
\State \textbf{Initialization specific to SnAp-1:} $J_n := 0_{q \times (m+q+1)}$: compressed influence matrix
\State 
\State \textbf{Learning and prediction}
\For{$n = 1,2,...$}
\State
\State \textbf{Forward propagation and computation of derivatives related to $W_{c,n}$ in standard RNNs} 
\State $z_n := W_{a,n}x_n + W_{b,n}u_n$, $x_{n+1} := \Phi(z_n)$ (hidden state update)
\State $y_{n+1} := W_{c,n} x_{n+1}$ (prediction), $e_{n+1} := y^*_{n+1} - y_{n+1}$ (error vector)
\State $[\Delta\theta_{1+|W_{a}|+|W_{b}|}, ..., \Delta\theta_{|W|}] := -[{(e_{n+1} x_{n+1}^T)}_{1,1}, ..., (e_{n+1} x_{n+1}^T)_{p,q}]$ (loss gradient $\partial{L_{n+1}} / \partial{W_{c,n}}$)
\State $\nabla_x L_{n+1} := - W_{c,n}^T e_{n+1}$ (gradient of the loss with respect to the states, column vector)
\State
\State \textbf{Computation of the loss gradient with respect to $W_a$ and $W_b$}
\State   $\overline{D_n} :=
  \begin{bmatrix}
    \Phi'(z_n)_1 (W_{a,n})_{1,1}& & 0 \\
    & \ddots & \\
    0 & & \Phi'(z_n)_q (W_{a,n})_{q,q}
  \end{bmatrix}$ 
  (sparse approximation)
\State $I_n := \Phi'(z_n) [x_n^T, u_n^T]$ (compressed immediate Jacobian matrix, Eq. \ref{eq:compact immediate Jacobian})
\State $J_{n+1} := \overline{D_n} J_n + I_n$ (reformulation of Eq. \ref{eq:influence update SnAp-1}) 
\State $[\Delta\theta_1, ..., \Delta\theta_{|W_{a}|+|W_{b}|}] := [ (\nabla_x L_{n+1} * J_{n+1})_{1,1}, ..., (\nabla_x L_{n+1} * J_{n+1})_{q,m+q+1}]$
\State \hspace{\algorithmicindent} $*$ is the element-wise multiplication operator.
\State \hspace{\algorithmicindent} Because $\nabla_x L_{n+1}$ is a column vector of size $q$ and $J_{n+1}$ is a matrix of size $q \times (m+q+1)$, 
\State \hspace{\algorithmicindent} each column of $J_{n+1}$ is multiplied element-wise by $\nabla_x L_{n+1}$ (broadcasting).
\State
\State \textbf{Parameter update in standard RNNs with gradient clipping} 
\State $\theta_n := [(W_{a,n})_{1,1}, ..., (W_{a,n})_{q,q}, (W_{b,n})_{1,1}, ..., (W_{b,n})_{q,m+1}, (W_{c,n})_{1,1}, ..., (W_{c,n})_{p,q}]$
\If{$ \|\Delta\theta\|_2 > \tau $} 
\State $\Delta\theta := \dfrac{\tau}{\|\Delta\theta\|_2} \Delta\theta $ (gradient clipping)
\EndIf 
\State $\theta_{n+1} := \theta_n - \eta \Delta\theta$ (weight update)
\State \resizebox{.9\hsize}{!}{$ W_{a,n+1} := \begin{bmatrix} 
(\theta_{n+1})_1 &...& (\theta_{n+1})_{q(q-1)+1}\\
... &...& ...\\
(\theta_{n+1})_q &...& (\theta_{n+1})_{|W_a|}
\end{bmatrix} $ ,
$ W_{b,n+1} := \begin{bmatrix} 
(\theta_{n+1})_{|W_a|+1} &...& (\theta_{n+1})_{|W_a|+qm+1}\\
... &...& ...\\
(\theta_{n+1})_{|W_a|+q} &...& (\theta_{n+1})_{|W_a|+|W_b|}
\end{bmatrix} $ }
\State $ W_{c,n+1} := \begin{bmatrix} 
(\theta_{n+1})_{|W_a|+|W_b|+1} & ... & (\theta_{n+1})_{|W_a|+|W_b|+p(q-1)+1}\\
... & ... & ...\\
(\theta_{n+1})_{|W_a|+|W_b|+p} & ... & (\theta_{n+1})_{|W_a|+|W_b|+|W_c|}
\end{bmatrix} $
\EndFor
\State
\State \textit{Convention: for} $A \in \mathbb{R}^{M} \times \mathbb{R}^{N}$ \textit{we define} $[A_{1,1}, ..., A_{M,N}] = [A_{1,1}, ..., A_{M,1}, A_{1,2}, ..., A_{M,N}]$
\end{algorithmic}
\end{algorithm}
\end{minipage}
\end{figure*}

\begin{figure*}[hbt!]
\centering
\begin{minipage}{.90\textwidth}
\begin{algorithm}[H]
\small
\caption{Decoupled Neural Interfaces}
\label{alg:RNN-DNI}
\begin{algorithmic}[1]
\State \textbf{Standard RNN initialization}
\State Parameters $L$, $n_{\text{M}}$, $m$, $q$, $p$, $\eta$, $\tau$, and $\sigma_{\text{init}}$: same as in lines 2-5 of Algorithm \ref{alg:RNN-SnAp-1}
\State Variables $W_{a,n=1}$, $W_{b,n=1}$, $W_{c,n=1}$, $x_{n=1}$, and $\Delta\theta$: same as in lines 8-11 of Algorithm \ref{alg:RNN-SnAp-1}
\State 
\State \textbf{Initialization of variables specific to DNI}
\State $\eta_A \in \mathbb{R}_{>0} $: learning rate associated with the credit assignment update
\State $\tilde{x}_{n=1} := [0_{1 \times (p+q)}, 1]$: line feature vector, including a bias term, for linear prediction of the credit assignment
\State $A_n$: coefficient matrix associated with credit assignment, of size $(p+q+1) \times q$, whose elements are initialized following a normal distribution $ \mathcal{N}(0,\, \sigma^2 = 1/q)$
\State 
\State \textbf{Learning and prediction}
\For{$n = 1,2,...$}
\State
\State \textbf{Forward propagation and computation of derivatives related to $W_{c,n}$ in standard RNNs} 
\State Computation of $z_n$, $x_{n+1}$, $y_{n+1}$, $e_{n+1}$, $\partial{L_{n+1}} / \partial{W_{c,n}}$, and $\nabla_x L_{n+1}$: same as in lines 19-22 of Algorithm \ref{alg:RNN-SnAp-1}
\State
\State \textbf{Computation of the loss gradient with respect to $W_a$ and $W_b$}
\State $D_n := \Phi'(z_n) * W_{a,n}$ (dynamic matrix, * denotes the element-wise and column-wise multiplication)
\State $\tilde{x}_{n+1} := [x_{n+1}^T, y^{*T}_{n+1}, 1]$ (features for credit assignment prediction, Eq. \ref{eq: DNI feature vector})
\State $f(A_n) := \tilde{x}_n A_n - \nabla_x L_{n+1}^T - \tilde{x}_{n+1} A_n D_n$ (function whose squared $l^2$ norm we aim to minimize, Eq. \ref{eq:function whose square norm minimization gives A in DNI})
\State $\Delta{A} := \tilde{x}_n^T f(A_n) - \tilde{x}_{n+1}^T f(A_n) D_n^T$ (gradient of $\| f \| ^2$ evaluated at $A_n$, Eq. \ref{eq:final expression of grad(norm(f(A)))}) 
\State $A_{n+1} := A_n - \eta_A \Delta{A}$  (update of the linear coefficients associated with credit assignment estimation)
\State $c_n := \tilde{x}_n A_{n+1}$ (credit assignment vector, Eq. \ref{eq: credit assignment proportionality relationship})
\State $\varphi_n := c_n^T * \Phi'(z_n)$ (auxiliary variable)
\State $[\Delta\theta_1, ..., \Delta\theta_{|W_{a}|+|W_{b}|}] := [(\varphi_n [x_n^T, u_n^T])_{1,1}, ..., (\varphi_n [x_n^T, u_n^T])_{q,m+q+1}]$ (proof in Appendix \ref{appendix: derivative of loss with respect to non-output weights})
\State
\State \textbf{Parameter update in standard RNNs with gradient clipping} 
\State Computation of $W_{a,n+1}$, $W_{b,n+1}$, and $W_{c,n+1}$: same as in lines 34-40 of Algorithm \ref{alg:RNN-SnAp-1}
\State
\EndFor
\end{algorithmic}
\end{algorithm}
\end{minipage}
\end{figure*}

\subsection{Experimental design}
\label{section: experimental design}

In the following, we represent the normalized 3D motion of marker $j \in \{1, 2, 3\}$ at time $t_k$ as $\vec{u}_j(t_k) = [u_j^x(t_k), u_j^y(t_k), u_j^z(t_k)]$. The RNN input is formed by concatenating the vectors $\vec{u}_j(t_n)$, ..., $\vec{u}_j(t_{n+L-1})$ for each marker $j$. Here, $L$ denotes the SHL expressed in number of time steps. Feeding the displacement information of the three markers altogether to the prediction algorithm helps leverage information concerning the correlations between each object's motion. The output vector $y_{n+1}$ comprises their positions at time $t_{n+L+h-1}$, with $h$ denoting the horizon value, also expressed in number of time steps (Eq. \ref{eq:RNN_in_out_def}).
\begin{equation} \label{eq:RNN_in_out_def}
u_n
=
\begin{pmatrix}
1 \\
u_1^x(t_n)\\
u_1^y(t_n)\\
u_1^z(t_n)\\
...\\
u_3^z(t_n)\\
u_1^x(t_{n+1})\\
...\\
u_3^z(t_{n+L-1})\\
\end{pmatrix} , 
\quad
y_{n+1}
=
\begin{pmatrix}
u_1^x(t_{n+L+h-1})\\
u_1^y(t_{n+L+h-1})\\
u_1^z(t_{n+L+h-1})\\
...\\
u_3^z(t_{n+L+h-1})\\
\end{pmatrix}
\end{equation}

We compare RNNs trained with RTRL, UORO, SnAp-1, and DNI with SVR with an RBF kernel \cite{drucker1996support, smola2004tutorial} and linear methods, namely LMS and multivariate linear regression (Table \ref{table:models comparison}). To provide baseline scenarios for comparison, we also include results when using the latest input $[u_1^x(t_n), ..., u_3^z(t_n)]$ as the predicted value $y_{n+1}$, which we refer to as "no prediction," and when initializing the hidden layer weights randomly and then "freezing" them during inference. We denote the second configuration as "RNN with fixed weights," although the output layer parameters are still updated at every time step. Last, to assess the contribution of the $\tilde{x}_{n+1}^T f(A) D_n^T$ term in the proposed update for DNI in Eq. \ref{eq:final expression of grad(norm(f(A)))}, we evaluate the performance of a baseline with a simplified update rule neglecting that term (i.e., only the $\tilde{x}_n^T f(A)$ term is kept), as an ablation experiment. RNNs updated using the gradient descent rule (and online algorithms in general) may exhibit instability. Therefore, we clip the estimated gradient of the instantaneous loss (Eq. \ref{eq:loss function}) with respect to the weight vector $\vec{\nabla}_{\theta} L_n $ for RTRL, UORO, SnAp-1, DNI, LMS, and also for the case of an RNN with a fixed hidden layer, when $ \| \vec{\nabla}_{\theta} L_n \|_2 > \tau$ \cite{pascanu2013difficulty}. We set the threshold $\tau$ to the same value, $\tau = 100.0$, for each of these algorithms instead of the lower value, $\tau = 2.0$, selected in \cite{pohl2022prediction}. 

Compared to the grid of hyperparameter values in \cite{pohl2022prediction}, we chose a higher upper limit for the number of hidden units (180 instead of 90), as that study showed that more hidden units led, on average, to higher prediction performance. One exception was RTRL, whose hidden layer size was kept under $q=40$ units because of its higher computational complexity $\mathcal{O}(q^4)$. We set the standard deviation of the normal distribution of the initial RNN parameters to $\sigma_{\text{init}} = 0.02$, as it was found in the same article that this value experimentally minimized the nRMSE and that $\sigma_{\text{init}}$ was the hyperparameter whose variations had the least influence on cross-validation accuracy. We also examined learning rates, $\eta$, lower than those in~\cite{pohl2022prediction} due to our higher gradient clipping threshold $\tau$. We varied the range of $\eta$ for LMS depending on the input signal frequency, $f$, because we experimentally found that LMS performance with respect to $\eta$ was particularly sensitive to changes in $f$ despite prior input signal normalization. In other words, without such adaptation, no common range for $\eta$ made LMS perform well for all the frequencies $f$ considered. A higher value of $\eta$ was needed at low frequencies due to relatively greater variations in the input signal and vice-versa. By contrast, the same range of values of $\eta$ was adopted regardless of the input frequency for all the RNN algorithms considered, as that experimentally resulted in acceptable performance. Regarding DNI, we set $\eta_A = 0.002$ as the learning rate used for updating $A$ at each time step $n$ and did not apply gradient clipping during this process.

\begin{table*}[htb!]
\small
\setlength{\tabcolsep}{6pt}
\begin{center}
\begin{tabular}{llll}
\hline
Prediction    &  Mathematical                               & Development set        & Range of hyperparameters \\
method        &  model                                      & partition              & for cross-validation     \\
\hline
\hline
RTRL, UORO    & $x_{n+1} = \Phi(W_{a,n} x_n + W_{b,n} u_n)$ & Training 30s           & $\eta \in \{0.005, 0.01, 0.02\}$ \\
SnAp-1, DNI   & $y_{n+1} = W_{c,n} x_{n+1}$                 & Cross-validation 30s   & $L \in \{1.2\text{s}, 2.4\text{s}, ..., 6.0\text{s}\}$ \\
              &                                             &                        & $q \in \{30, 60, 90, ..., 180\}$ except \\ 
              &                                             &                        & for RTRL: $q_{\text{RTRL}} \in \{10, 25, 40\}$\\
\hline
LMS           & $y_{n+1} = W_n u_n$                         & Training 30s           & $L \in \{1.2\text{s}, 2.4\text{s}, ..., 6.0\text{s}\}$ \\
              &                                             & Cross-validation 30s   & 3.33Hz: $\eta \in \{ 0.0002, 0.0005, 0.001\}$\\
              &                                             &                        & 10.0Hz: $\eta \in \{ 0.0001, 0.0002, 0.0005\}$\\
              &                                             &                        & 30.0Hz: $\eta \in \{ 0.00005, 0.0001, 0.0002\}$\\
\hline
Linear        & $y_{n+1} = W u_n$                           & Training 54s           & $L \in \{1.2\text{s}, 2.4\text{s}, ..., 6.0\text{s}\}$ \\
regression    &                                             & Cross-validation 6s    & \\
\hline
RNN with      & $x_{n+1} = \Phi(W_a x_n + W_b u_n)$         & Training 30s           & $\eta \in \{0.005, 0.01, 0.02\}$ \\
a frozen layer  & $y_{n+1} = W_{c,n} x_{n+1}$                 & Cross-validation 30s   & $L \in \{1.2\text{s}, 2.4\text{s}, ..., 6.0\text{s}\}$ \\
              &                                             &                        & $q \in \{30, 60, 90, ..., 180\}$ \\ 
\hline
Kernel SVR    & $y_{n+1, i} = \sum_{k<=N_{\text{train}}} \alpha_{k,i} K(x_k, x_n) + \beta_i$ & Training 54s & $L \in \{1.2\text{s}, 2.4\text{s}, ..., 6.0\text{s}\}$ \\
              & with $K(x_k, x_l) = \text{exp}(- ||x_k - x_l ||^2/(2 \sigma^2))$       & Cross-validation 6s & $\sqrt{2} \sigma \in \{100, 200, 500, 1000\}$ \\
              &                                             &                        & $\epsilon \in \{0.005, 0.01, 0.02, 0.05\}$ \\ 
              &                                             &                        & $C \in \{100, 200, 500, 1000\}$ \\ 
\hline                            
\end{tabular}
\end{center}
\caption{Outline of the different forecasting algorithms compared in this work. The input vector \text{\normalsize $u_n$} and output vector \text{\normalsize $y_{n+1}$}, containing respectively the past and predicted positions, and appearing in the second column, are defined in Eq. \ref{eq:RNN_in_out_def}. The fourth column describes the hyperparameter range used during cross-validation with grid search. $\eta$, $\sigma_{\text{init}}$, $L$, and $q$ designate the learning rate, the standard deviation of the Gaussian distribution of the initial synaptic parameters, the SHL expressed in seconds\protect\footnotemark, and the hidden layer size, respectively. The matrices $W_n$ and $W$, of size $p \times (m+1)$, are used respectively in LMS and linear regression. The parameters $N_{\text{train}}$, $\sigma$, $\epsilon$, and $C$ intervening in kernel SVR are the (time) index of the last training example, the standard deviation of the Gaussian kernel, the half-width of the $\epsilon$-insensitive band, and the regularization coefficient controlling the penalty imposed on observations lying outside the $\epsilon$-margin \cite{drucker1996support, smola2004tutorial}. The SVR implementation that we used outputs a single scalar; the model with coefficients $(\alpha_{k,i},\beta_i)$ corresponds to the $i^{\text{th}}$ output, $y_{n+1, i}$, and the same hyperparameters (in the fourth column) are shared across those models.}
\label{table:models comparison}
\end{table*}

\footnotetext{For example, an SHL of 2.4s corresponds to 24 time steps (in the past) when the input is sampled at $f=10\text{Hz}$ and 72 time steps when $f=30\text{Hz}$.}

\begin{figure}[htb!]
	\centering
	\includegraphics[width=\columnwidth]{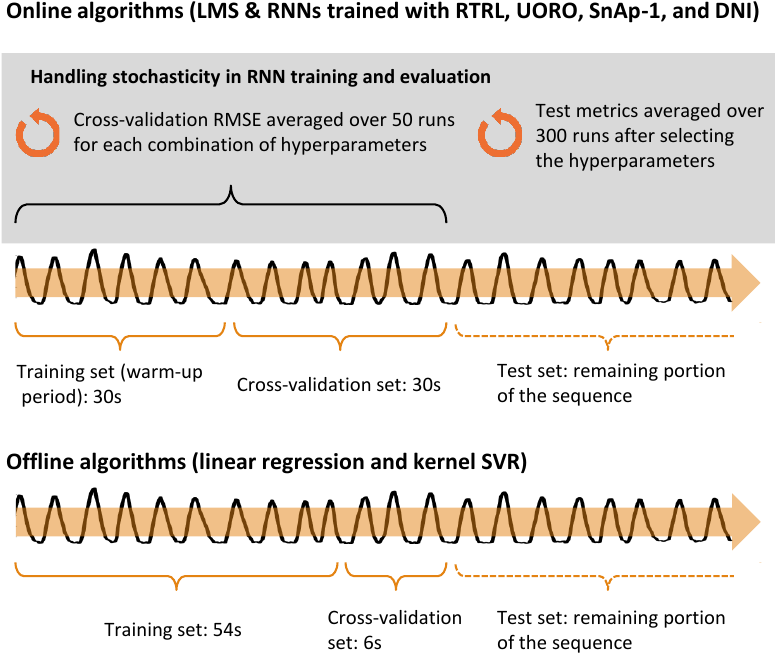}
	\caption{Partition of each nine-dimensional breathing sequence (containing the 3D positions of the three markers) into a training, cross-validation, and test set, and variability mitigation via metric averaging in the case of RNNs.}
	\label{fig:sequence_splitting}
\end{figure}

We perform prediction for horizons $h$ ranging from 0.1s to 2.1s to study its impact on performance. When the input signal is sampled at 3.33Hz, the values of $h$ considered are exactly in $\{0.3\text{s}, 0.6\text{s},...,  2.1\text{s}\}$, and when it is sampled at 10Hz or 30Hz, the horizon range is exactly $h \in \{0.1\text{s}, 0.2\text{s}, ...,  2.1\text{s}\}$\footnotemark. The forecasting models in our work are subject-specific. In other words, learning is conducted solely with one respiratory sequence (i.e., the information from the 3D positions of the three markers for a single subject) among the nine in the dataset, and we conduct testing using that exact sequence. Each time series undergoes division into training and development sets spanning together 1 minute and the remaining test set (Fig. \ref{fig:sequence_splitting}). The data from 0s to 30s is used as the training set, except for kernel SVR and linear regression, as allocating a larger proportion of data to training generally improves accuracy in offline learning. We select the data between 0s and 54s, and between 54s and 1min as the training set and cross-validation set, respectively, for the two latter algorithms. Online algorithms do not stop learning, as weights are constantly updated. Hence, the "training set" mentioned above refers to a "warm-up" period for those. To facilitate learning, we subtract from the original time series the mean of the training set, $\mu_{\text{train}}$, and divide it by the standard deviation of the training set, $\sigma_{\text{train}}$, to obtain the inputs $u_n$. The predicted values $y_n$ are then replaced by $\sigma_{\text{train}} y_n + \mu_{\text{train}}$. Evaluation with the test set is conducted using the hyperparameters minimizing the RMSE of the cross-validation set during the grid search process. To remove the bias from random initialization of the RNN weights and stochastic updates, we average the RMSE of the cross-validation set over $n_{\text{cv}} = 50$ successive runs given each set of hyperparameters. Similarly, each evaluation metric computed using the test set is averaged over $n_{\text{test}} = 300$ runs.

\footnotetext{Even if the horizons expressed in seconds are the same for $f=10\text{Hz}$ and $f=30\text{Hz}$, they differ when expressed in number of time steps. For example, prediction with a response time equal to 0.3s corresponds to 1 time step, 3 time steps, and 9 time steps ahead for a signal sampled at 3.33Hz, 10Hz, and 30Hz, respectively.}

Those metrics include the RMSE, nRMSE, MAE, and maximum error of the test set. Additionally, we calculate the jitter of the test set, which quantifies the average jump between two successive positions or data points in the predicted signal. On the one hand, increased fluctuations in the latter can pose challenges regarding robot control during treatment. On the other hand, constant prediction minimizes jitter; therefore, there is a trade-off between jitter and accuracy. The precise definitions of those metrics can be found in \cite{pohl2022prediction}. Specifically, they use 3D Euclidean distances and averaging over the three markers altogether, and the nRMSE is normalized using the standard deviation of the ground-truth signal\footnotemark. The experimental setting and overall characteristics of the RNNs considered in this study can be found in Table \ref{table:RNNs_configuration}.

\footnotetext{We use a generalization of standard deviation normalization for three multi-dimensional signals, each corresponding to the position of one marker (cf Eq. 11 in \cite{pohl2022prediction}).}

\begin{table}[thb!]
\small
\setlength{\tabcolsep}{1pt}
\begin{tabular}{ll}
\hline
RNN parameters &  \\
\hline
Output layer size       & $p = 3 n_{\text{M}} $ \\
Input layer size        & $m = 3 n_{\text{M}} L$\\
Number of hidden layers & 1 \\
Size of the hidden layer & $q$ \\
Activation function $\phi$ &  Hyperbolic tangent \\
Training algorithm &  RTRL, UORO, SnAp-1, or DNI\\
Optimization method & Stochastic gradient descent \\ 
Gradient clipping   & Yes, with threshold $\tau = 100$ \\
Weight initialization & Gaussian $\mathcal{N}(0, \sigma_{\text{init}} = 0.02)$\\
Input data normalization & Yes, with training set statistics\\
Cross-validation metric & RMSE \\
Nb. of runs for cross-val. & $n_{\text{cv}}=50$\\
Nb. of runs for evaluation & $n_{\text{test}}=300$ \\
Training time interval & 30s \\
Cross-val. time interval & 30s \\
\hline
\end{tabular}
\caption{Parameters related to the experimental setup and RNN configuration. $n_{\text{M}}$ and $L$ designate the number of external markers and the SHL expressed in number of time steps, respectively.}
\label{table:RNNs_configuration}
\end{table}

\section{Results}

\subsection{Accuracy and oscillatory behavior of the prediction}
\label{section:accuracy and jitter}

\begin{table*}[tb!]
\small 
\setlength{\tabcolsep}{8pt}
\begin{center}
\begin{tabular}{lllll}
\hline
Error     &  Prediction & Sampling       & Sampling   & Sampling \\
type      &  method     & at 3.33Hz     & at 10Hz   & at 30Hz \\
\hline \hline
MAE       & RTRL          & $1.3513 \pm 0.0010$ & $0.6531 \pm 0.0003$ & $0.3680 \pm 0.0001$ \\
(in mm)   & UORO          & $1.2266 \pm 0.0016$ & $0.5347 \pm 0.0003$ & $0.3087 \pm 0.0001$ \\
          & SnAp-1        & $1.0890 \pm 0.0005$ & $0.4933 \pm 0.0001$ & $0.3132 \pm 0.0001$ \\
          & DNI (full update rule for $A$)      & $1.1215 \pm 0.0026$ & $0.5433 \pm 0.0004$ & $0.3131 \pm 0.0001$ \\          
          & DNI (simplified update of $A$) & $1.1925 \pm 0.0014$ & $0.6035 \pm 0.0003$ & $0.3067 \pm 0.0001$ \\  
          & LMS               & 1.6204              & 1.0276              & 0.5931              \\
          & Linear regression   & 4.9290              & 4.5683              & 5.1387              \\
          & No prediction     & 3.6363              & 3.3780              & 3.3888              \\
          & RNN with a frozen layer & $1.3963 \pm 0.0029$ & $2.5890 \pm 0.0079$ & $2.1707 \pm 0.0044$ \\
          & Kernel SVR    & 2.7676              & 3.2639              & 3.7243 \\
 
\hline               
RMSE      & RTRL          & $1.8817 \pm 0.0016$ & $0.9260 \pm 0.0004$ & $0.4837 \pm 0.0002$ \\
(in mm)   & UORO          & $1.7406 \pm 0.0025$ & $0.7549 \pm 0.0007$ & $0.4015 \pm 0.0002$ \\
          & SnAp-1        & $1.5309 \pm 0.0009$ & $0.6994 \pm 0.0001$ & $0.4142 \pm 0.0001$ \\
          & DNI (full update rule for $A$)          & $1.5464 \pm 0.0035$ & $0.7522 \pm 0.0007$ & $0.4018 \pm 0.0002$ \\          
	      & DNI (simplified update of $A$)  & $1.6425 \pm 0.0020$ & $0.8522 \pm 0.0005$ & $0.3940 \pm 0.0001$ \\            
          & LMS               & 2.2126 & 1.4192 & 0.7967 \\
          & Linear regression   & 6.9404 & 6.3739 & 7.2572 \\
          & No prediction     & 4.6975 & 4.3753 & 4.3827 \\
          & RNN with a frozen layer & $1.9191 \pm 0.0047$ & $3.5159 \pm 0.0118$ & $3.0316 \pm 0.0070$ \\
          & Kernel SVR    & 3.5994              & 4.2378              & 4.8180  \\
\hline
nRMSE     & RTRL          & $0.40319 \pm 0.00021$ & $0.19499 \pm 0.00006$ & $0.10156 \pm 0.00002$ \\
(no unit) & UORO          & $0.38435 \pm 0.00039$ & $0.16602 \pm 0.00012$ & $0.08573 \pm 0.00003$ \\
          & SnAp-1        & $0.33468 \pm 0.00017$ & $0.15674 \pm 0.00003$ & $0.08965 \pm 0.00002$ \\
          & DNI (full update rule for $A$)          & $0.33658 \pm 0.00045$ & $0.16466 \pm 0.00011$ & $0.08784 \pm 0.00003$ \\          
	      & DNI (simplified update of $A$)  & $0.36277 \pm 0.00035$ & $0.18729 \pm 0.00009$ & $0.08639 \pm 0.00003$ \\                      
          & LMS               & 0.48956 & 0.31420 & 0.17462 \\
          & Linear regression   & 1.66276 & 1.53738 & 1.80327 \\
          & No prediction     & 1.02853 & 0.95947 & 0.96017 \\
          & RNN with a frozen layer & $0.43079 \pm 0.00102$ & $0.79985 \pm 0.00252$ & $0.67087 \pm 0.00148$ \\   
          & Kernel SVR    & 0.80091             & 0.95998              & 1.10122  \\
\hline
Max error & RTRL          & $9.754 \pm 0.015$ & $5.929 \pm 0.008$ & $3.539 \pm 0.005$ \\
(in mm)   & UORO          & $9.759 \pm 0.022$ & $5.483 \pm 0.010$ & $3.294 \pm 0.007$ \\
          & SnAp-1        & $8.449 \pm 0.014$ & $5.602 \pm 0.006$ & $3.588 \pm 0.005$ \\
          & DNI (full update rule for $A$)          & $8.668 \pm 0.020$ & $5.500 \pm 0.009$ & $2.940 \pm 0.005$ \\   
		  & DNI (simplified update of $A$)  & $8.937 \pm 0.018$ & $6.119 \pm 0.008$ & $3.055 \pm 0.005$ \\              
          & LMS               & 11.090 & 8.576 & 5.854 \\
          & Linear regression   & 35.262 & 32.537 & 36.715 \\
          & No prediction     & 15.797 & 15.173 & 15.429 \\
          & RNN with a frozen layer & $9.285 \pm 0.026$ & $13.956 \pm 0.048$ & $14.031 \pm 0.040$ \\  
          & Kernel SVR    & 15.501             & 16.819              & 18.854  \\
\hline
Jitter    & RTRL          & $1.2944 \pm 0.0017$ & $0.6466 \pm 0.0006$ & $0.3044 \pm 0.0002$ \\
(in mm)   & UORO          & $1.4230 \pm 0.0020$ & $0.6552 \pm 0.0004$ & $0.3224 \pm 0.0001$ \\
          & SnAp-1        & $1.6189 \pm 0.0010$ & $0.7200 \pm 0.0002$ & $0.3923 \pm 0.0002$ \\
          & DNI (full update rule for $A$)          & $1.8678 \pm 0.0025$ & $0.8443 \pm 0.0005$ & $0.3123 \pm 0.0001$ \\          
	      & DNI (simplified update of $A$)  & $2.0301 \pm 0.0018$ & $0.9787 \pm 0.0005$ & $0.3169 \pm 0.0001$ \\             
          & LMS               & 2.0479 & 1.4480 & 0.8636 \\
          & Linear regression   & 1.7860 & 0.8219 & 0.4147 \\
          & No prediction     & 1.1550 & 0.4395 & 0.2456 \\
          & RNN with a frozen layer & $1.6821 \pm 0.0057$ & $4.8245 \pm 0.0158$ & $4.1680 \pm 0.0088$ \\                            
          & Kernel SVR    & 0.9864             & 0.3911              & 0.1558  \\
\hline
\end{tabular}
\end{center}
\caption{Performance of each forecasting algorithm for different input signal sampling rates. Each measure in the table represents the average of a given performance metric of the test set over the nine records and response times $h$ between 0.1s and 2.1s, using the best hyperparameters for each individual sequence and value of $h$. The 95\% confidence intervals for the mean metrics corresponding to the RNNs are computed assuming a Gaussian distribution\protect\footnotemark. DNI with the full update rule for $A$ refers to our implementation (Section \ref{section:material and methods - DNI}), whereas DNI with the simplified update of $A$ refers to the implementation in \cite{marschall2020unified} where the second term in the right-hand side of Eq. \ref{eq:final expression of grad(norm(f(A)))} is neglected; in the rest of the article, "DNI" refers to the former version, unless specified otherwise.}
\label{table:pred perf}
\end{table*}

\footnotetext{The confidence interval calculation method is the same as that described in Section 2.4 in \cite{pohl2022prediction}. When $f=3.33\text{Hz}$, the performance metrics are averaged over the response times $0.3\text{s} \leq h \leq 2.1\text{s}$.}

\begin{figure*}[htb!]
    \centering
    \subfloat[\normalsize Prediction with an RNN trained with UORO]{{\includegraphics[width=.95\textwidth]{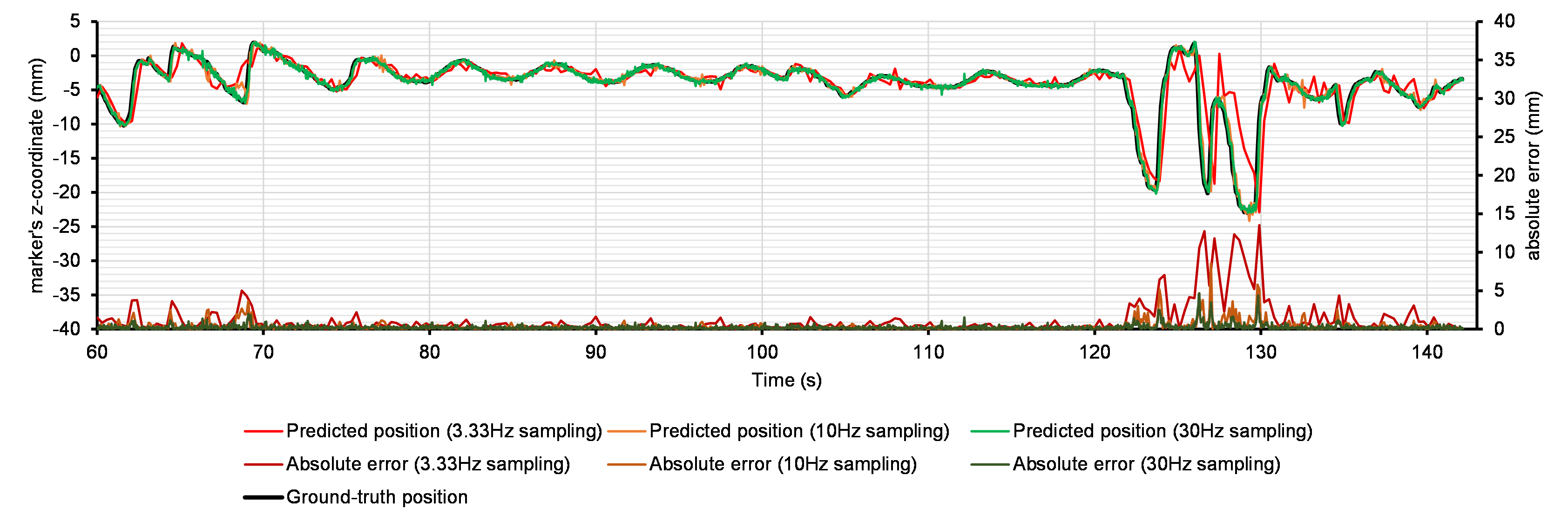} }}
    \quad
    \subfloat[\normalsize Prediction with an RNN trained with SnAp-1]{{\includegraphics[width=.95\textwidth]{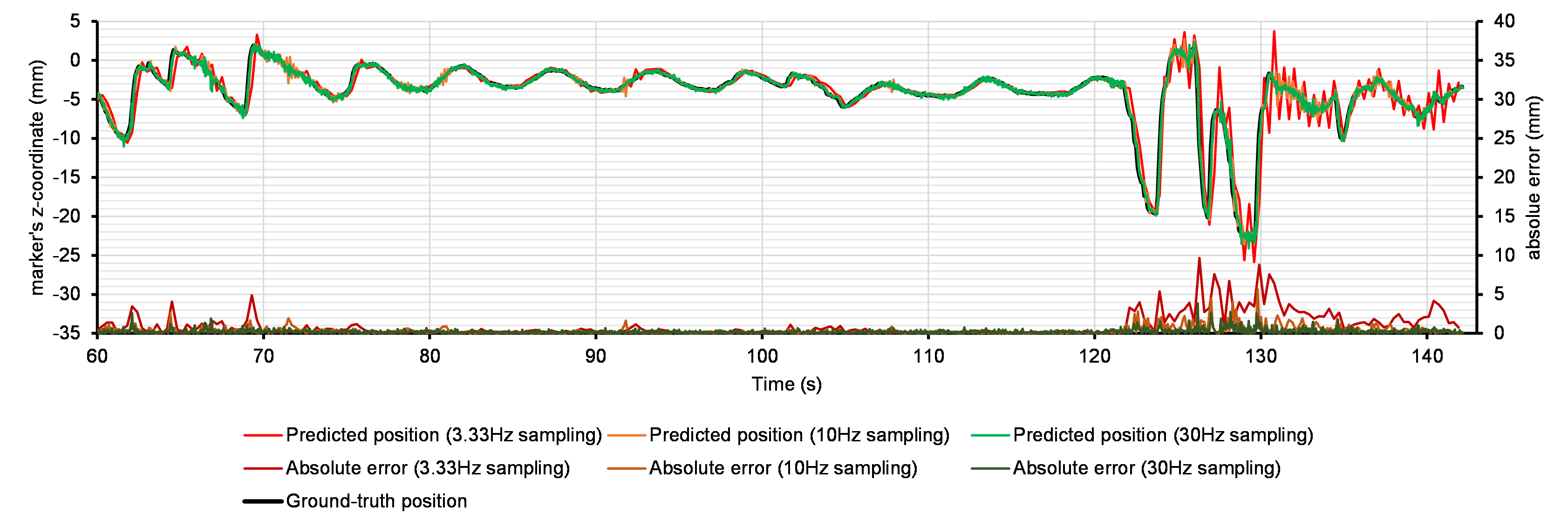} } \label{fig:coordz_marker3_seq4_SnAp-1}}%
    \quad
    \subfloat[\normalsize Prediction with an RNN trained with DNI]{{\includegraphics[width=.95\textwidth]{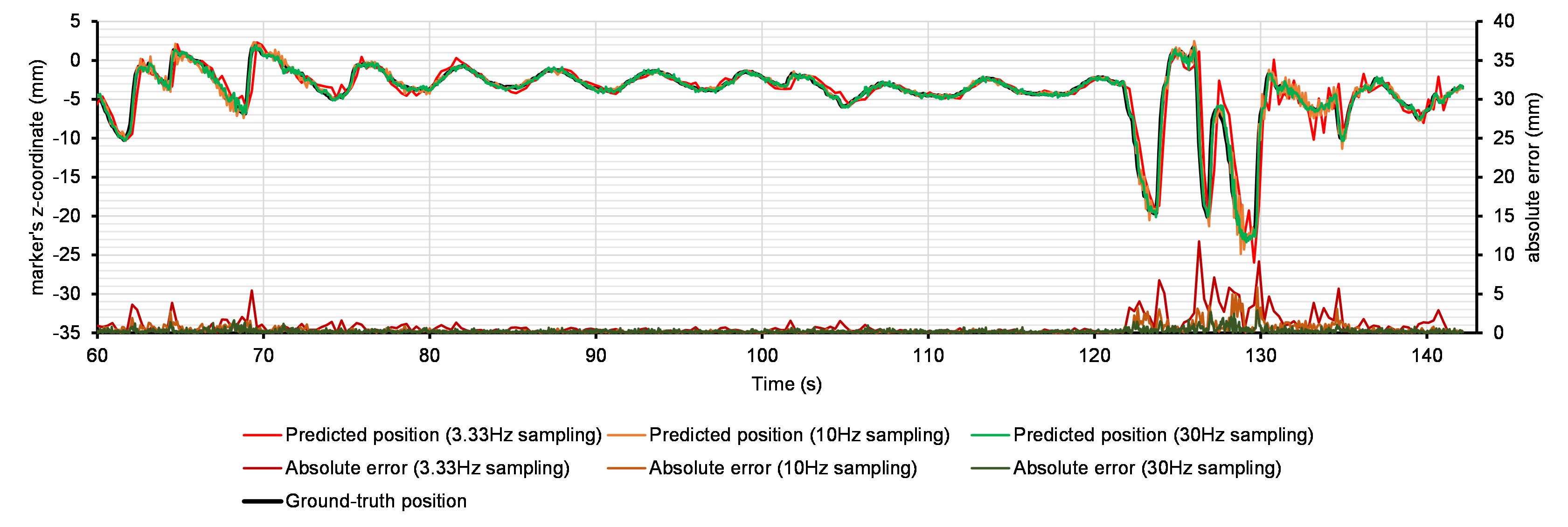} }}%
    \caption{Comparison between the ground-truth z-coordinate (longitudinal axis) of marker 3 in sequence 4 (person laughing and talking) and its prediction with UORO, SnAp-1, and DNI for different input signal sampling frequencies $f$. The forecasting horizon is set to 1.2s. For each algorithm and value of $f$, we selected the optimal hyperparameters for that horizon.}%
    \label{fig:coordz_marker3_seq4}%
\end{figure*}

SnAp-1 achieved the lowest MAEs, RMSEs, and nRMSEs averaged over all the sequences and response times considered at $f=3.33\text{Hz}$ and $f=10\text{Hz}$ (Table \ref{table:pred perf}). UORO attained the lowest nRMSE, and DNI with the simplified partial update rule for $A$, where the second term on the right-hand side of Eq. \ref{eq:final expression of grad(norm(f(A)))} was suppressed, reached the lowest MAE and RMSE, at $f=30\text{Hz}$. DNI with the full update rule consistently ranked second regarding these three errors on average across all records and horizons at 3.33Hz and 10Hz, except for the MAE at 10Hz, where it ranked third. For the rest of this article, "DNI" will denote our proposed version with the full update rule for $A$ (Eq. \ref{eq:final expression of grad(norm(f(A)))}) unless explicitly stated otherwise. UORO performed worse than SnAp-1 and DNI in terms of these three measures at 3.33Hz, as reflected in Fig. \ref{fig:coordz_marker3_seq4}. SnAp-1, UORO, and DNI respectively achieved the lowest maximum errors at 3.33Hz, 10Hz, and 30Hz, with some overlap of the confidence intervals of UORO and DNI at $f=10\text{Hz}$. LMS led to MAEs, RMSEs, and nRMSEs higher than those associated with the RNN algorithms considered by approximately 34\% at 3.33Hz, 83\% at 10Hz, and 87\% at 30Hz. Likewise, the maximum errors characterizing LMS were about 21\%, 52\%, and 75\% higher than those corresponding to the RNNs at 3.33Hz, 10Hz, and 30Hz, respectively\footnotemark. Kernel SVR performed worse than LMS regarding all the accuracy metrics. 

\footnotetext{These percentages correspond to averages of relative differences; we did not consider DNI with the partial update for $A$ when calculating those.\label{footnote: relative diffs not over DNI partial}}

\begin{figure}[htb!]
	\centering
	\includegraphics[width=\columnwidth]{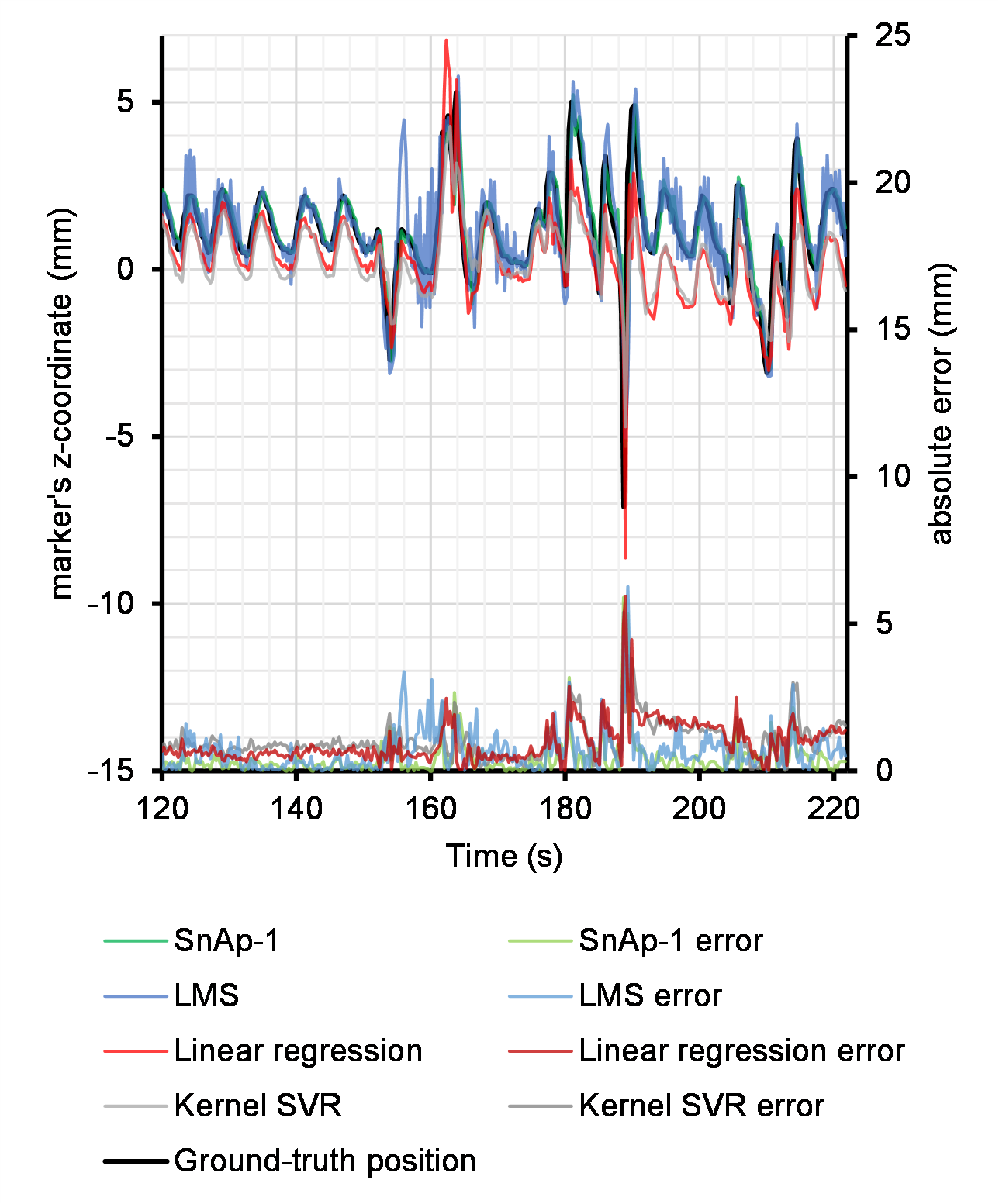}
	\caption{Comparison between the ground-truth z-coordinate (longitudinal axis) of marker 3 in sequence 1 (person talking) and its prediction with SnAp-1, LMS, linear regression, and kernel SVR at 3.33Hz. The forecasting horizon is set to $h=0.3\text{s}$; the hyperparameters selected for each method were those optimal for that record and value of $h$.}
	\label{fig:coordz_marker3_seq1}
\end{figure}

\begin{figure}[htb!]
	\centering
	\includegraphics[width=\columnwidth]{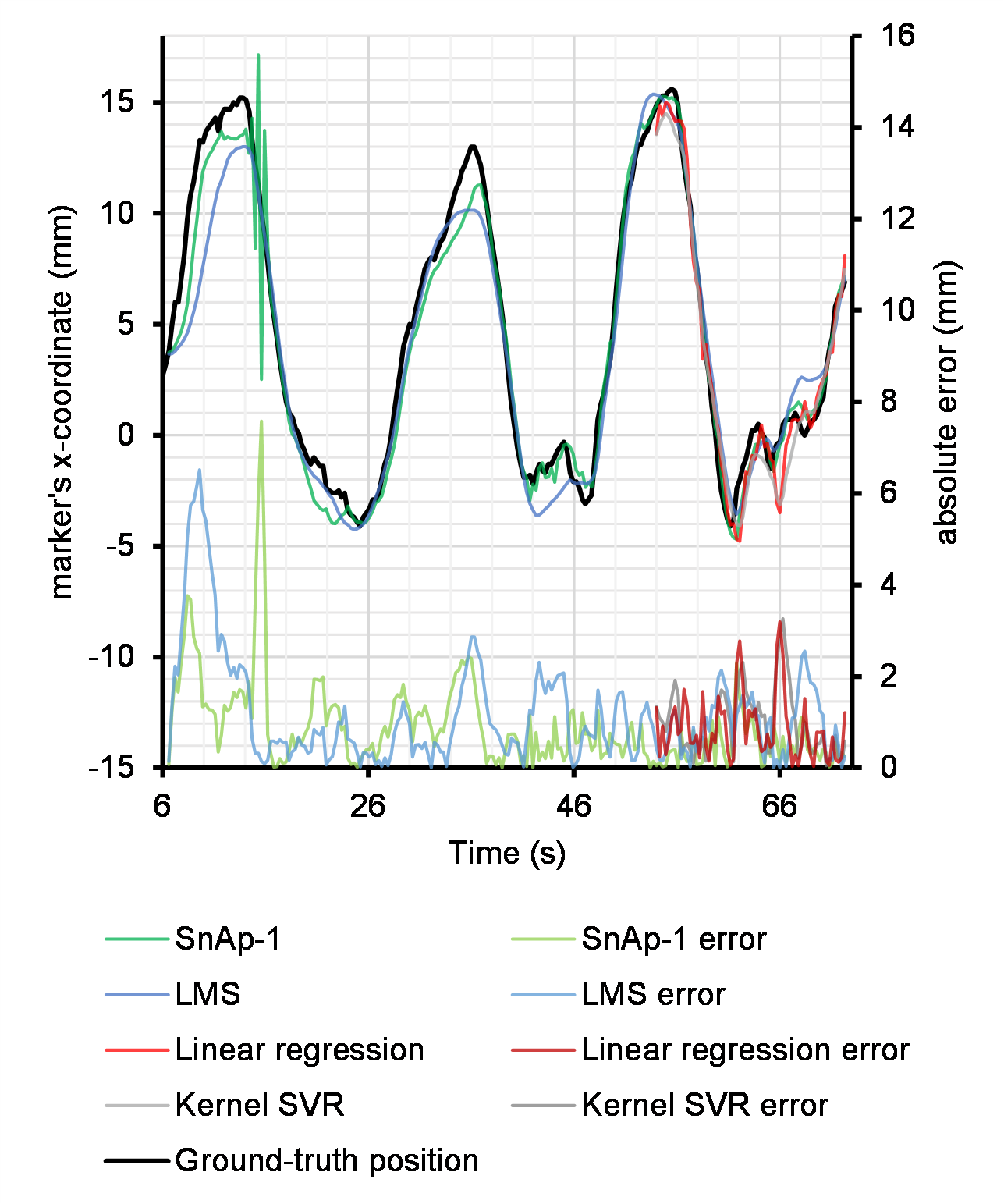}
	\caption{Comparison between the ground-truth x-coordinate of marker 1 in sequence 7 (respiratory pattern classified as "other" in \cite{krilavicius2016predicting} and characterized by high-amplitude slow motion) and its prediction with SnAp-1, LMS, linear regression, and kernel SVR at 3.33Hz. Linear regression and kernel SVR are fit using the data between 0s and 54s, so forecasting starts after that period for those two algorithms. By contrast, online algorithms can start predicting data sooner, although early time points are considered part of the warm-up interval. The horizon is set to $h=0.9\text{s}$; the hyperparameters selected for each method were those optimal for that record and value of $h$.}
	\label{fig:coordx_marker1_seq7}
\end{figure}

\begin{figure}[htb!]
	\centering
	\includegraphics[width=\columnwidth]{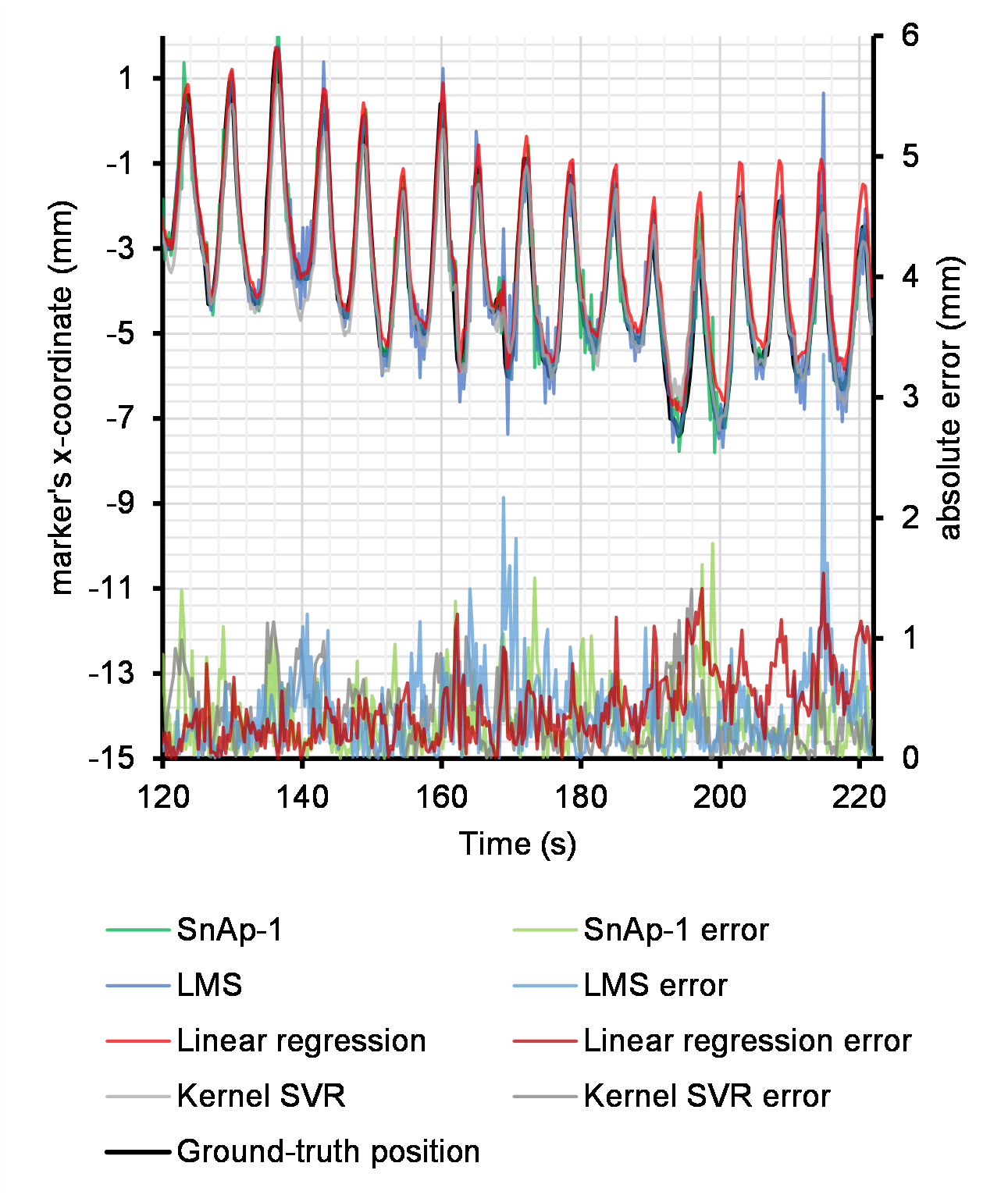}
	\caption{Comparison between the ground-truth x-coordinate of marker 3 in sequence 8 (normal breathing exhibiting drift) and its prediction with SnAp-1, LMS, linear regression, and kernel SVR at 3.33Hz. The forecasting horizon is set to $h=0.3\text{s}$; the hyperparameters selected for each method were those optimal for that record and value of $h$.}
	\label{fig:coordx_marker3_seq8}
\end{figure}

The lowest, second lowest, and third lowest jitter corresponded to kernel SVR, the non-prediction setting, and RTRL, respectively. Conversely, LMS and the RNN with fixed hidden layer parameters invariably resulted in the highest jitter regardless of $f$, except at $f=3.33\text{Hz}$, where DNI with the simplified update rule had the second highest jitter. The oscillatory behavior of LMS, observed in sequences 1 and 8, showcasing irregular breathing patterns and drift, was associated with high maximum errors, attained at $t \approx 184\text{s}$ and $t \approx 215\text{s}$ in these two examples, respectively (Figs. \ref{fig:coordz_marker3_seq1} and \ref{fig:coordx_marker3_seq8}). In general, the extreme phases of the respiratory cycle appeared the hardest to forecast, which was visible as well in the predictions associated with sequence 7, featuring deep and slow breathing (Fig. \ref{fig:coordx_marker1_seq7}). Kernel SVR tended to underestimate the x-coordinates of marker 3 in sequence 8 at those peaks when $120\text{s} \leq t \leq 150\text{s}$, linear regression overestimated them when $t \geq 192\text{s}$, while the predictions of LMS and SnAp-1 were more oscillatory around them (Fig. \ref{fig:coordx_marker3_seq8}). Nonetheless, the latter behavior may be less apparent at higher frequencies, as illustrated in Fig. \ref{fig:coordz_marker3_seq4_SnAp-1}. That is in agreement with the observations in the literature regarding chest video prediction, with some works mentioning the difficulty to predict the end-of-inhale phase due to its high fluctuations among cycles \cite{romaguera2021probabilistic, romaguera2023conditional}. Although SnAp-1 had the highest accuracy at 3.33Hz, very unstable motion was challenging to predict even at low horizons, as, for instance, no algorithm could reliably predict the local minimum of the z-coordinate of marker 3 in sequence 1 at $t \approx 184\text{s}$ (Fig. \ref{fig:coordz_marker3_seq1}). Furthermore, SnAp-1 might exhibit signs of instability, as evidenced by the large-amplitude oscillations appearing during the warm-up period near $t \approx 16\text{s}$ in sequence 7 (Fig. \ref{fig:coordx_marker1_seq7}).

\begin{figure*}[htb!]
    \centering
    \subfloat[\normalsize Sampling at 3.33Hz]{\includegraphics[width=.30\textwidth]{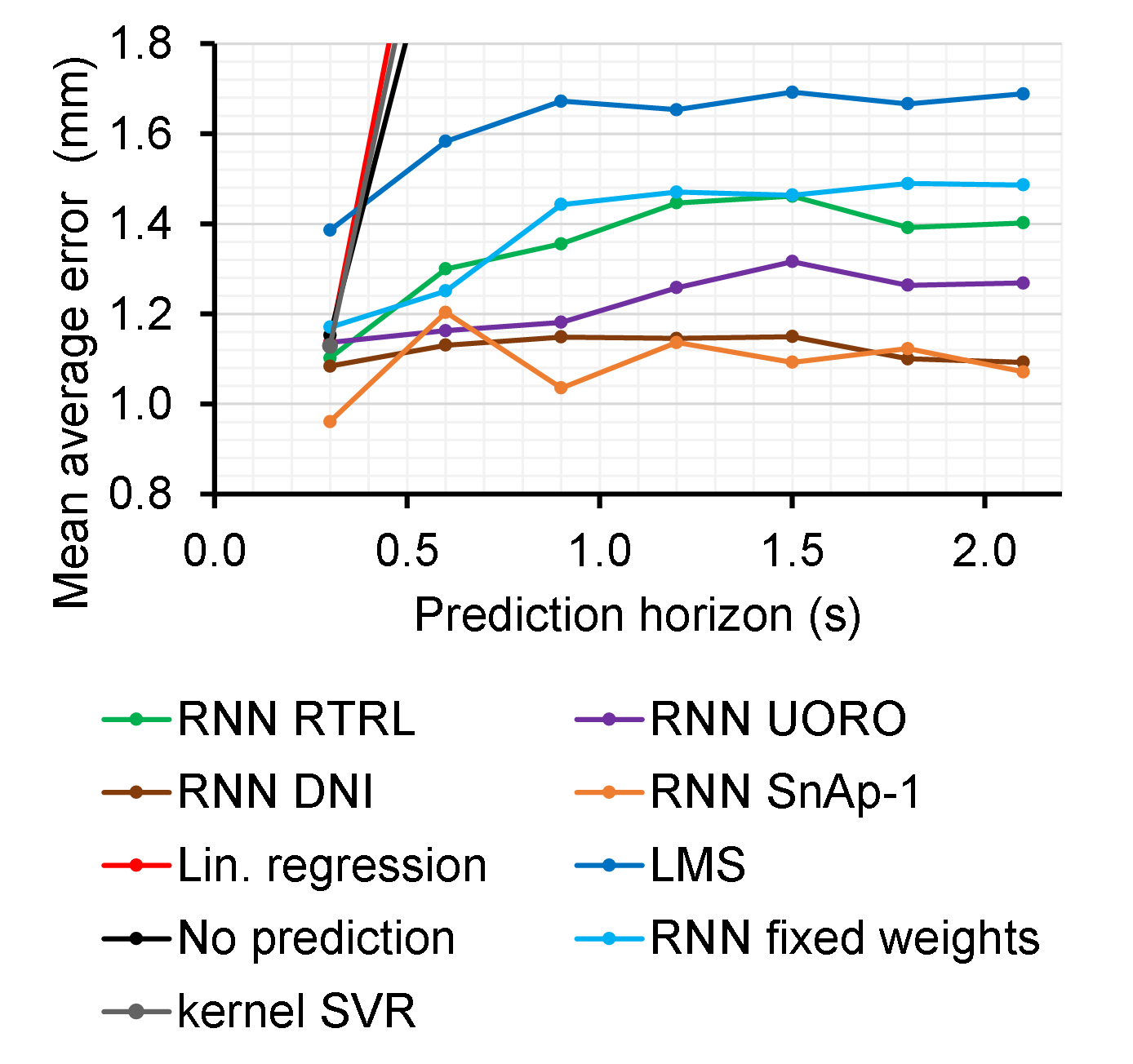} \label{fig:MAE_vs_horizon_3.33Hz}}%
    \quad
    \subfloat[\normalsize Sampling at 10.0Hz]{\includegraphics[width=.30\textwidth]{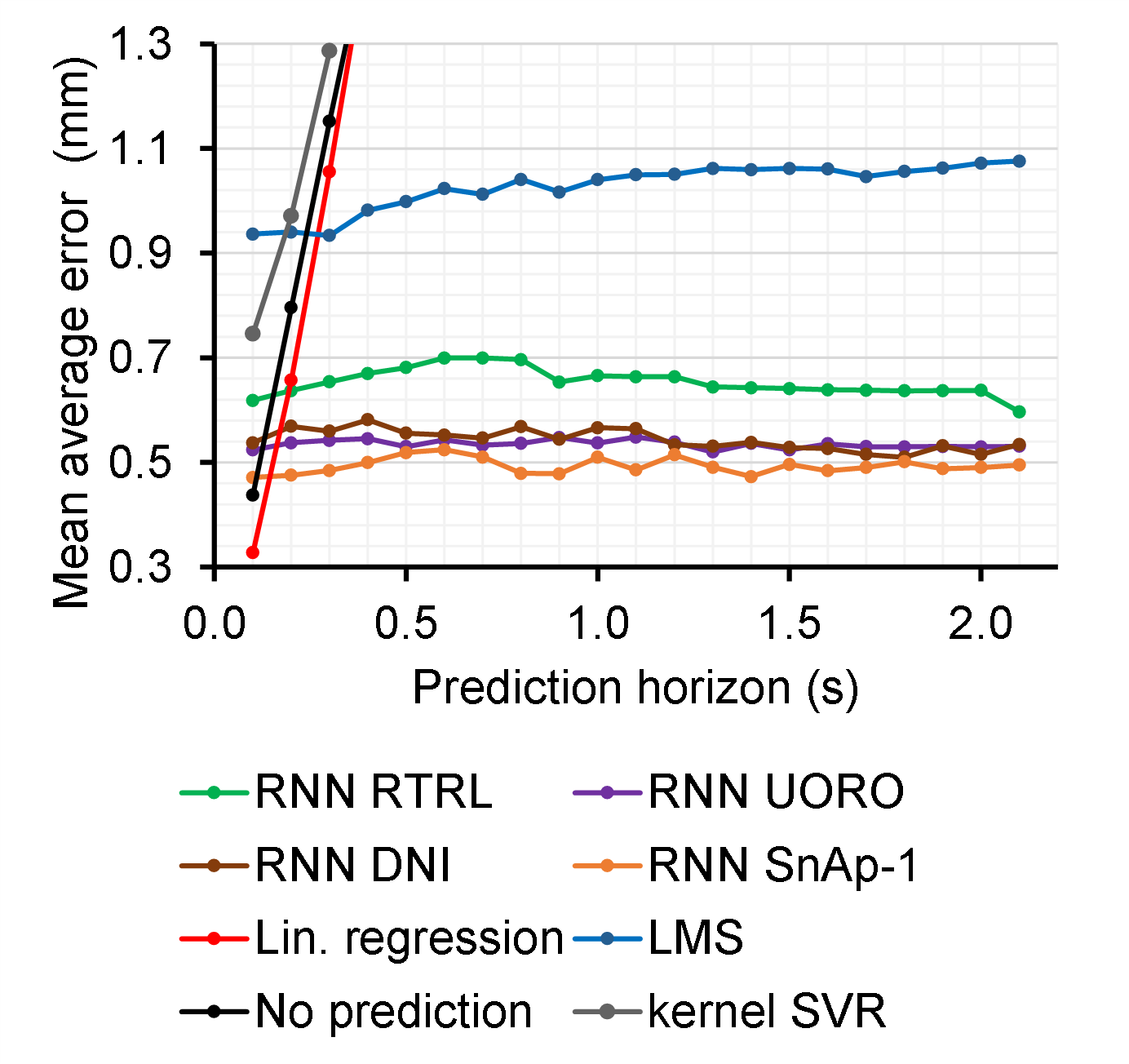} \label{fig:MAE vs horizon 10Hz}}%
    \quad
    \subfloat[\normalsize Sampling at 30.0Hz]{\includegraphics[width=.30\textwidth]{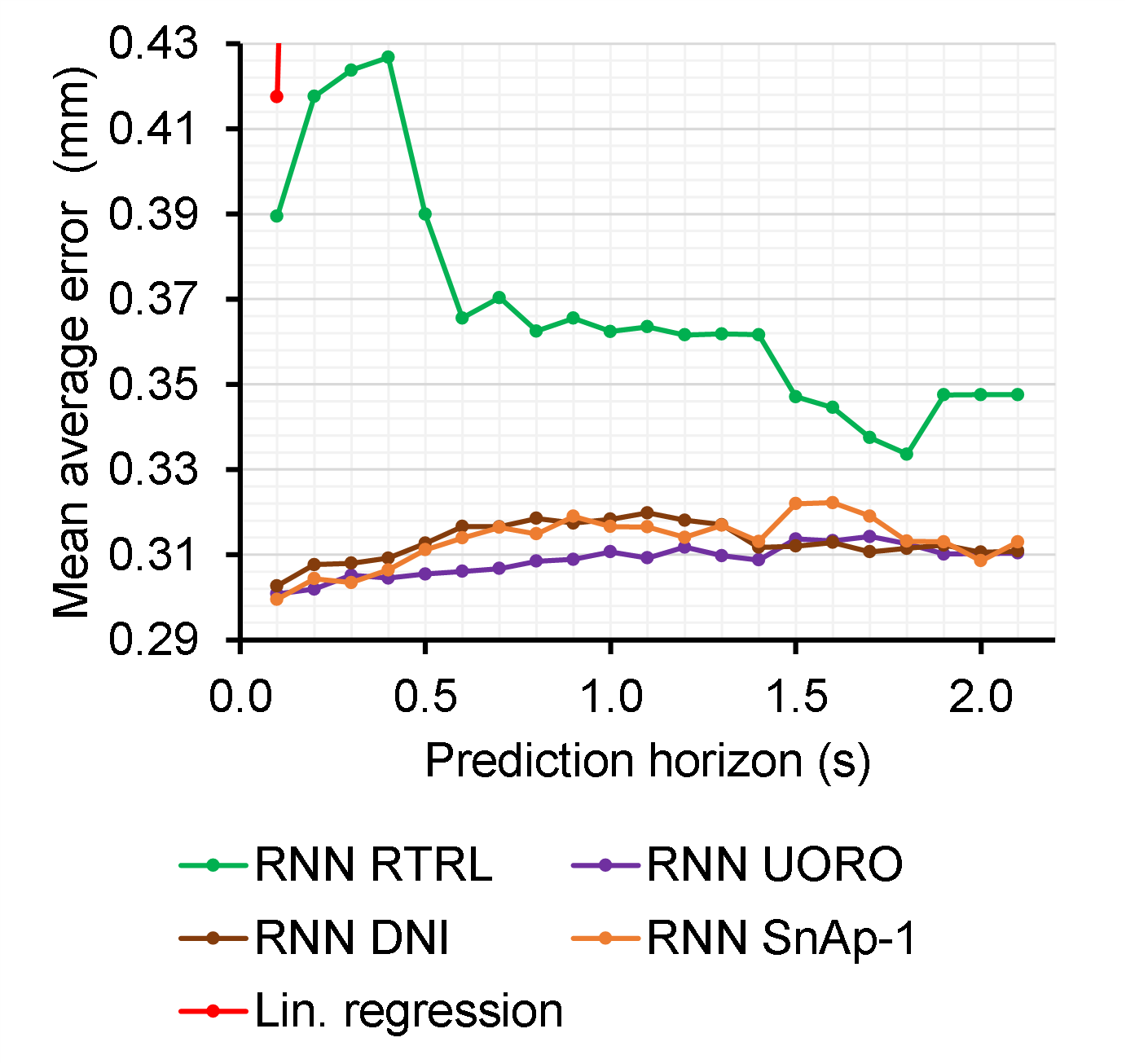} \label{fig:MAE vs horizon 30Hz}}%
    \caption{MAE of each algorithm as a function of the forecasting horizon for different input signal sampling rates. Each point represents the average MAE of the test set across the nine sequences for a given horizon using the best hyperparameters for that horizon (and each sequence individually)\protect\footnotemark.}%
    \label{fig:MAE_for_different_horizons}%
\end{figure*}

\begin{figure*}[htb!]
    \centering
    \subfloat[\normalsize Sampling at 3.33Hz]{\includegraphics[width=.30\textwidth]{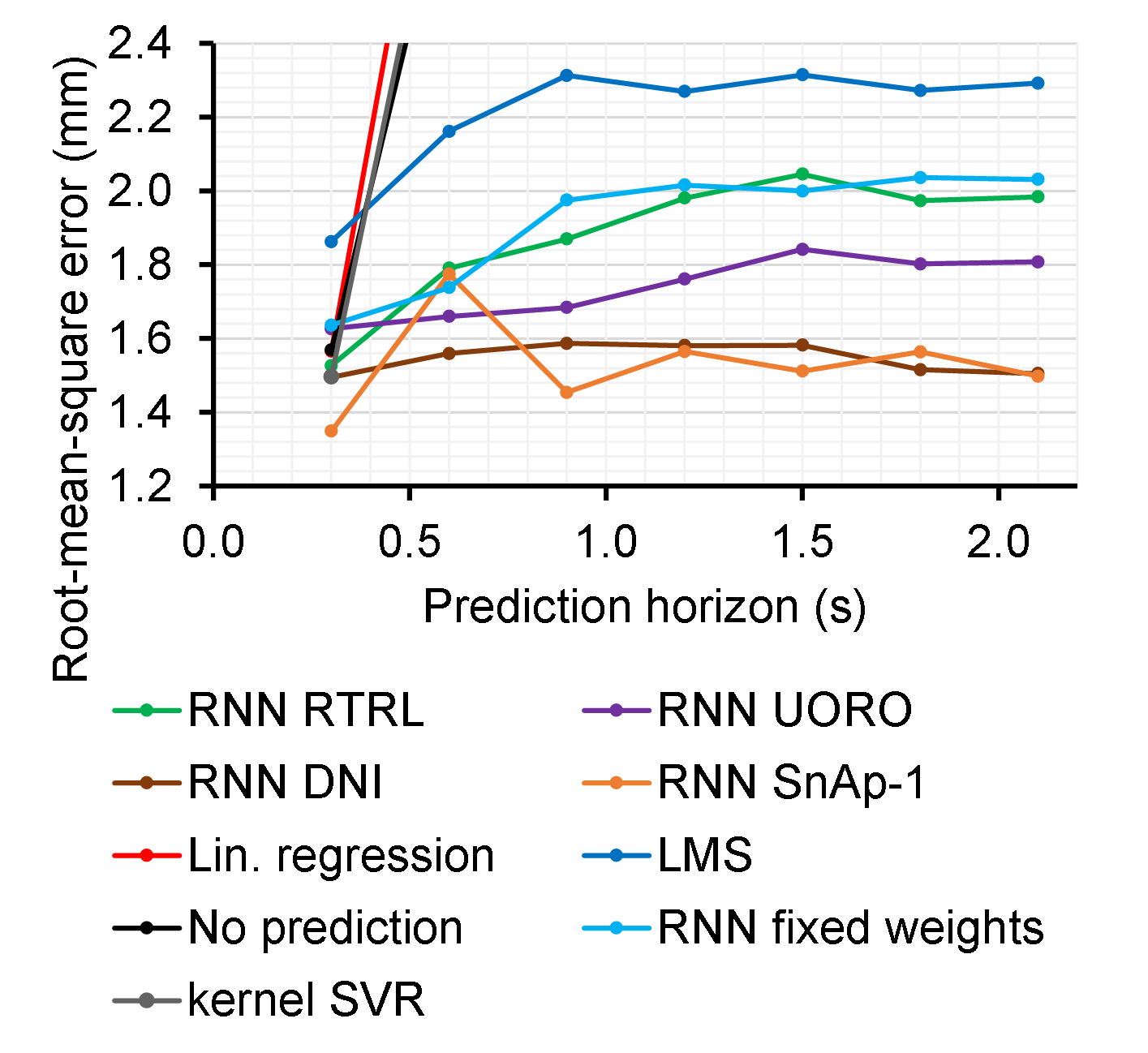} \label{fig:RMSE vs horizon 3.33Hz}}%
    \quad
    \subfloat[\normalsize Sampling at 10.0Hz]{\includegraphics[width=.30\textwidth]{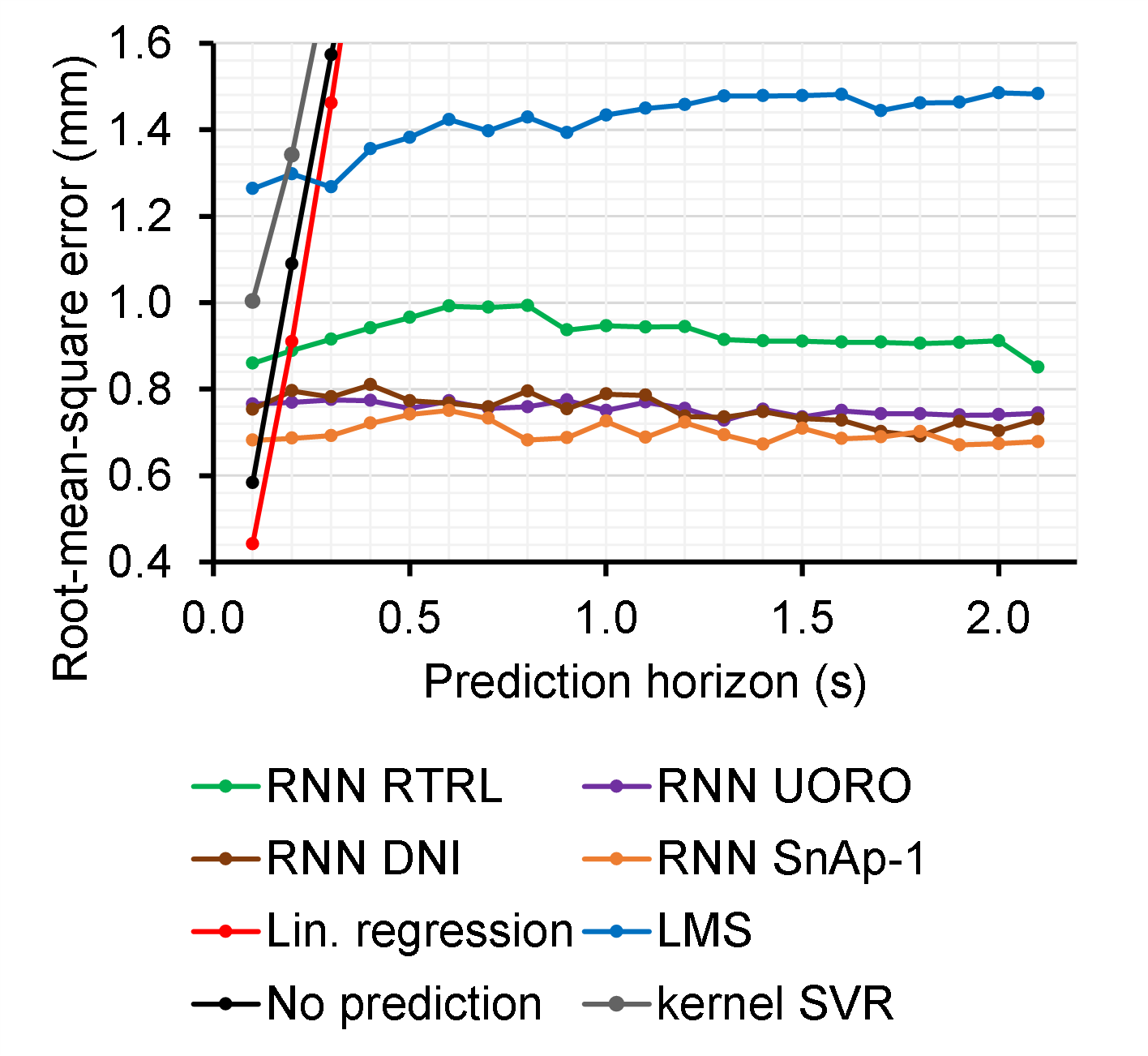} \label{fig:RMSE vs horizon 10Hz}}%
    \quad
    \subfloat[\normalsize Sampling at 30.0Hz]{\includegraphics[width=.30\textwidth]{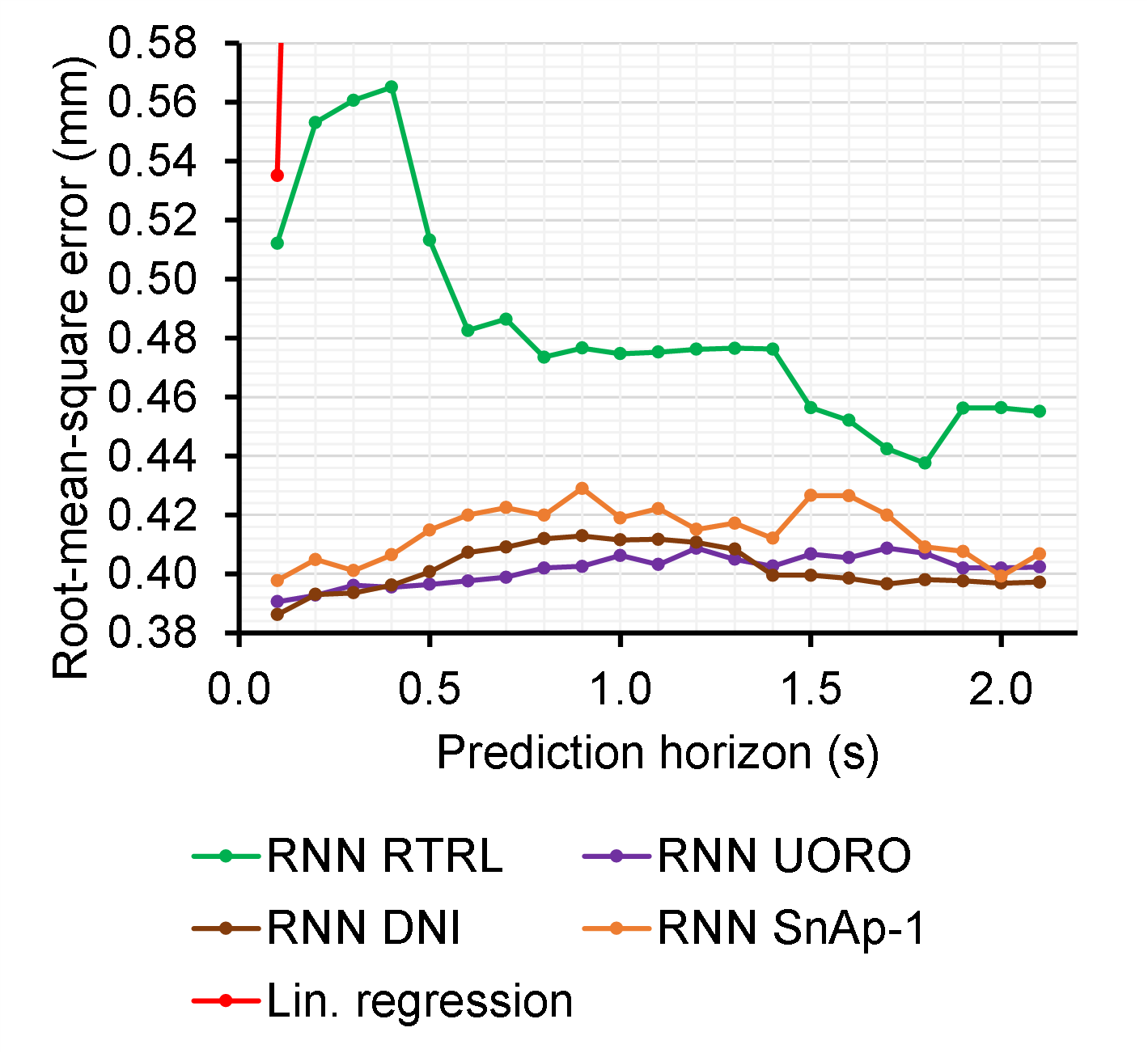} \label{fig:RMSE vs horizon 30Hz}}%
    \caption{RMSE of each algorithm as a function of the forecasting horizon for different input signal sampling rates. Each point represents the average RMSE of the test set across the nine sequences for a given horizon using the best hyperparameters for that horizon (and each sequence individually)\protect\footnotemark.}%
    \label{fig:RMSE_for_different_horizons}%
\end{figure*}

The MAEs, RMSEs, and nRMSEs associated with the online learning algorithms for RNNs decreased by approximately 53\% and 44\%, as $f$ increased from 3.33Hz to 10Hz and from 10Hz to 30Hz, respectively\footnote{Same as footnote \ref{footnote: relative diffs not over DNI partial}}. Similarly, concerning LMS, the same errors were reduced by 36\% and 44\%, as $f$ increased from 3.33Hz to 10Hz and from 10Hz to 30Hz, respectively. That is because more information is available for making a single prediction at higher sampling rates. The RNN with fixed weights led to lower performance on average over the horizons and sequences considered compared with the other RNN algorithms, except in a few cases at $f=3.33\text{Hz}$ \footnote{The RNN with fixed weights achieved a maximum error approximately 5\% lower than that of RTRL and UORO at $f=3.33\text{Hz}$.}. This confirms that efficient representation learning at the hidden layer level impacts performance positively. The comparable accuracy of RTRL and the RNN with frozen weights at the latter sampling rate can be attributed to the relatively low maximum value of $q$ allowed for RTRL in our experiments\footnote{cf Figs. \ref{fig:MAE_vs_horizon_3.33Hz} and \ref{fig:RMSE vs horizon 3.33Hz}, for instance}. Almost all the observed errors and jitters associated with DNI trained with the simplified partial update rule for $A$ were higher than those corresponding to our proposed update (Eq. \ref{eq:final expression of grad(norm(f(A)))}), demonstrating the latter's effectiveness. The MAE, RMSE, and nRMSE corresponding to DNI with the full update rule (which we refer to as "DNI" in the rest of the article unless mentioned otherwise) at 30Hz were slightly higher, but its jitter was lower; perhaps the additional term helps smooth prediction, reducing fluctuations, while introducing a slight bias.

\footnotetext{The errors corresponding to an RNN with fixed hidden layer weights were very high compared to the other methods at 10Hz and 30Hz. Therefore, they were not plotted in the corresponding graphs to improve readability. Same remark for LMS, kernel SVR, and the scenario without prediction when $f=30\text{Hz}$. \label{footnote: high RNN and LMS errors}} 

\footnotetext{Same as footnote \ref{footnote: high RNN and LMS errors}.} 

\begin{figure*}[htb!]
    \centering
    \subfloat[\normalsize Sampling at 3.33Hz]{\includegraphics[width=.30\textwidth]{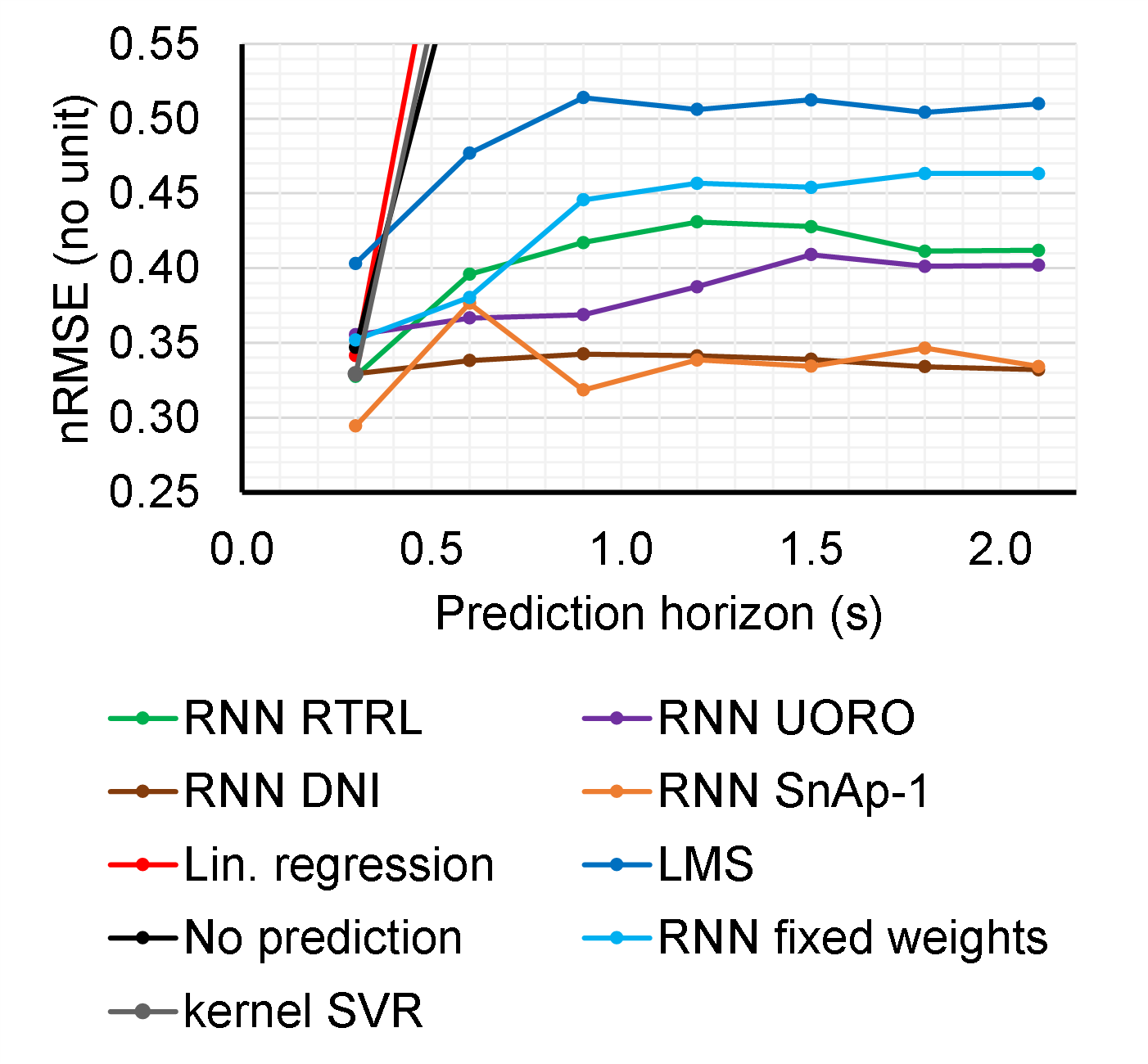} \label{fig:nRMSE vs horizon 3.33Hz}}%
    \quad
    \subfloat[\normalsize Sampling at 10.0Hz]{\includegraphics[width=.30\textwidth]{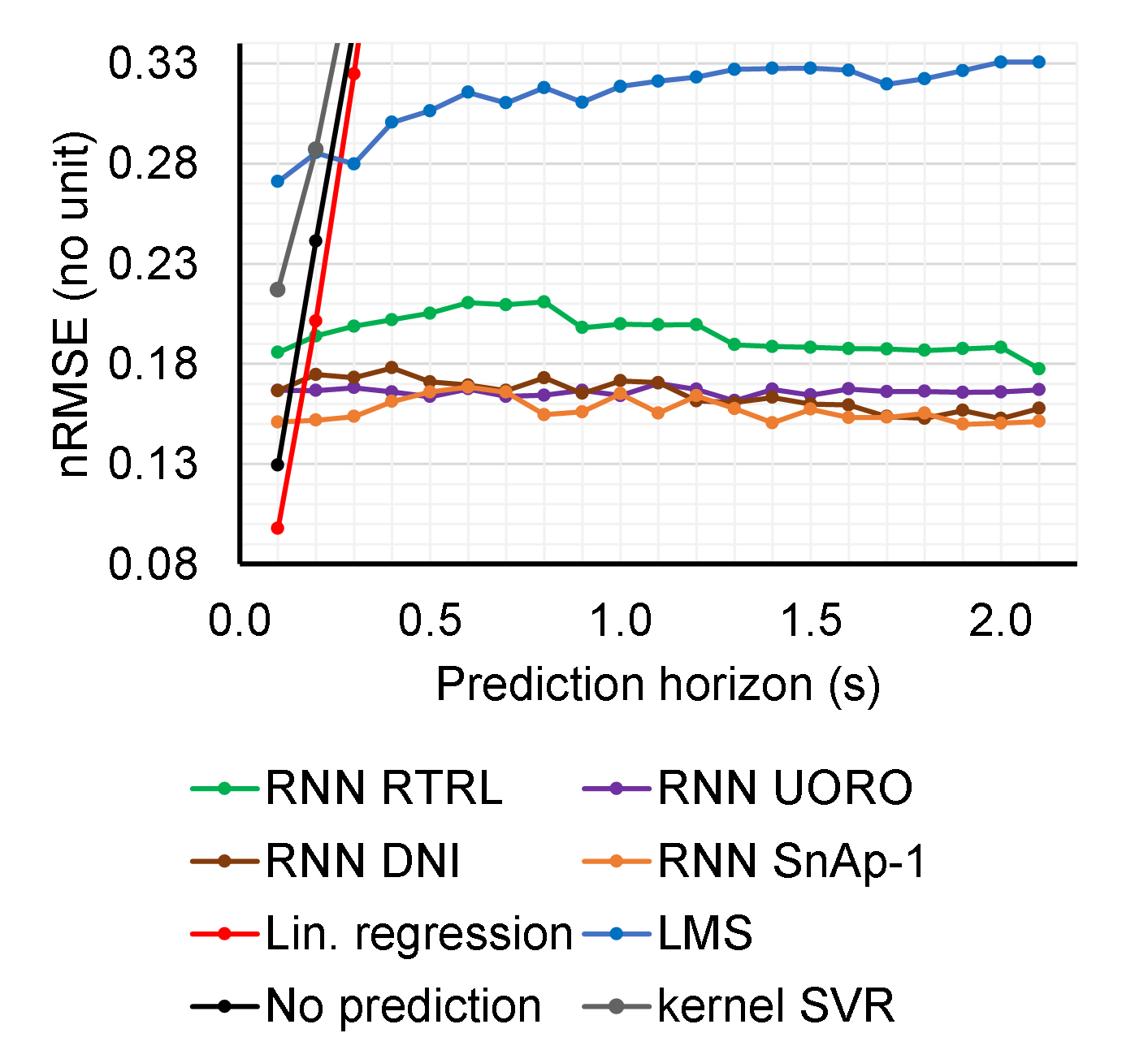} \label{fig:nRMSE vs horizon 10Hz}}%
    \quad
    \subfloat[\normalsize Sampling at 30.0Hz]{\includegraphics[width=.30\textwidth]{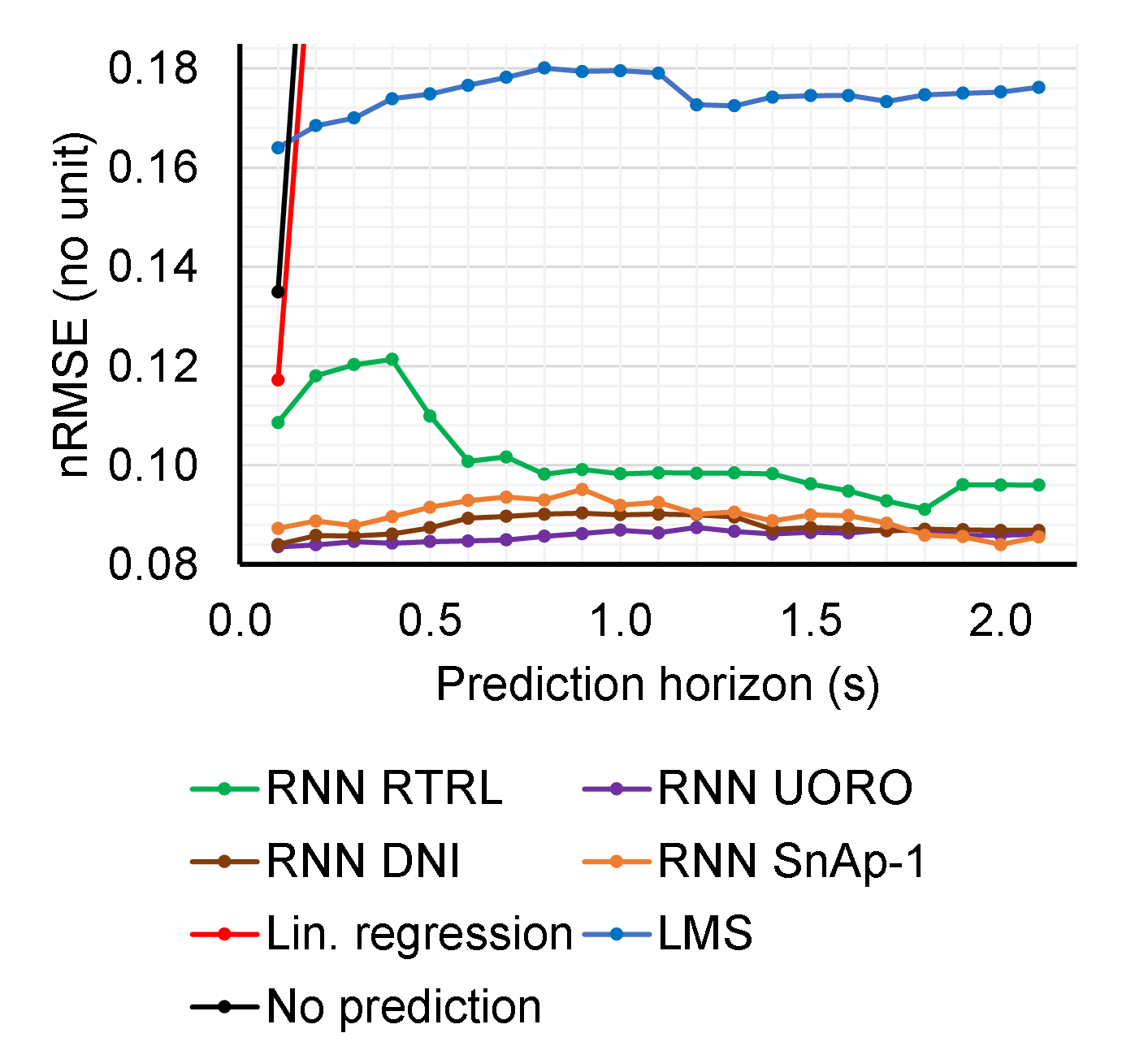} \label{fig:nRMSE vs horizon 30Hz}}%
    \caption{nRMSE of each algorithm as a function of the forecasting horizon for different input signal sampling rates. Each point represents the average nRMSE of the test set across the nine sequences for a given horizon using the best hyperparameters for that horizon (and each sequence individually)\protect\footnotemark.}%
    \label{fig:nRMSE_for_different_horizons}%
\end{figure*}

\footnotetext{Same remark as footnote \ref{footnote: high RNN and LMS errors} concerning kernel SVR and the RNN with a frozen hidden layer.} 

\begin{figure*}[htb!]
    \centering
    \subfloat[\normalsize Sampling at 3.33Hz]{\includegraphics[width=.30\textwidth]{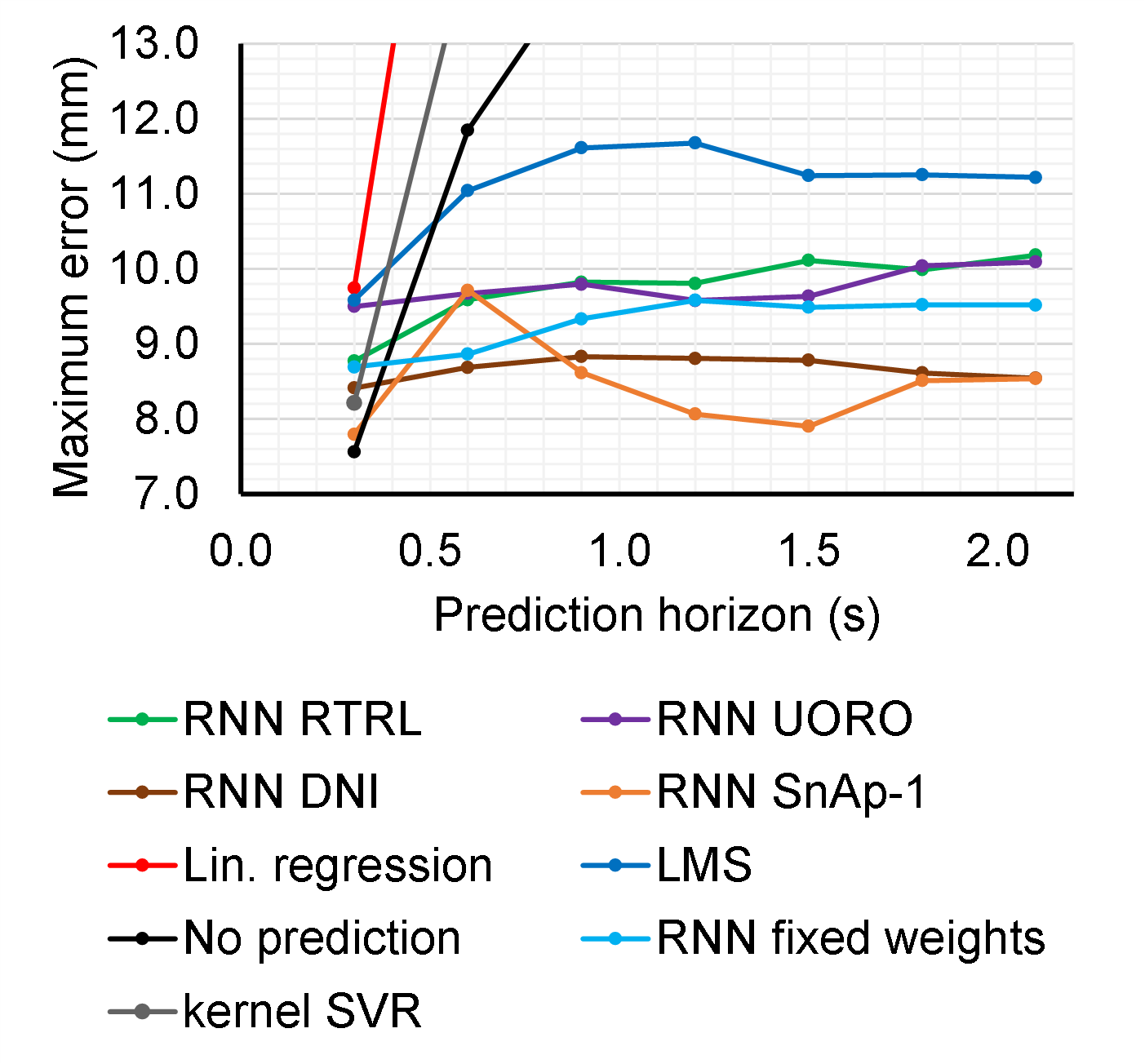} \label{fig:max error vs horizon 3.33Hz}}%
    \quad
    \subfloat[\normalsize Sampling at 10.0Hz]{\includegraphics[width=.30\textwidth]{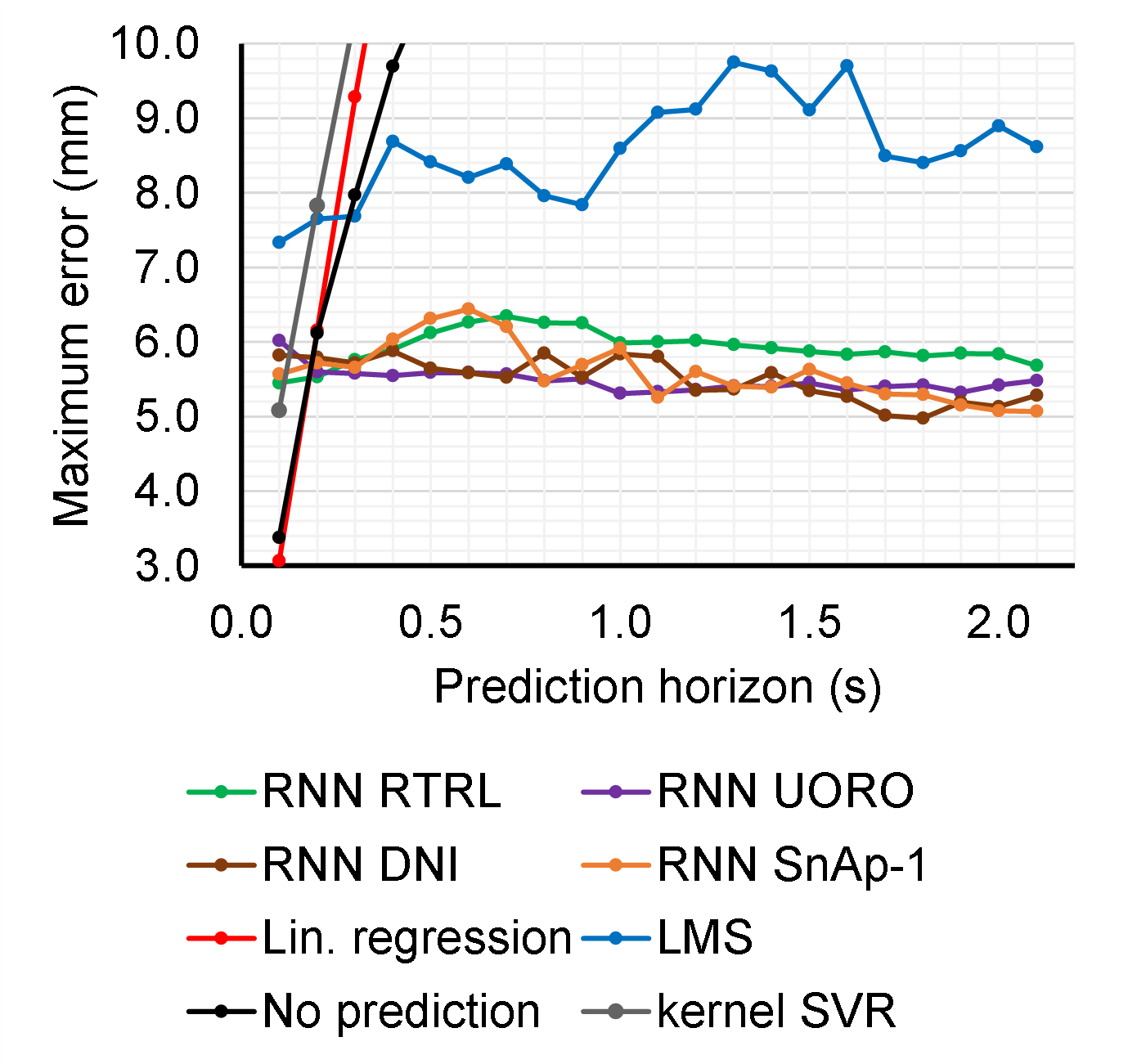} \label{fig:max error vs horizon 10Hz}}%
    \quad
    \subfloat[\normalsize Sampling at 30.0Hz]{\includegraphics[width=.30\textwidth]{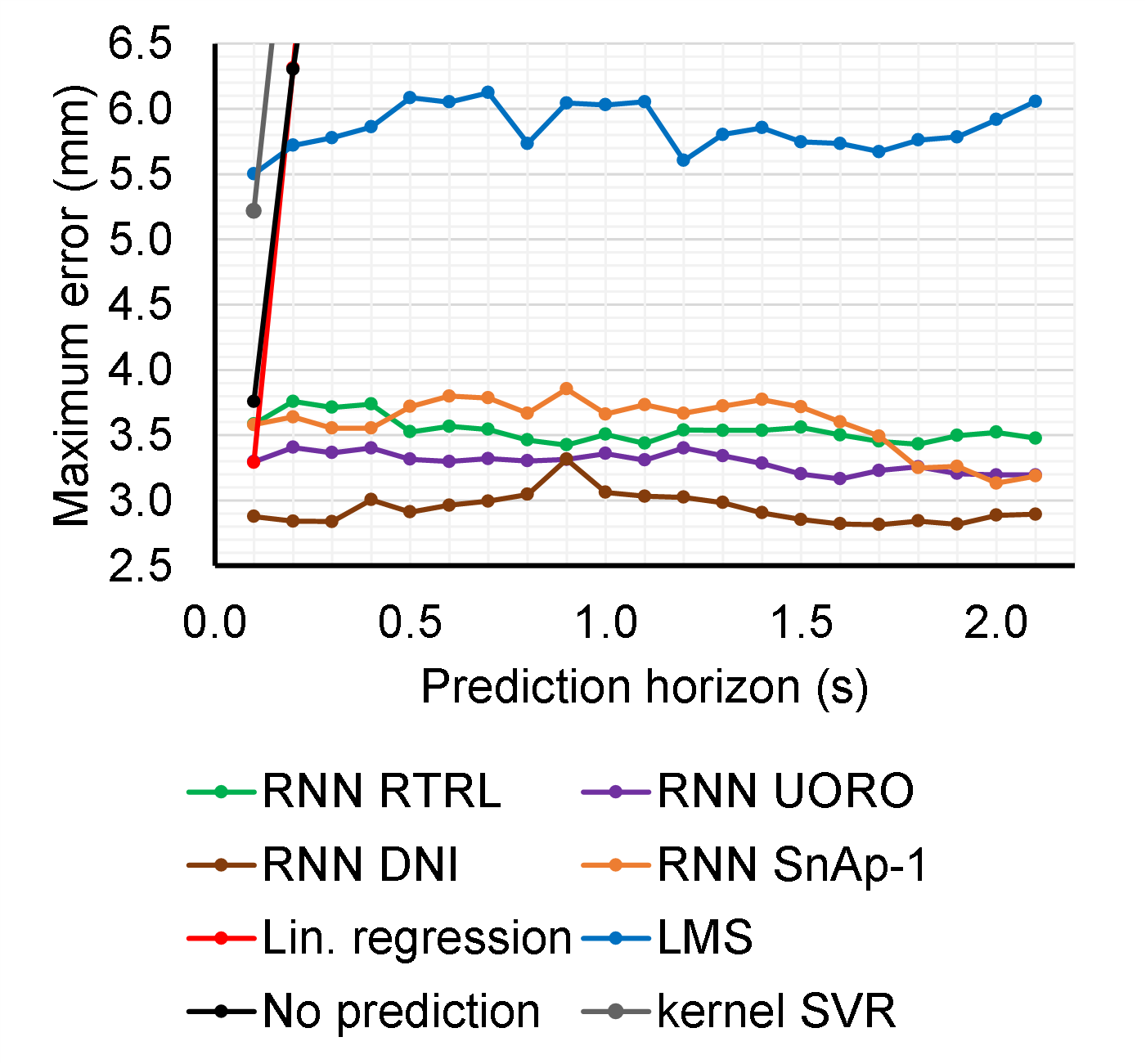} \label{fig:max error vs horizon 30Hz}}%
    \caption{Maximum error of each algorithm as a function of the forecasting horizon for different input signal sampling rates. Each point represents the average maximum error of the test set across the nine sequences for a given horizon using the best hyperparameters for that horizon (and each sequence individually)\protect\footnotemark.}%
    \label{fig:max_error_for_different_horizons}%
\end{figure*}

\footnotetext{Same as footnote \ref{footnote: high RNN and LMS errors} concerning the RNN with a frozen hidden layer. \label{footnote: high RNN errors}} 

\begin{figure*}[htb!]
    \centering
    \subfloat[\normalsize Sampling at 3.33Hz]{\includegraphics[width=.30\textwidth]{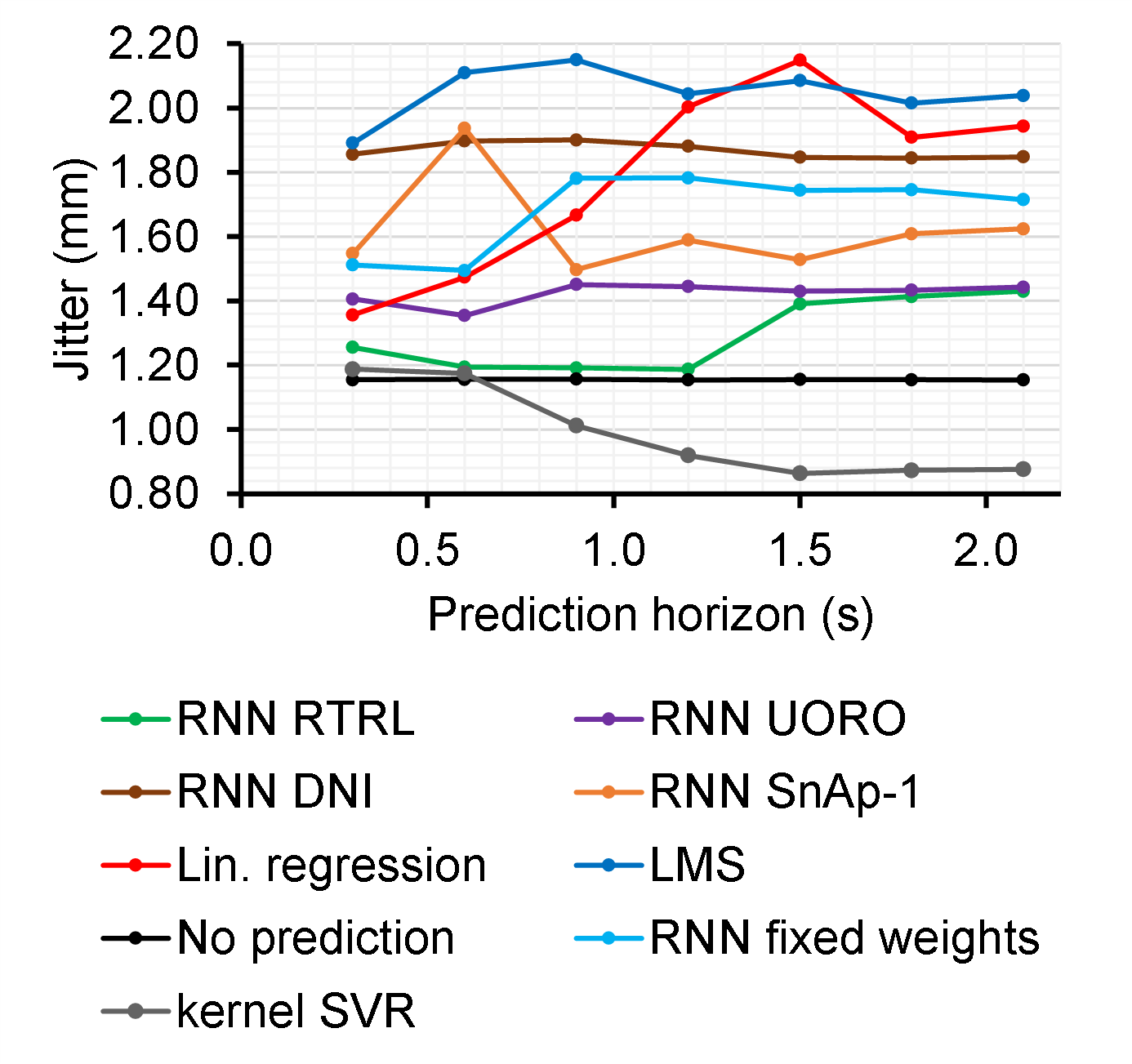} \label{fig:jitter vs horizon 3.33Hz}}%
    \quad
    \subfloat[\normalsize Sampling at 10.0Hz]{\includegraphics[width=.30\textwidth]{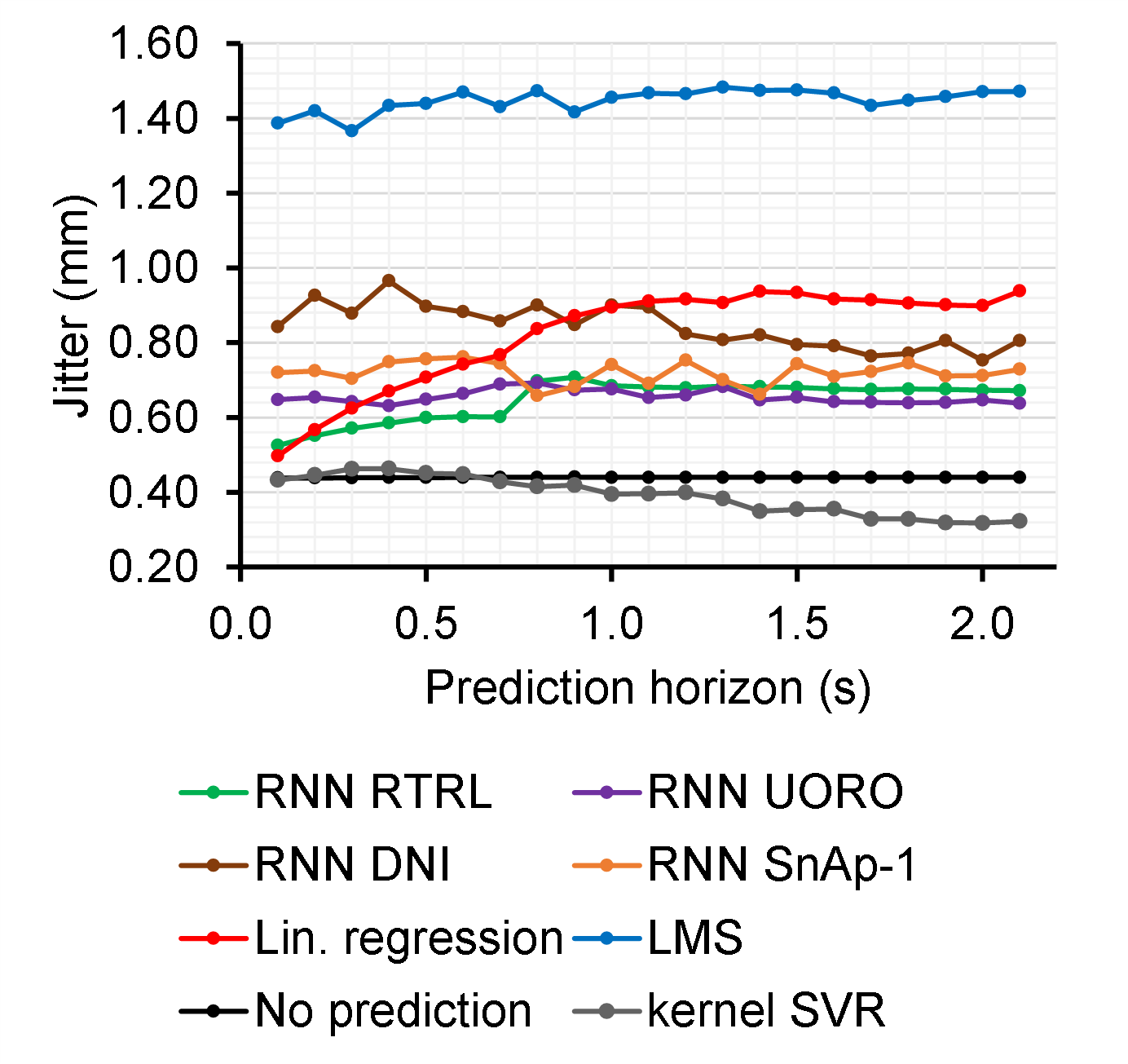} \label{fig:jitter vs horizon 10Hz}}%
    \quad
    \subfloat[\normalsize Sampling at 30.0Hz]{\includegraphics[width=.30\textwidth]{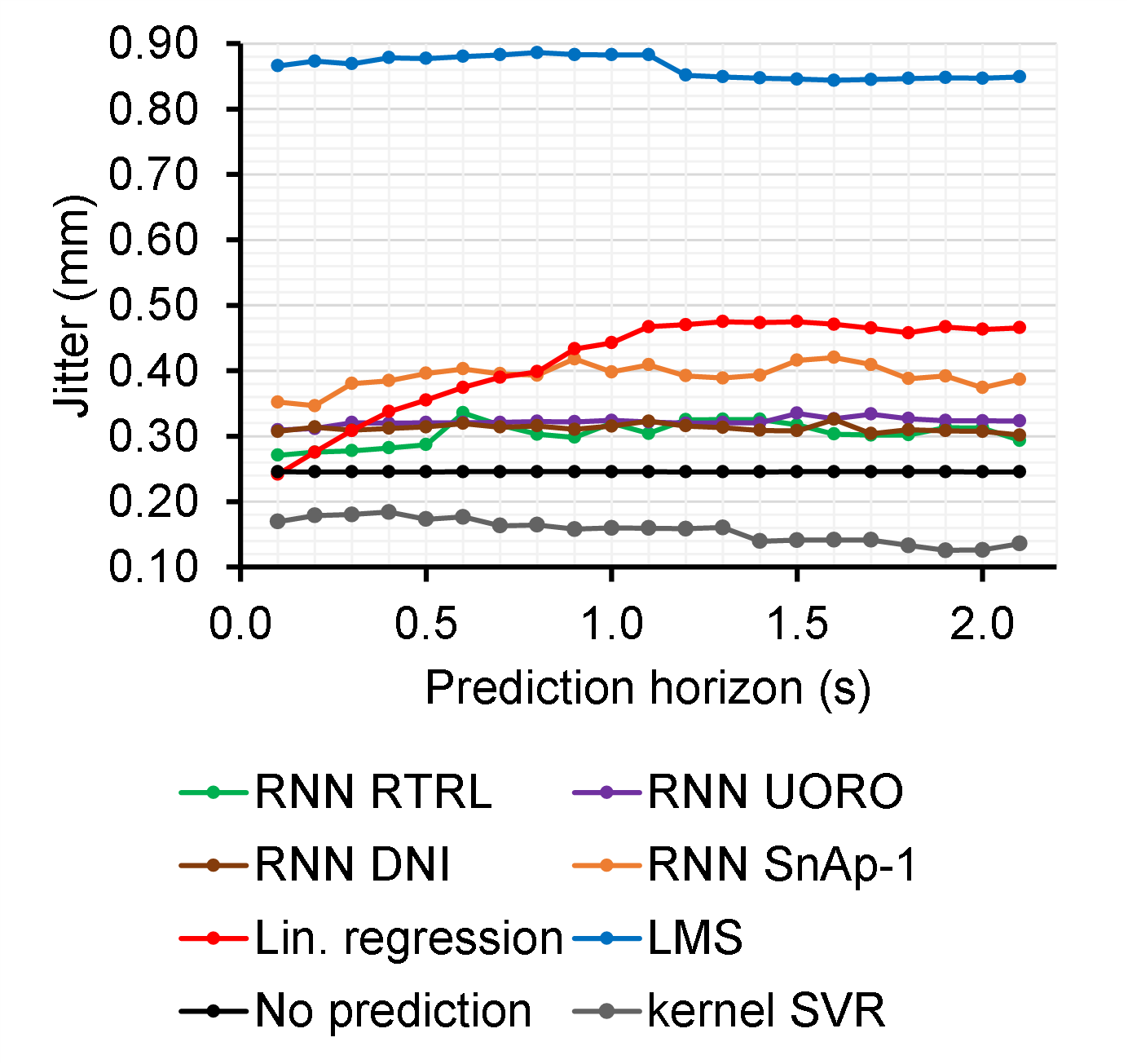} \label{fig:jitter vs horizon 30Hz}}%
    \caption{Jitter associated with each algorithm as a function of the forecasting horizon for different input signal sampling rates. Each point represents the average jitter of the test set across the nine sequences for a given horizon using the best hyperparameters for that horizon (and each sequence individually)\protect\footnotemark.}%
    \label{fig:jitter_for_different_horizons}%
\end{figure*}

\footnotetext{Same as footnote \ref{footnote: high RNN errors}.} 

The graphs characterizing forecasting performance (averaged over all the sequences) for each horizon value $h$ appear to have unsteady local variations (Figs. \ref{fig:MAE_for_different_horizons}, \ref{fig:RMSE_for_different_horizons}, \ref{fig:nRMSE_for_different_horizons}, \ref{fig:max_error_for_different_horizons}, and \ref{fig:jitter_for_different_horizons}). That is particularly visible in those corresponding to SnAp-1 at $f=3.33\text{Hz}$ and RTRL at $f=30\text{Hz}$. This instability is caused mainly by the following two factors. First, the set of hyperparameters automatically selected during cross-validation with grid search differs with each value of $h$. Secondly, there are relatively few respiratory traces in our dataset. The graphs displaying performance measures averaged over only the regular and irregular sequences exhibit even more instability with $h$, as the respiratory traces are fewer in each of these two subgroups (Fig. \ref{fig:regular vs irregular breathing} in Appendix \ref{appendix:regular vs irregular perf}). The accuracy of the RNNs and LMS averaged over all the records at $f=3.33\text{Hz}$ tended to decrease as $h$ increased, except for DNI, whose performance was relatively stable as $h$ varied. For instance, the nRMSEs associated with SnAp-1 at $h = 0.3\text{s}$ and $h = 2.1\text{s}$ were respectively equal to 0.294 and 0.334 (Fig. \ref{fig:nRMSE vs horizon 3.33Hz}). We could not observe such a trend at higher sampling frequencies, which may be due to the relatively small size of our dataset or the horizons considered, that may be low relative to $f$. That phenomenon may also be attributed to the inherent robustness of the RNN algorithms considered in our work.

Linear regression demonstrated high forecasting performance at short horizons. For instance, it was more effective than the other algorithms for all the metrics considered at $f=10\text{Hz}$ and $h = 0.1\text{s}$ (Figs. \ref{fig:MAE vs horizon 10Hz}, \ref{fig:RMSE vs horizon 10Hz}, \ref{fig:nRMSE vs horizon 10Hz}, \ref{fig:max error vs horizon 10Hz}, and \ref{fig:jitter vs horizon 10Hz}), with a corresponding RMSE and nRMSE equal to 0.442mm and 0.098, respectively. However, the RNNs had a higher accuracy at $f=30\text{Hz}$ and $h = 0.1\text{s}$ in terms of MAE, RMSE, and nRMSE. Nevertheless, for the latter frequency and horizon, linear regression still outperformed LMS regarding all metrics and had a lower maximum error and jitter than the RNNs, except for the maximum error of DNI (Figs. \ref{fig:MAE vs horizon 30Hz}, \ref{fig:RMSE vs horizon 30Hz}, \ref{fig:nRMSE vs horizon 30Hz}, \ref{fig:max error vs horizon 30Hz}, and \ref{fig:jitter vs horizon 30Hz}). We conjecture that it would perform similarly or better than the RNN algorithms for shorter response times at 30Hz (e.g., $h = 0.033\text{s}$ or $h = 0.066\text{s}$), given the strong decreasing trend of its associated errors as $h$ decreases. Using the last input as the predicted signal led to relatively high accuracy for low values of $h$, similar to linear regression. Nonetheless, the latter consistently resulted in lower errors for the shortest horizons considered\footnote{$h = 0.3\text{s}$ at 3.33Hz and $h = 0.1\text{s}$ at 10Hz and 30Hz}, except for the maximum error at $f=3.33\text{Hz}$ and $h = 0.3\text{s}$. In the latter setting, kernel SVR notably reached a lower average MAE, RMSE, and nRMSE than linear regression and the "no prediction" scenario without introducing much additional jitter compared to the latter. However, those three error metrics were still higher for SVR than for SnAp-1.

\begin{figure}[htb!]
	\centering
		\includegraphics[width=\columnwidth]{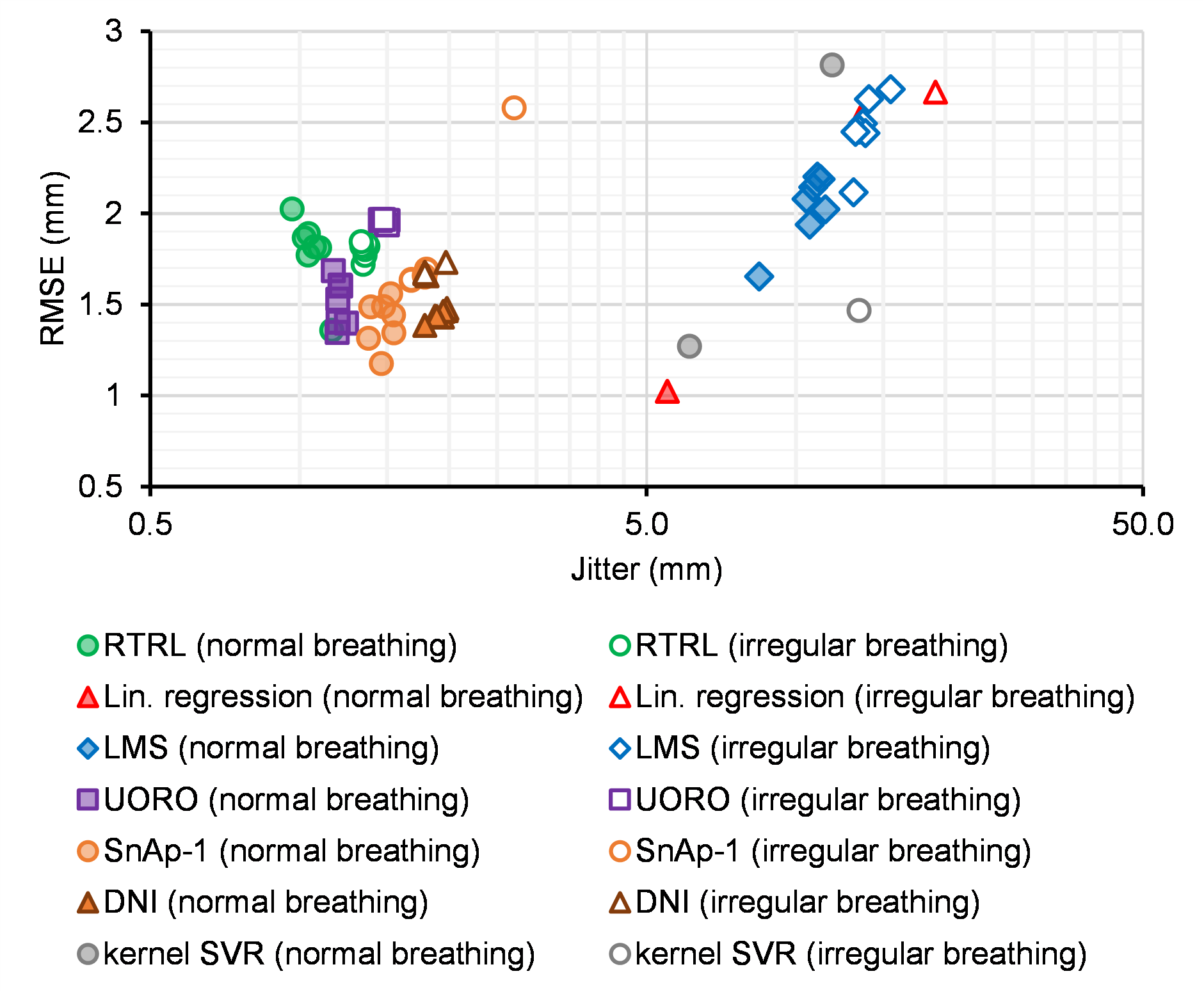}
	\caption{Average RMSE and jitter of the test set when the breathing signal is sampled at $f=3.33\text{Hz}$. Each point in the graph represents the mean of those two metrics over either the steady or irregular respiratory traces, for each algorithm and horizon $h$ considered, using the best hyperparameters for that value of $h$ and each record individually\protect\footnotemark. Data points associated with linear regression and kernel SVR forecasting at high response times were not displayed for readability as they correspond to high RMSEs.}
	\label{fig:RMSE vs jitter 3.33Hz}
\end{figure}

At $f=3.33\text{Hz}$, most metrics indicated lower performance with irregular motion, but this became less pronounced as $f$ increased (Table \ref{table:regular vs irregular breathing} in Appendix \ref{appendix:regular vs irregular perf}). For instance, UORO, SnAp-1, and DNI all had higher maximum errors and RMSEs for irregular breathing sequences at 3.33Hz and 10Hz, but that was not always true at 30Hz. At $f=3.33\text{Hz}$, the RMSE and maximum error averaged over the irregular breathing cases for each of those three algorithms were greater by approximately 25\% and 65\% than the same metrics averaged over the regular ones, respectively. In comparison, the respective increases at $f=10\text{Hz}$ were about 20\% and 44\%. The fact that RMSEs were higher for irregular respiratory records, in general, can also be observed in Fig. \ref{fig:RMSE vs jitter 3.33Hz}, illustrating the trade-off between maximizing accuracy and minimizing oscillations. Linear regression was less robust to unstable breathing at 3.33Hz than the other algorithms, as the corresponding RMSE and maximum error increased by 74\% and 96\%, respectively (Table \ref{table:relative error increases at 3.33Hz}). On the one hand, at $f=3.33\text{Hz}$ and $h=0.3\text{s}$, it achieved the lowest RMSE and maximum errors on average over the sequences with regular breathing patterns among the algorithms considered; these metrics were respectively equal to 1.02mm and 5.5mm (Fig. \ref{fig:regular vs irregular breathing} in Appendix \ref{appendix:regular vs irregular perf}). On the other hand, in that same setting, it performed the worst in terms of these two errors averaged over the records with irregular breathing patterns, as they reached 2.67mm and 19.1mm, respectively. Fig. \ref{fig:coordz_marker3_seq1} shows one instance of prediction with linear regression and kernel SVR of an unsteady breathing signal, where both algorithms mostly underestimated the z-coordinate throughout the test set. Noticeably, kernel SVR reached the lowest test RMSE averaged over the irregular breathing records at $f=3.33\text{Hz}$ and $h=0.3\text{s}$, equal to 1.467mm (Fig. \ref{fig:regular vs irregular breathing}).

\begin{table}[htb!]
\centering
\begin{tabular}{lll}
\hline
                  & RMSE          & Maximum error           \\
                  & increase      & increase                \\   
\hline               
\hline
RTRL              & 0.56\%        & 48.3\%                  \\
UORO              & 29.7\%        & 82.1\%                  \\
SnAp-1            & 27.8\%        & 62.2\%                  \\
DNI               & 16.4\%        & 51.5\%                  \\
LMS               & 21.4\%        & 30.3\%                  \\
Linear regression & 73.9\%        & 96.2\%                  \\
Kernel SVR        & 23.0\%        & 44.8\%                  \\            
\hline
\end{tabular}
\caption{Relative increase in RMSE and maximum error at $f=3.33\text{Hz}$ for each algorithm, calculated as the difference between errors averaged separately over irregular and regular breathing sequences, across all considered horizons (i.e., the values in Table \ref{table:regular vs irregular breathing} in Appendix \ref{appendix:regular vs irregular perf}).}
\label{table:relative error increases at 3.33Hz}
\end{table}

\footnotetext{Sequence 7, denoted as 201205111057-LACLARUAR-3-O-72 in \cite{krilavicius2016predicting}, was excluded from the records with irregular respiratory motion when reporting performance metrics associated with unsteady breathing, as it corresponds to slow motion and does not feature sudden or abrupt displacements that could make prediction particularly hard, yet its signal amplitude is high (leading to potentially high errors) and the time points in the test set are relatively few compared to the other sequences, due to its shorter duration. \label{footnote: slow motion sequence removed}}

\subsection{Influence of the hyperparameters on prediction accuracy}
\label{section:discussion hyperparameters influence} 

\begin{figure*}[ht!]%
    \centering
    \subfloat[UORO - 3.33Hz input sampling]{\includegraphics[width=.31\textwidth]{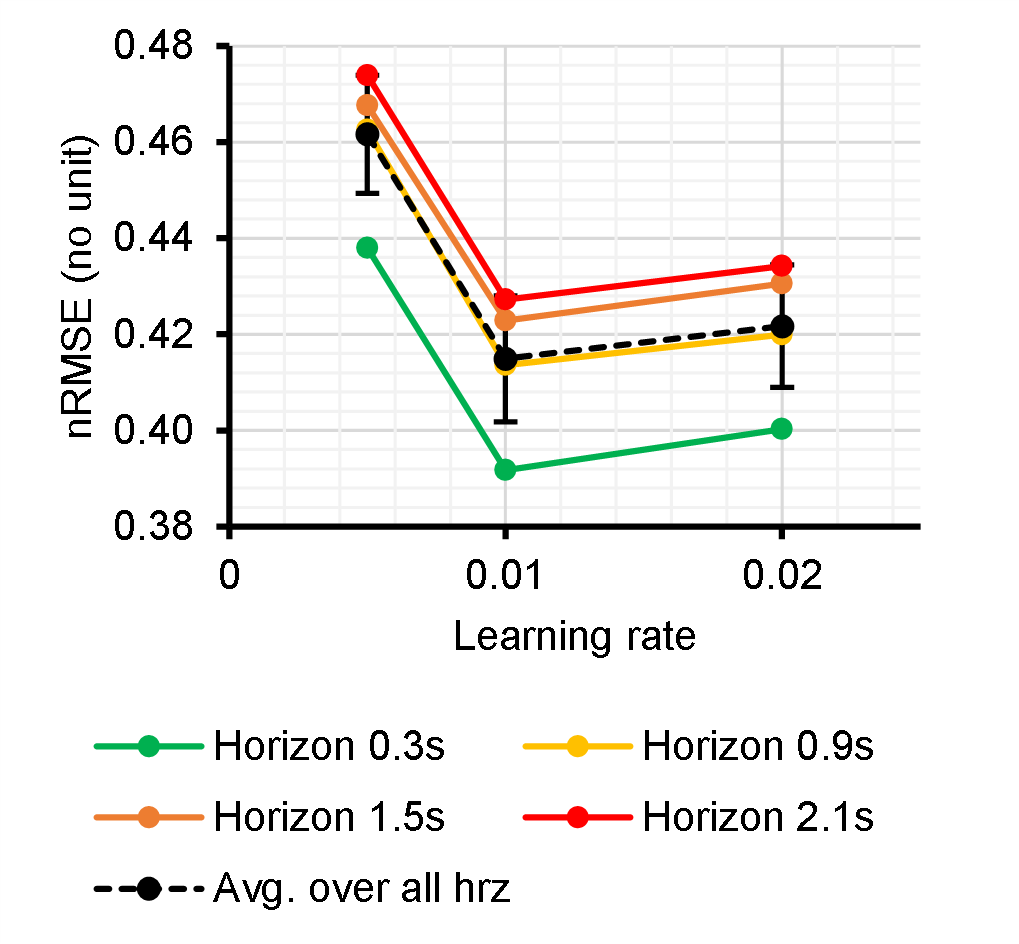} \label{fig:influence_of_learning_rate_UORO_3.33Hz}}
    \quad
    \subfloat[UORO - 10Hz input sampling]{\includegraphics[width=.31\textwidth]{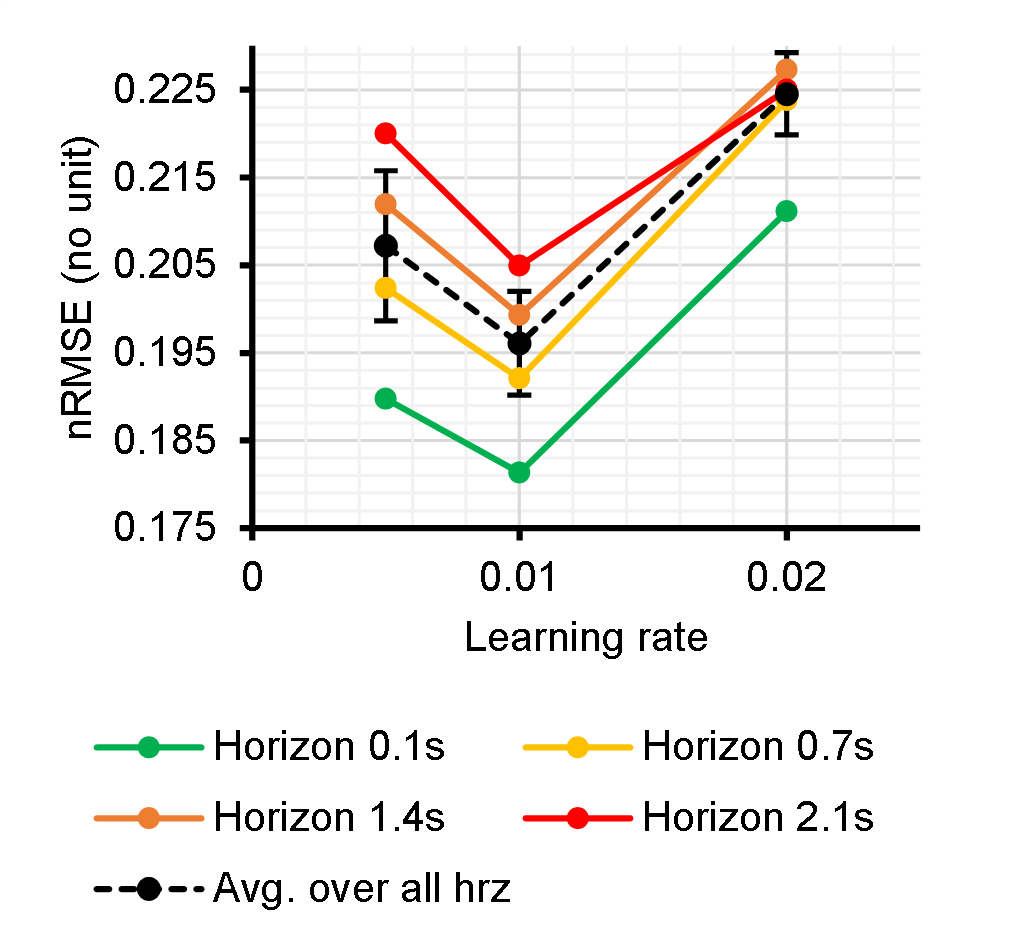} \label{fig:influence_of_learning_rate_UORO_10Hz}}
    \quad
    \subfloat[UORO - 30Hz input sampling]{\includegraphics[width=.31\textwidth]{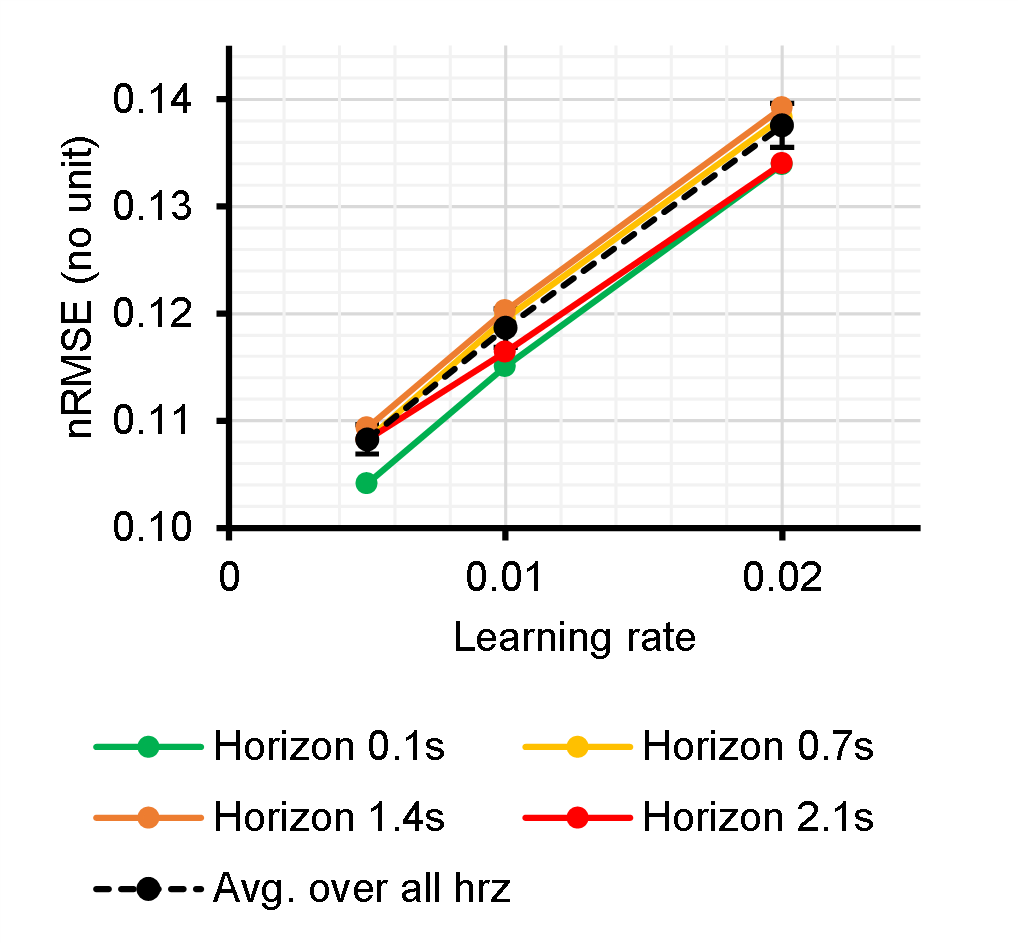} \label{fig:influence_of_learning_rate_UORO_30Hz}}
    \quad
    \subfloat[SnAp-1 - 3.33Hz input sampling]{\includegraphics[width=.31\textwidth]{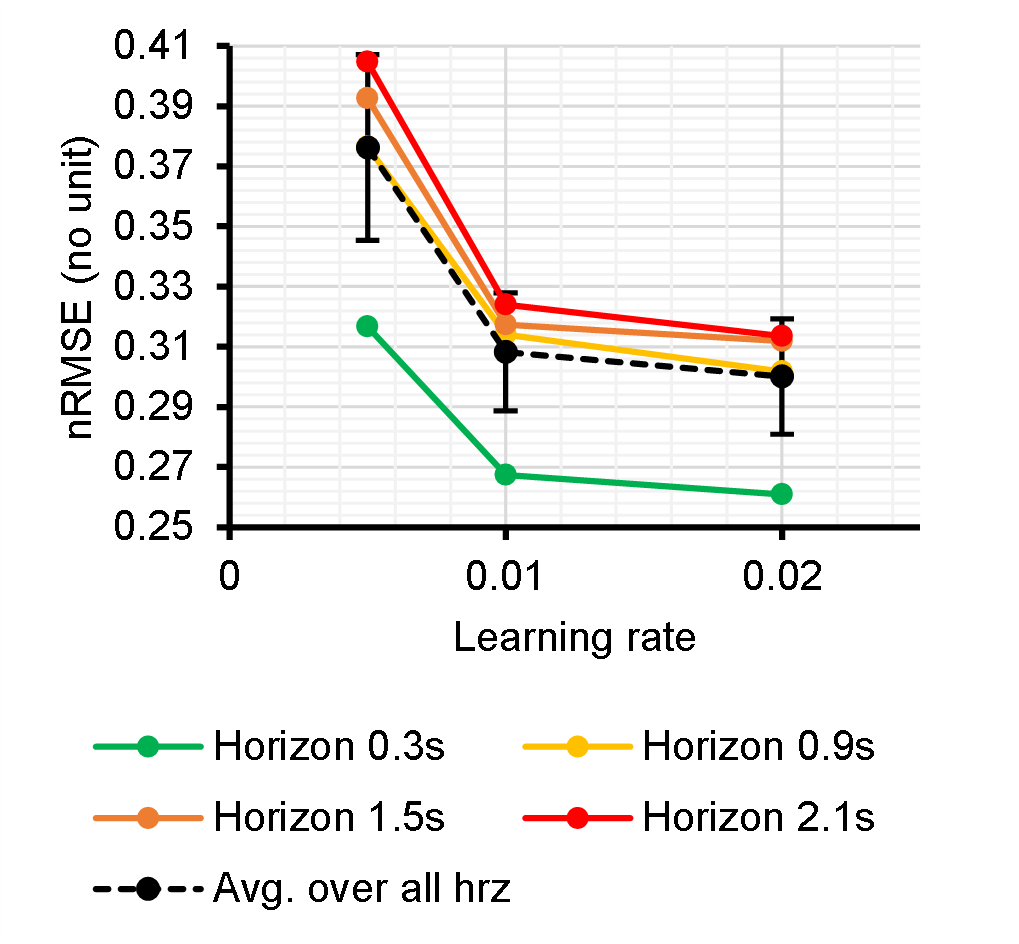} \label{fig:influence_of_learning_rate_SnAp-1_3.33Hz}}
    \quad
    \subfloat[SnAp-1 - 10Hz input sampling]{\includegraphics[width=.31\textwidth]{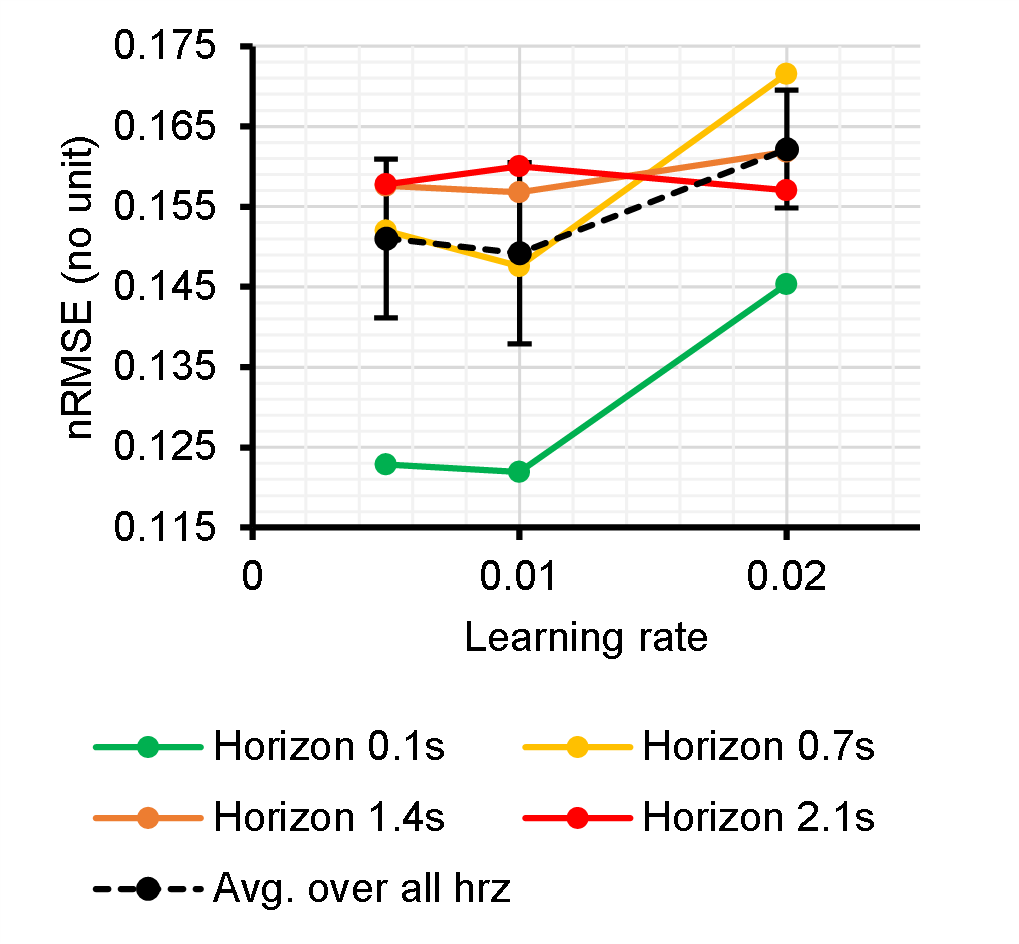} \label{fig:influence_of_learning_rate_SnAp-1_10Hz}}
    \quad
    \subfloat[SnAp-1 - 30Hz input sampling]{\includegraphics[width=.31\textwidth]{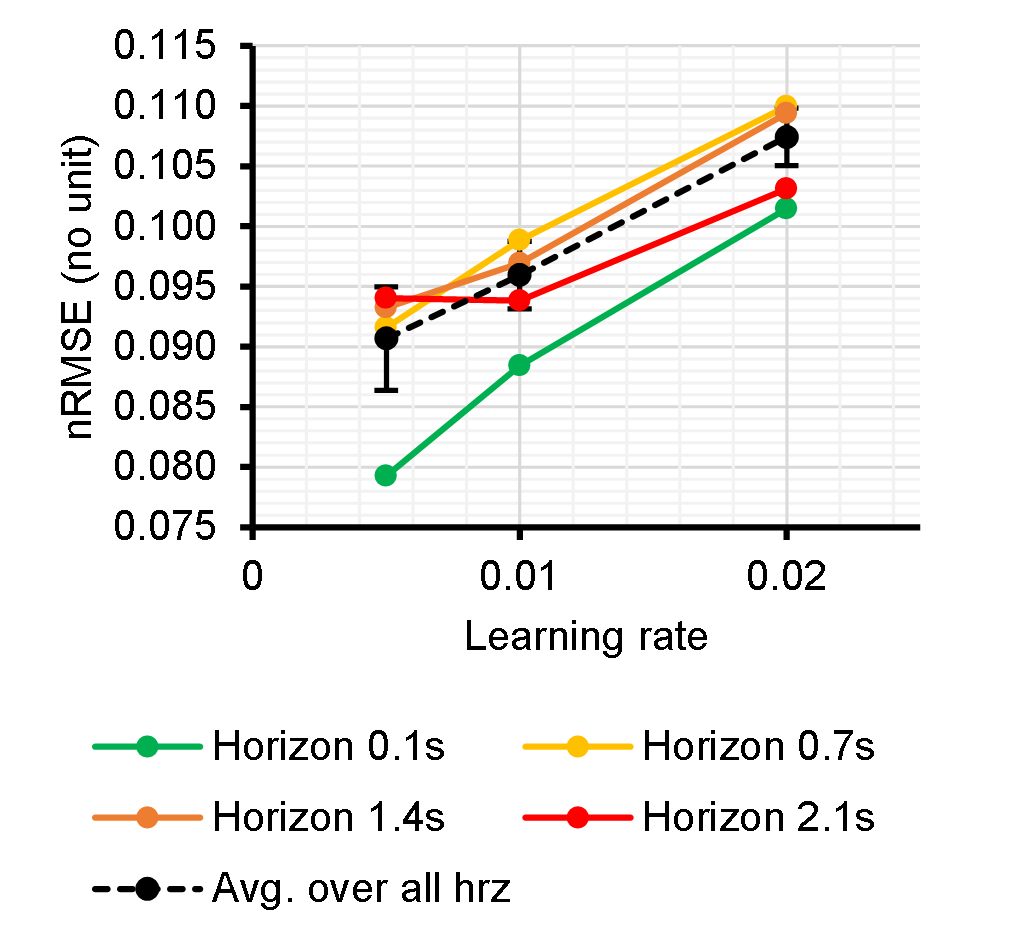} \label{fig:influence_of_learning_rate_SnAp-1_30Hz}}
    \quad
    \subfloat[DNI - 3.33Hz input sampling]{\includegraphics[width=.31\textwidth]{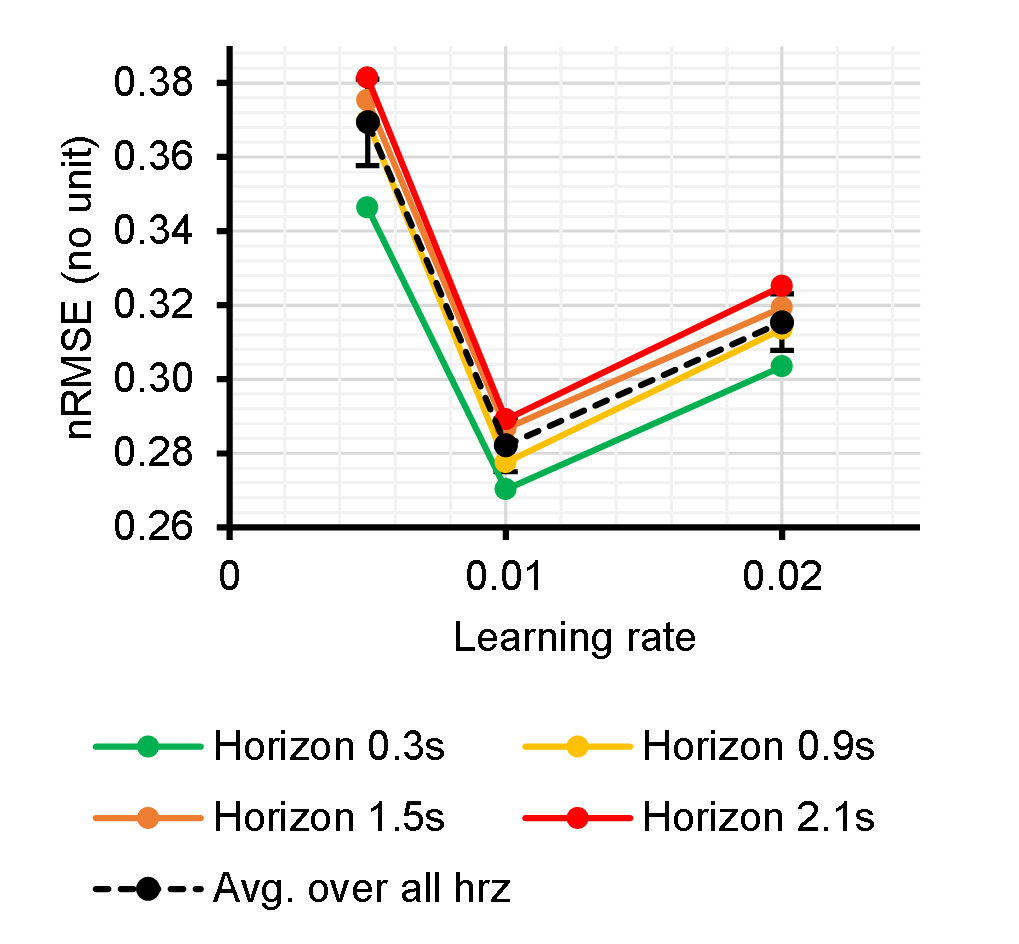} \label{fig:influence_of_learning_rate_DNI_3.33Hz}}
    \quad
    \subfloat[DNI - 10Hz input sampling]{\includegraphics[width=.31\textwidth]{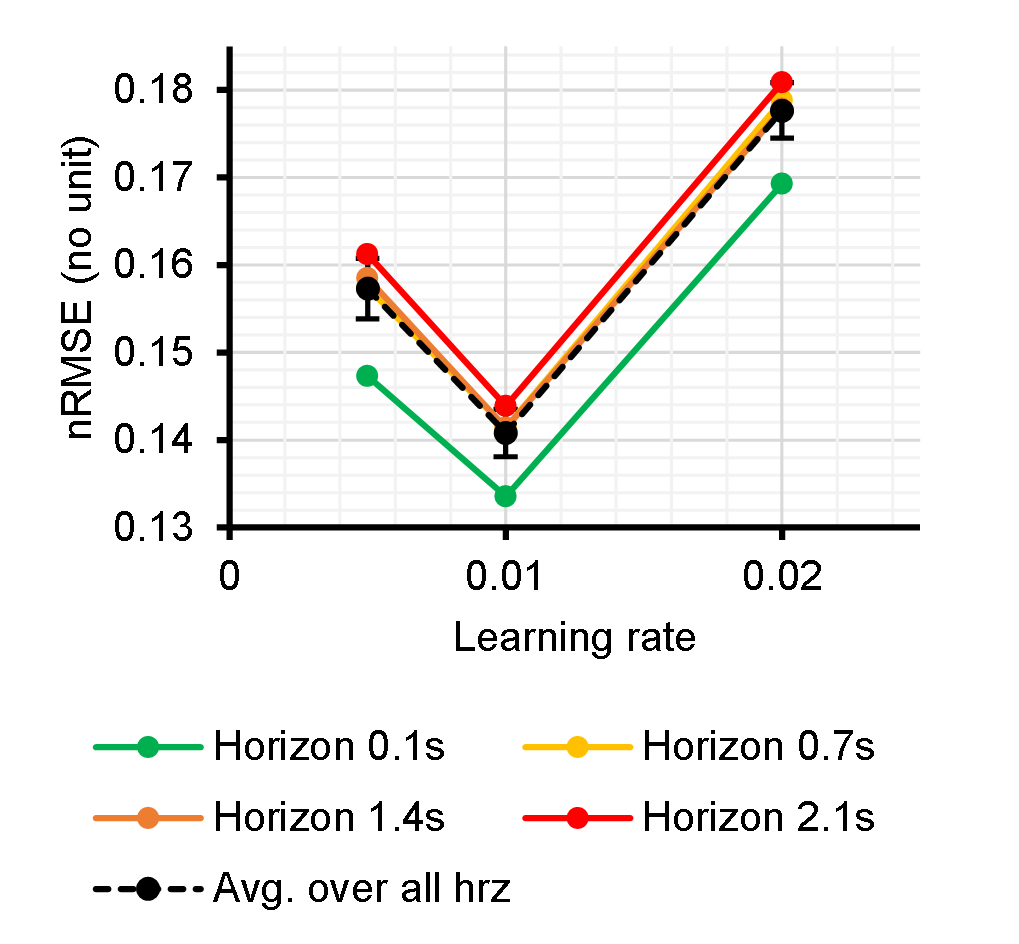} \label{fig:influence_of_learning_rate_DNI_10Hz}}
    \quad
    \subfloat[DNI - 30Hz input sampling]{\includegraphics[width=.31\textwidth]{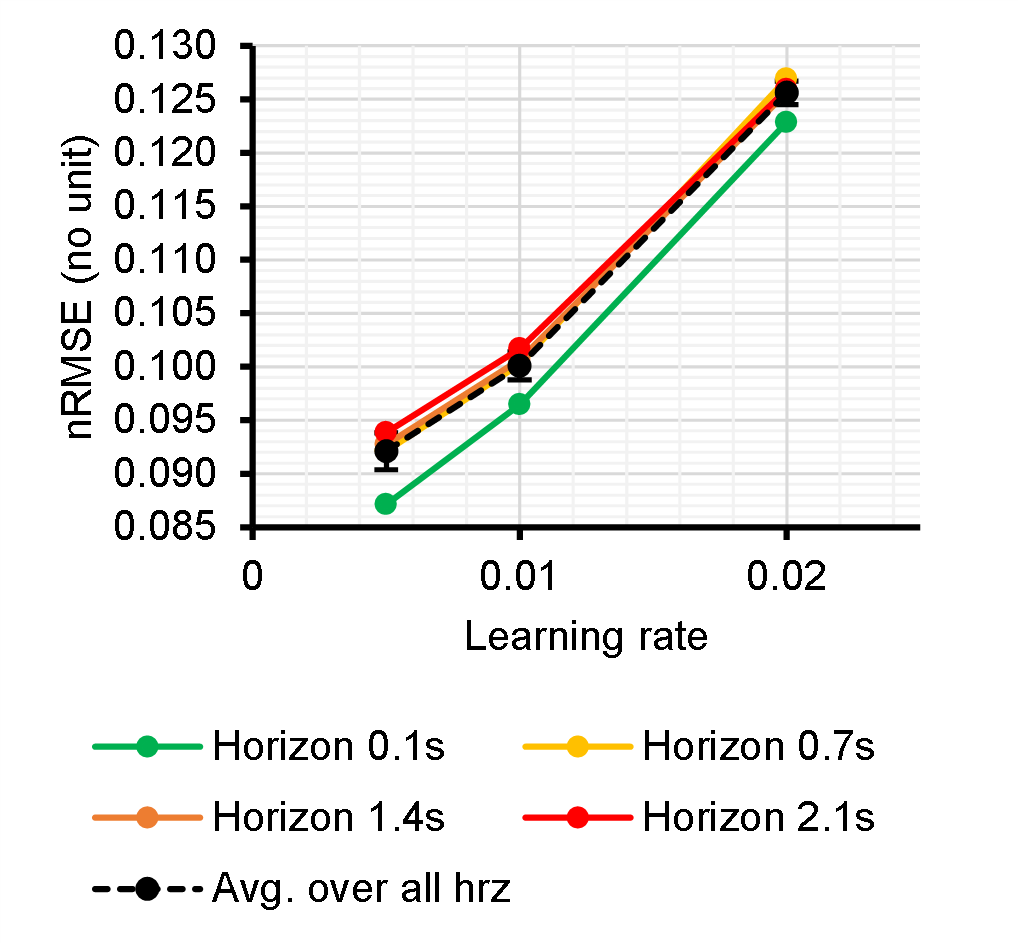} \label{fig:influence_of_learning_rate_DNI_30Hz}}
    \quad
    \caption{Forecasting nRMSE of UORO, SnAp-1, and DNI of the cross-validation set as a function of the learning rate $\eta$, for various response times $h$ and input signal sampling frequencies $f$. For each sequence and specific values of $\eta$ and $h$, we compute the nRMSE minimum over every possible combination of $q$ and $L$ within the cross-validation range (Table \ref{table:models comparison}); all errors in that grid are averaged over 50 runs to mitigate RNN stochasticity. Each colored point represents the average of these minimum errors over the nine records. The black dotted curves show the nRMSE minimum, averaged over both the nine respiratory traces and the response times considered, between 0.1s and 2.1s, or between 0.3s and 2.1s if $f=3.33\text{Hz}$. Error bars indicate its standard deviation over these values of $h$.}
    \label{fig:learning rate influence}
\end{figure*}

\begin{figure*}[ht!]%
    \centering
    \subfloat[UORO - 3.33Hz input sampling]{\includegraphics[width=.31\textwidth]{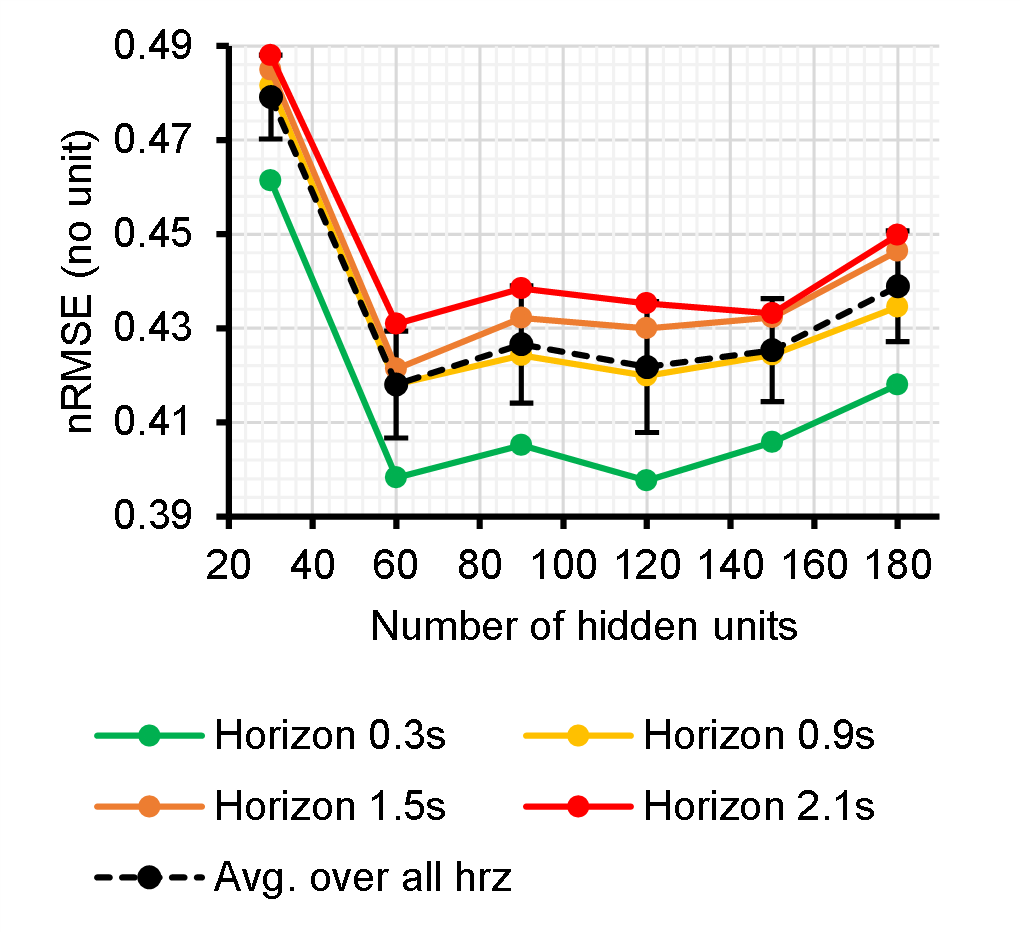} \label{fig:influence_of_n_hidden_units_UORO_3.33Hz}}
    \quad
    \subfloat[UORO - 10Hz input sampling]{\includegraphics[width=.31\textwidth]{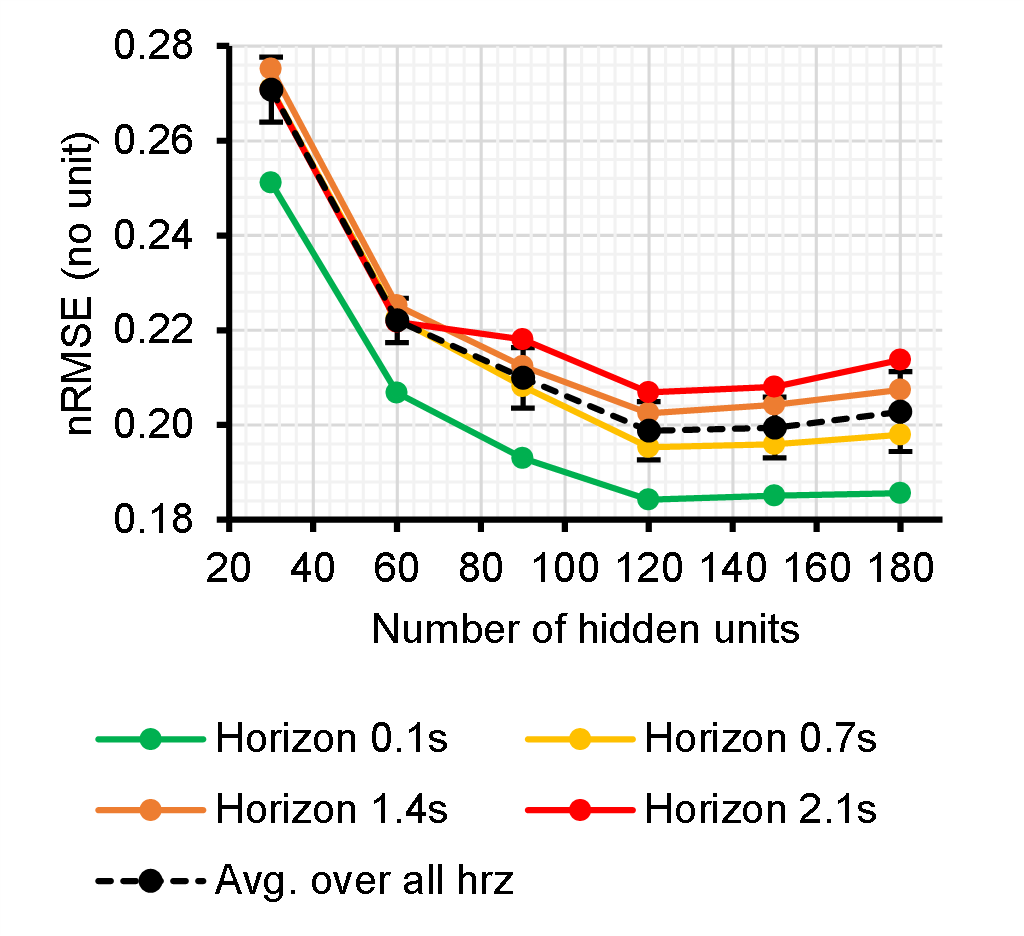} \label{fig:influence_of_n_hidden_units_UORO_10Hz}}
    \quad
    \subfloat[UORO - 30Hz input sampling]{\includegraphics[width=.31\textwidth]{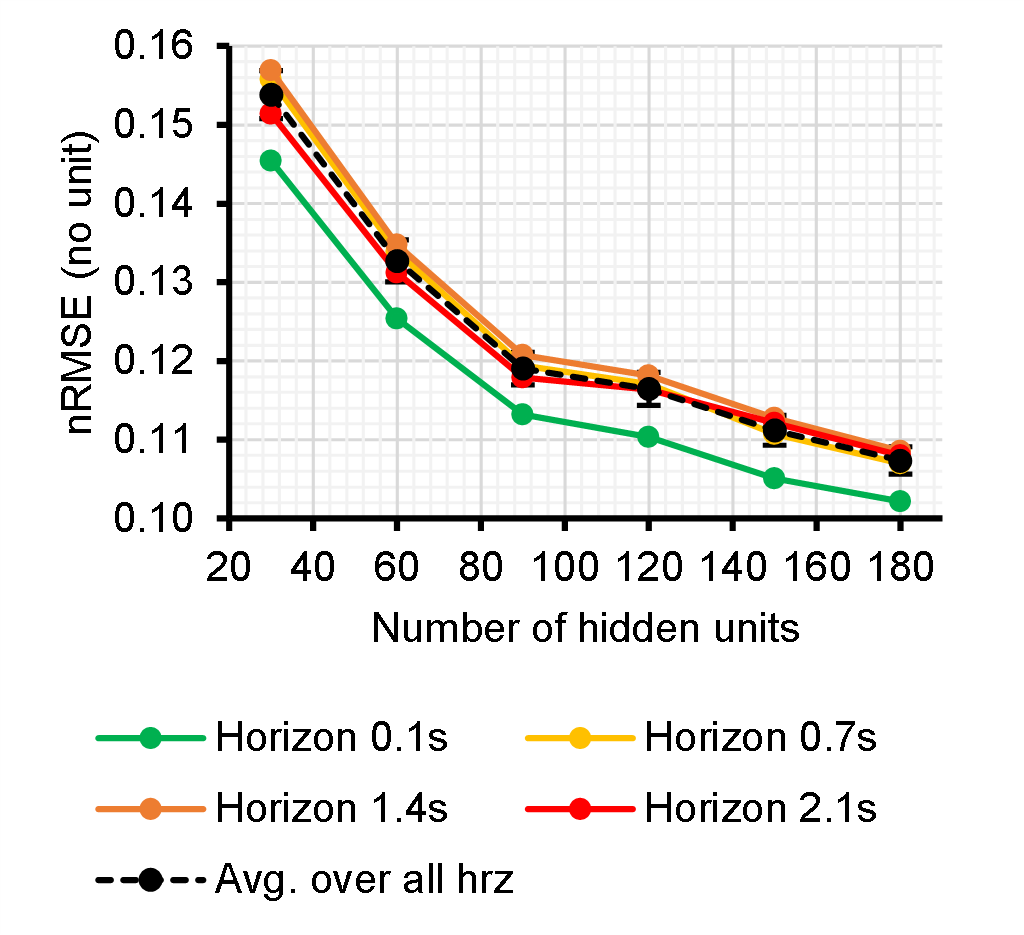} \label{fig:influence_of_n_hidden_units_UORO_30Hz}}
    \quad
    \subfloat[SnAp-1 - 3.33Hz input sampling]{\includegraphics[width=.31\textwidth]{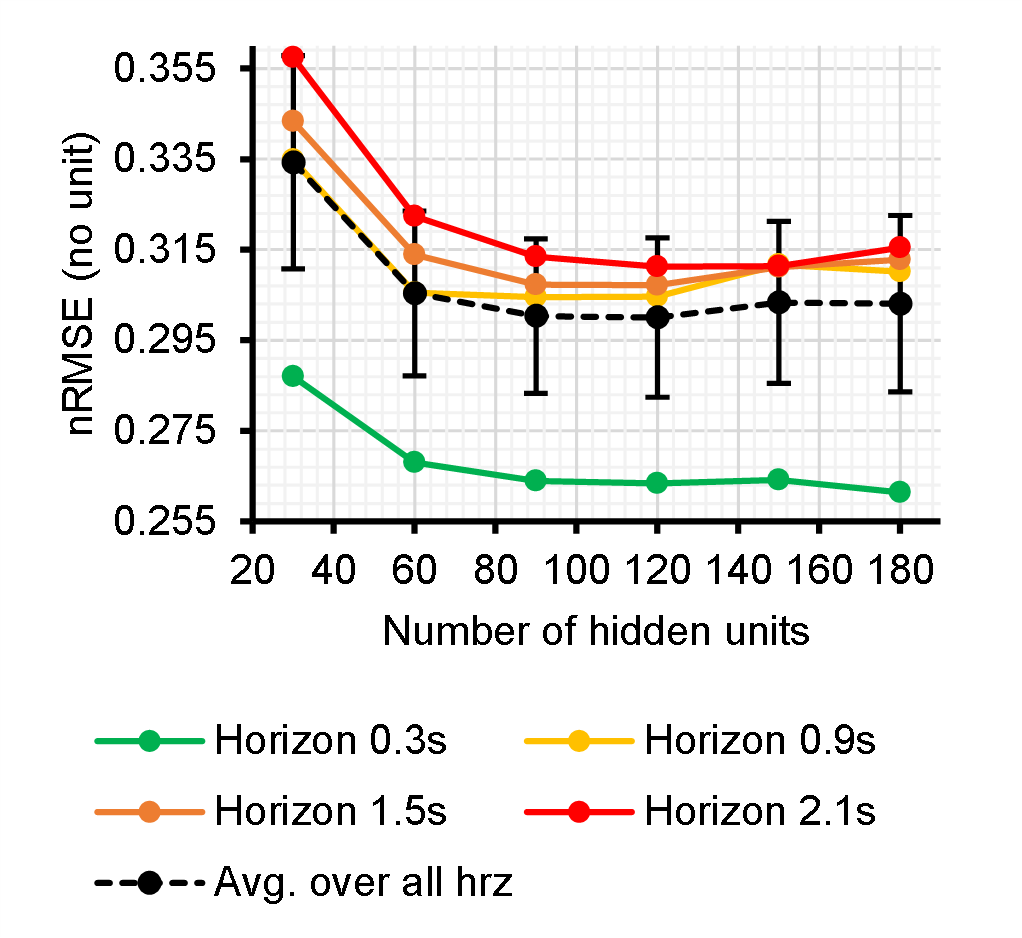} \label{fig:influence_of_n_hidden_units_SnAp-1_3.33Hz}}
    \quad
    \subfloat[SnAp-1 - 10Hz input sampling]{\includegraphics[width=.31\textwidth]{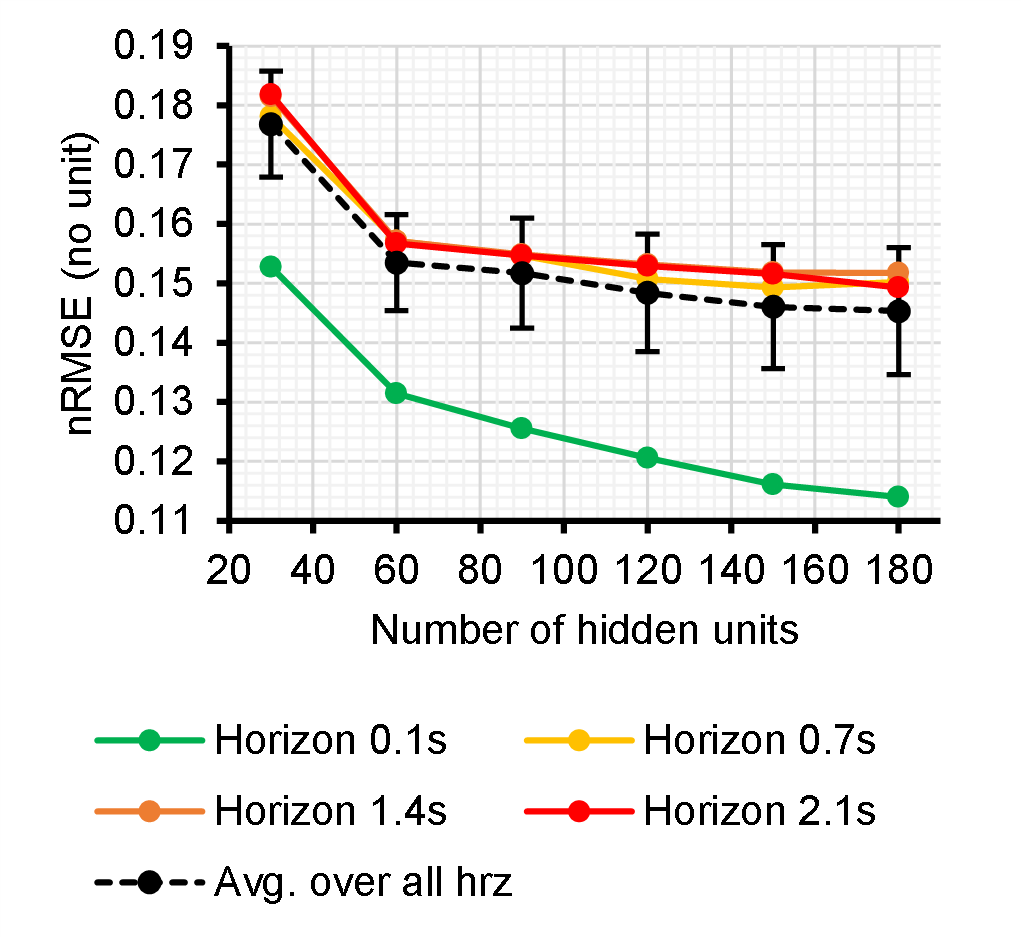} \label{fig:influence_of_n_hidden_units_SnAp-1_10Hz}}
    \quad
    \subfloat[SnAp-1 - 30Hz input sampling]{\includegraphics[width=.31\textwidth]{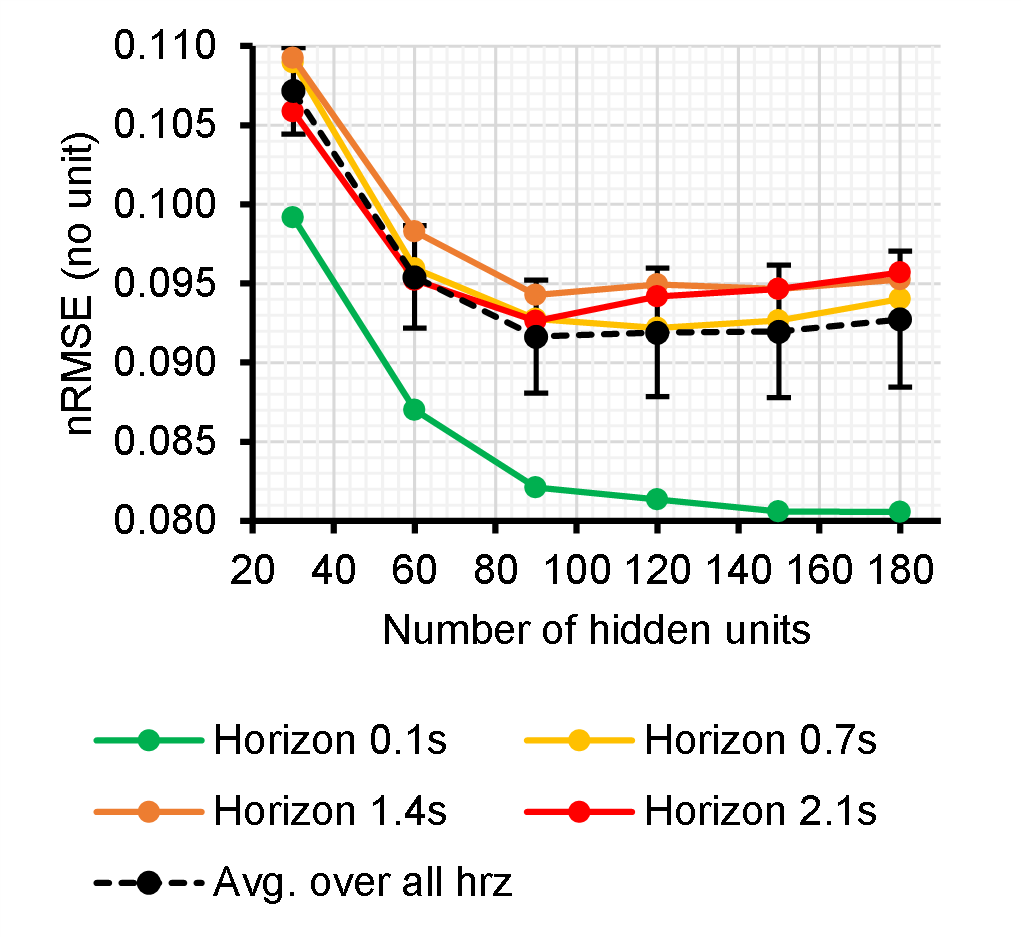} \label{fig:influence_of_n_hidden_units_SnAp-1_30Hz}}
    \quad
    \subfloat[DNI - 3.33Hz input sampling]{\includegraphics[width=.31\textwidth]{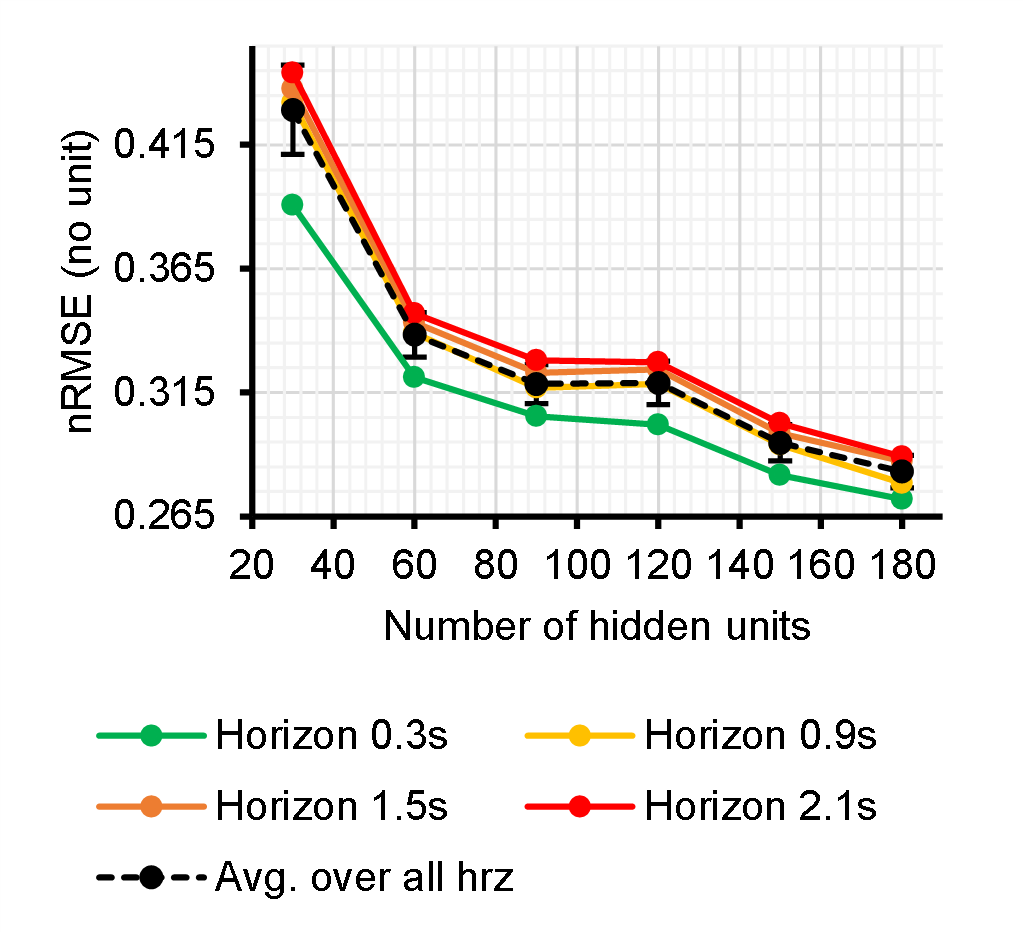} \label{fig:influence_of_n_hidden_units_DNI_3.33Hz}}
    \quad
    \subfloat[DNI - 10Hz input sampling]{\includegraphics[width=.31\textwidth]{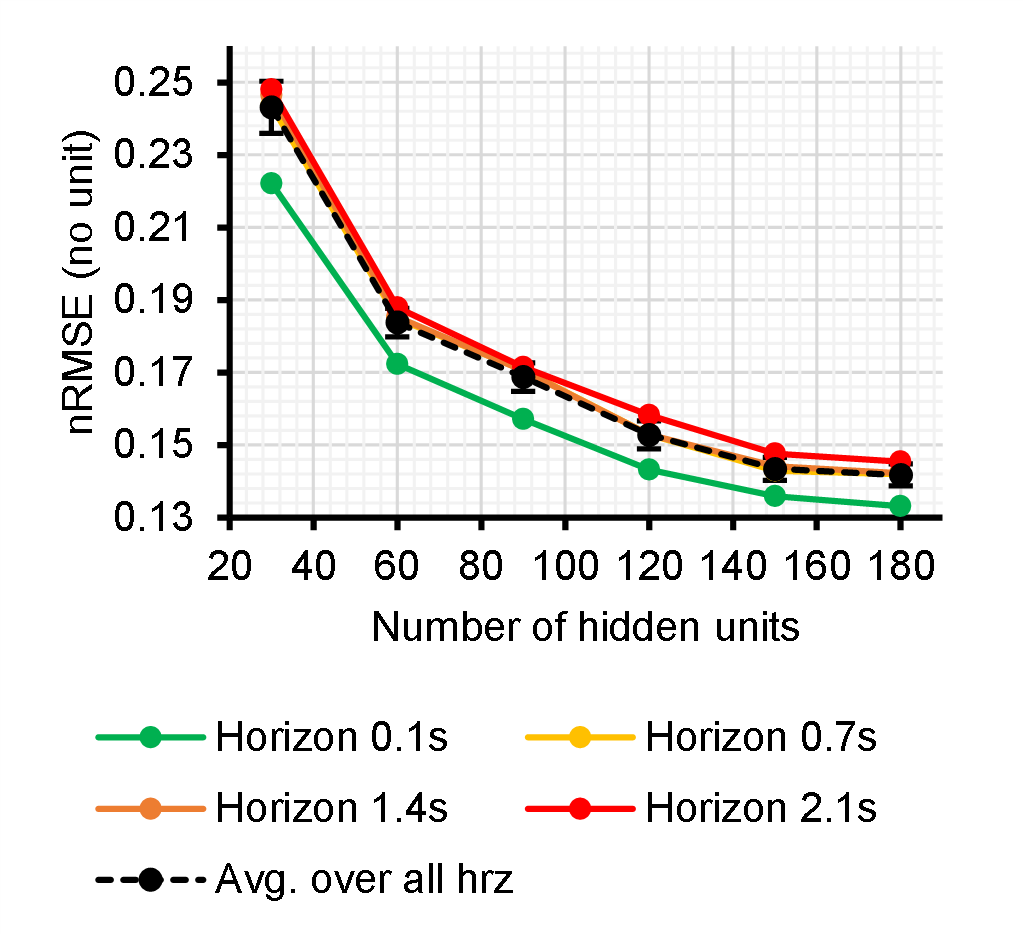} \label{fig:influence_of_n_hidden_units_DNI_10Hz}}
    \quad
    \subfloat[DNI - 30Hz input sampling]{\includegraphics[width=.31\textwidth]{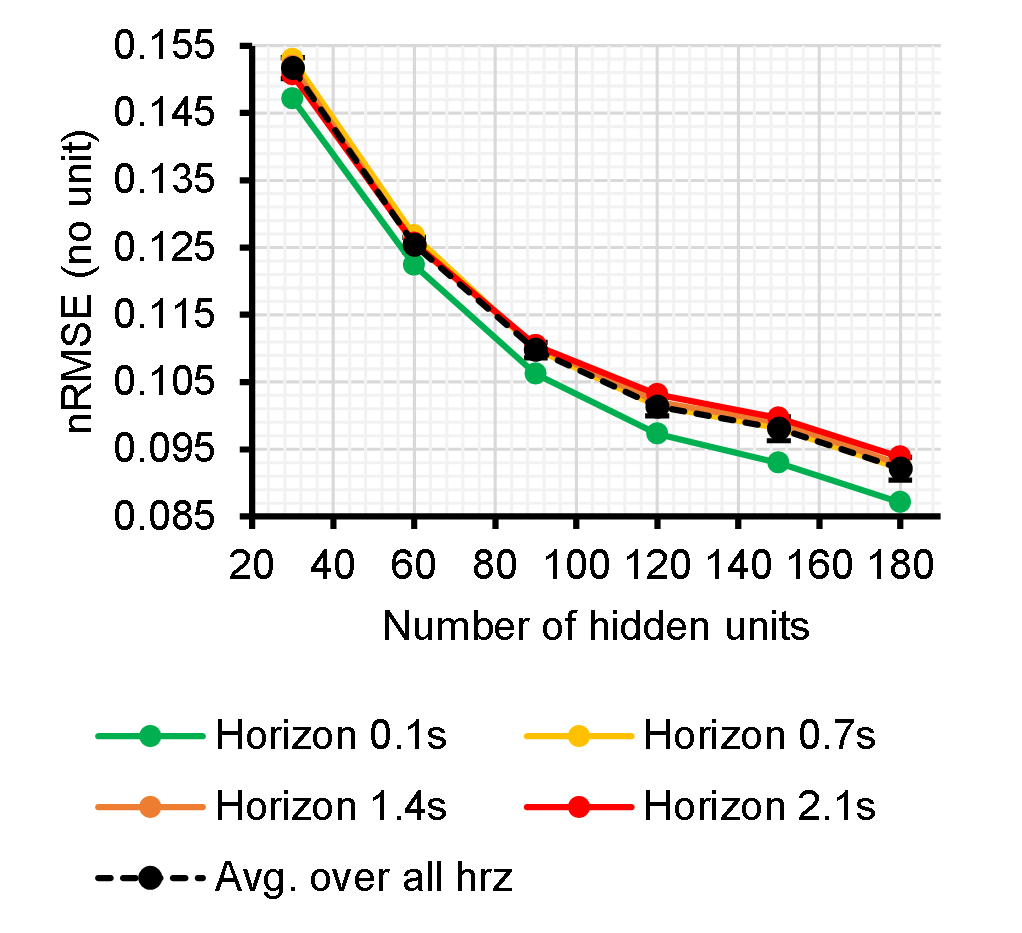} \label{fig:influence_of_n_hidden_units_DNI_30Hz}}
    \quad
    \caption{Forecasting nRMSE of UORO, SnAp-1, and DNI of the cross-validation set as a function of the number of hidden units $q$, for various response times $h$ and input signal sampling frequencies $f$. For each sequence and specific values of $q$ and $h$, we compute the nRMSE minimum over every possible combination of $\eta$ and $L$ within the cross-validation range (Table \ref{table:models comparison}); all errors in that grid are averaged over 50 runs to mitigate RNN stochasticity. Each colored point represents the average of these minimum errors over the nine records. The black dotted curves show the nRMSE minimum, averaged over both the nine respiratory traces and the response times considered, between 0.1s and 2.1s, or between 0.3s and 2.1s if $f=3.33\text{Hz}$. Error bars indicate its standard deviation over these values of $h$.}
    \label{fig:influence of the nb of hidden units}
\end{figure*}

\begin{figure*}[ht!]%
    \centering
    \subfloat[UORO - 3.33Hz input sampling]{\includegraphics[width=.31\textwidth]{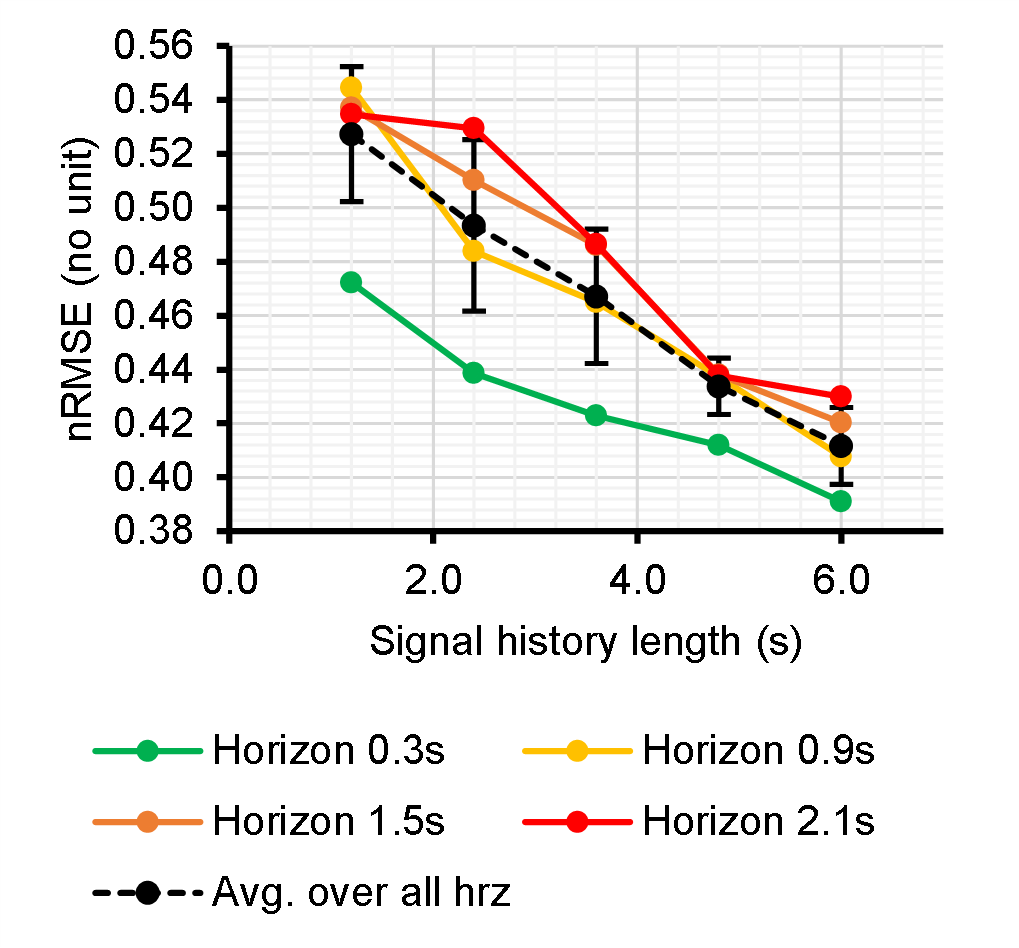} \label{fig:influence_of_SHL_UORO_3.33Hz}}
    \quad
    \subfloat[UORO - 10Hz input sampling]{\includegraphics[width=.31\textwidth]{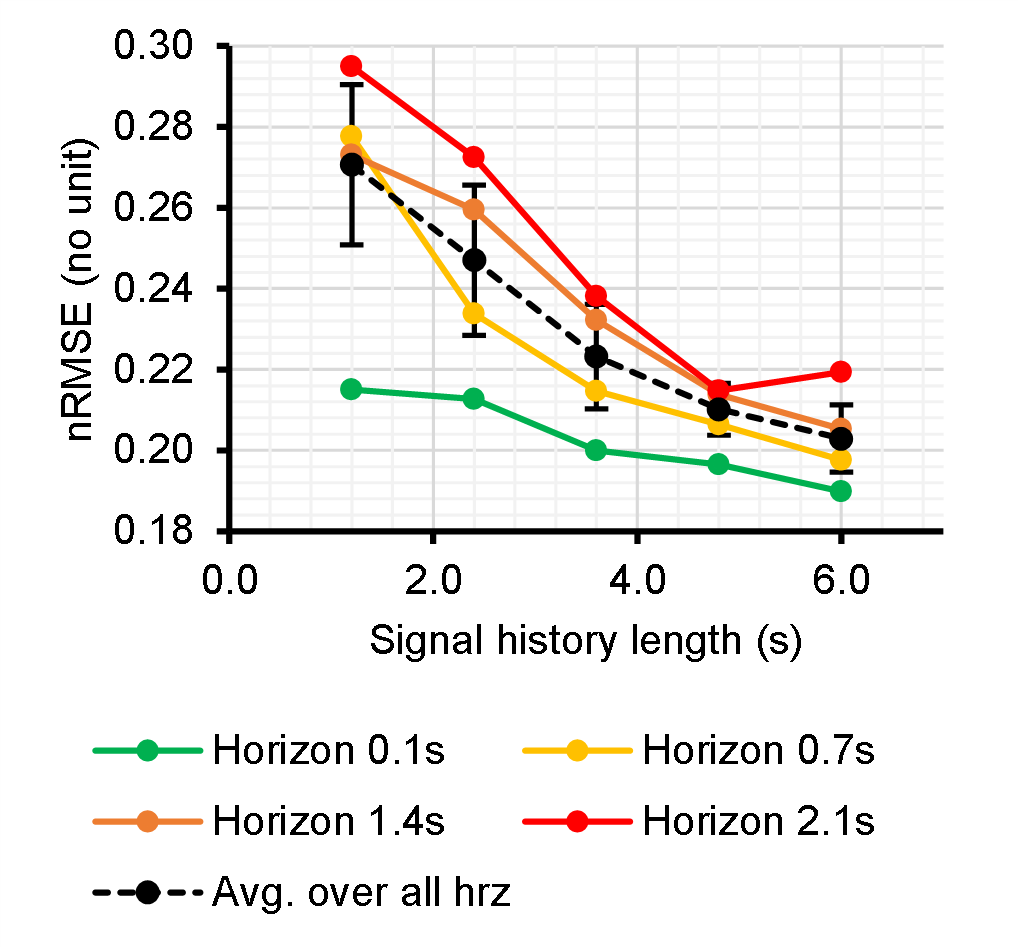} \label{fig:influence_of_SHL_UORO_10Hz}}
    \quad
    \subfloat[UORO - 30Hz input sampling]{\includegraphics[width=.31\textwidth]{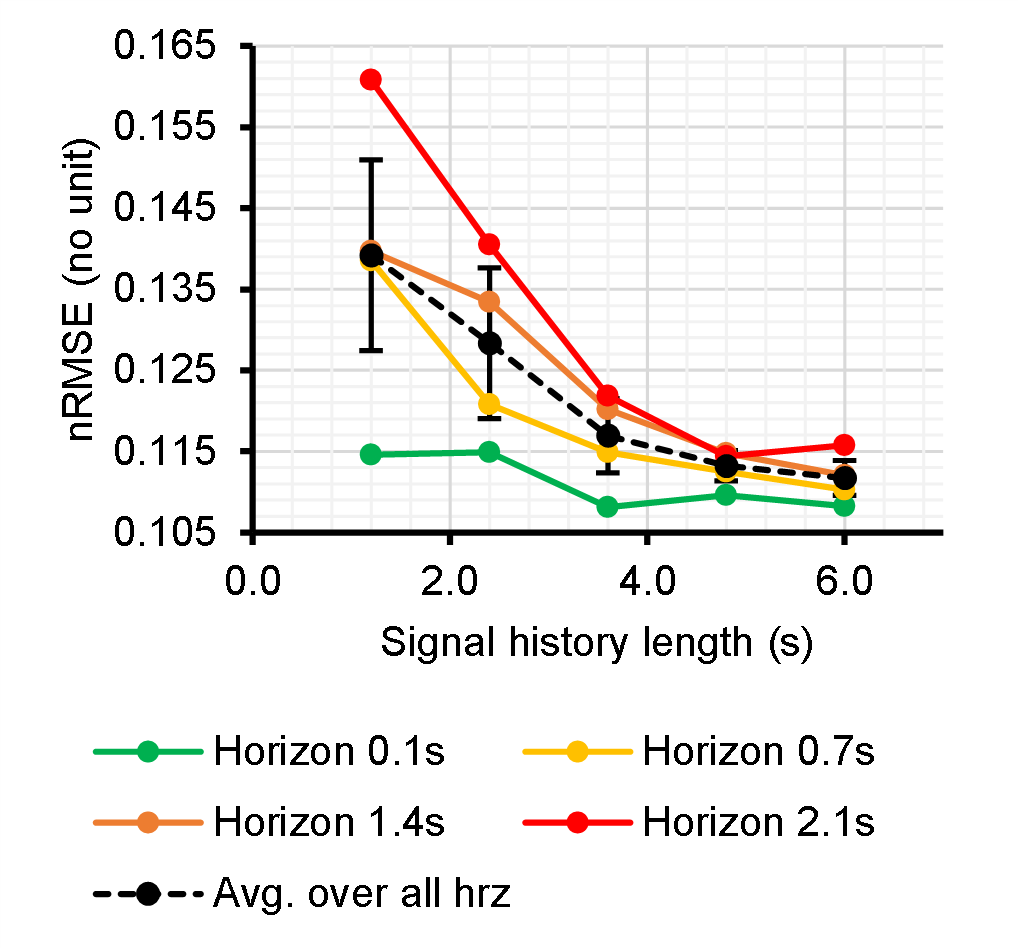} \label{fig:influence_of_SHL_UORO_30Hz}}
    \quad
    \subfloat[SnAp-1 - 3.33Hz input sampling]{\includegraphics[width=.31\textwidth]{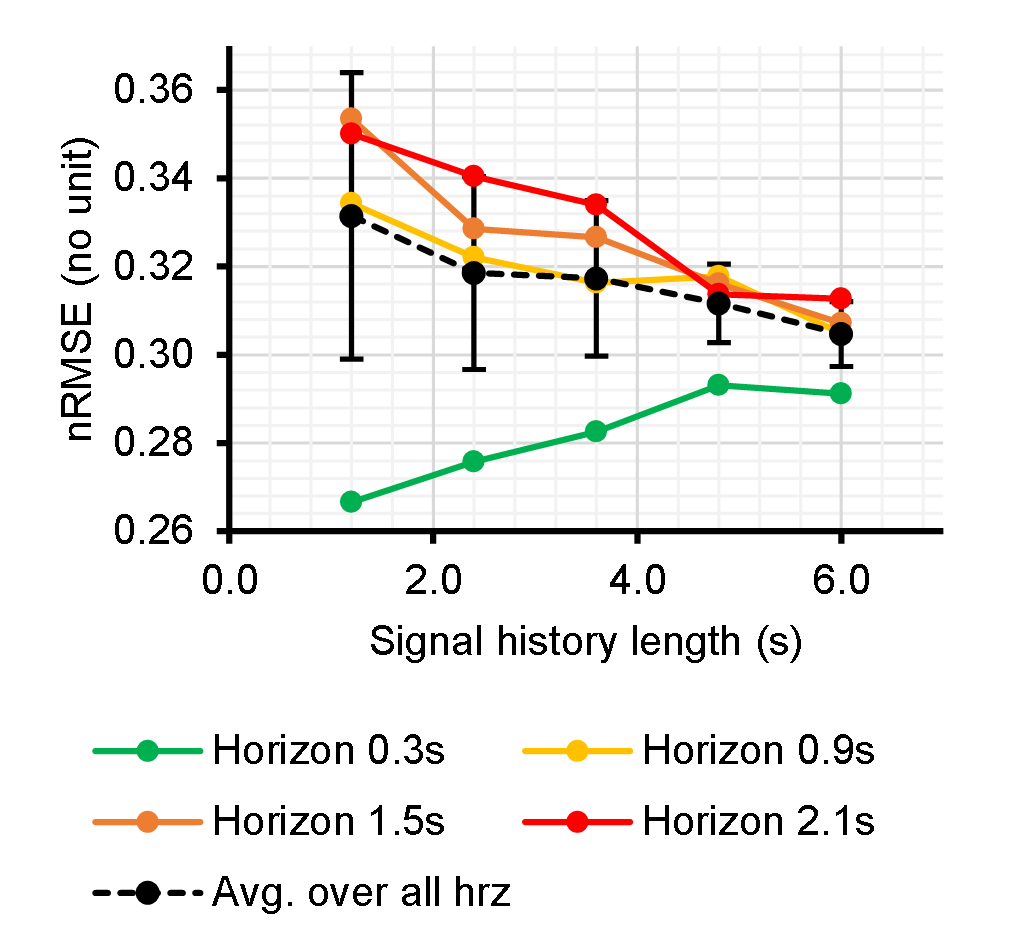} \label{fig:influence_of_SHL_SnAp-1_3.33Hz}}
    \quad
    \subfloat[SnAp-1 - 10Hz input sampling]{\includegraphics[width=.31\textwidth]{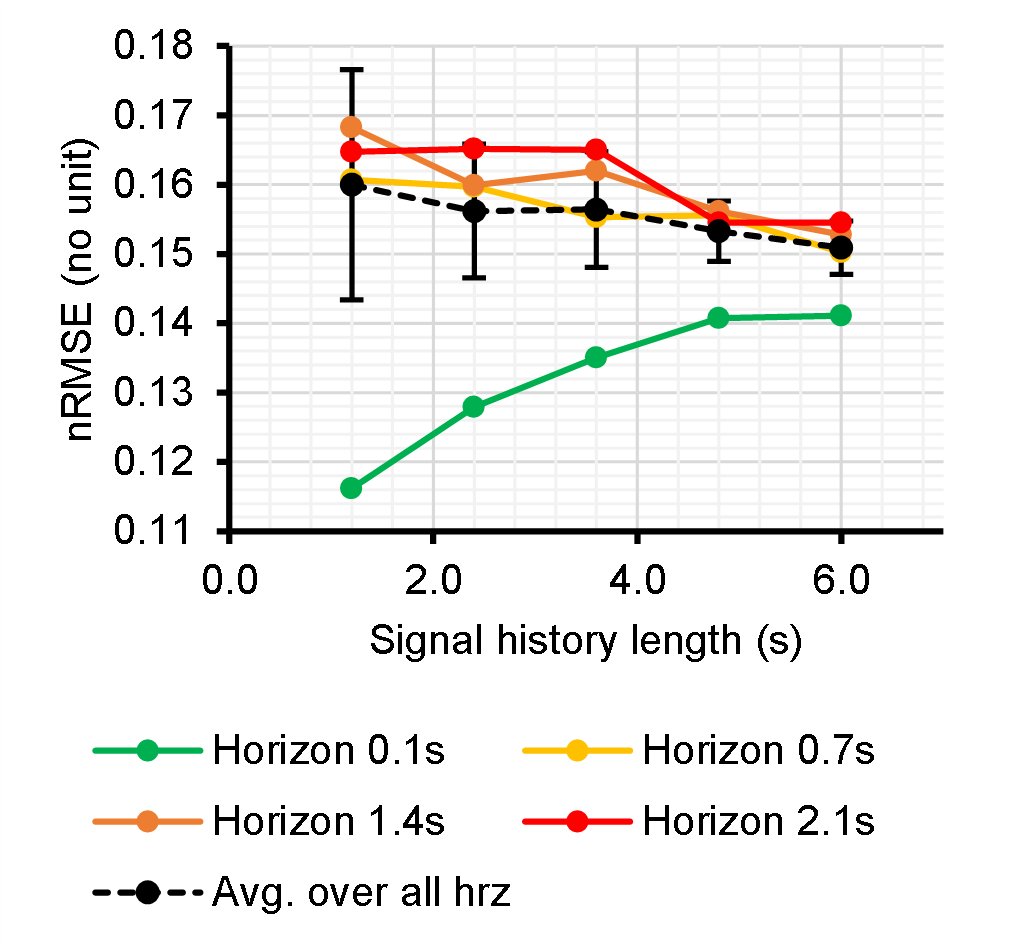} \label{fig:influence_of_SHL_SnAp-1_10Hz}}
    \quad
    \subfloat[SnAp-1 - 30Hz input sampling]{\includegraphics[width=.31\textwidth]{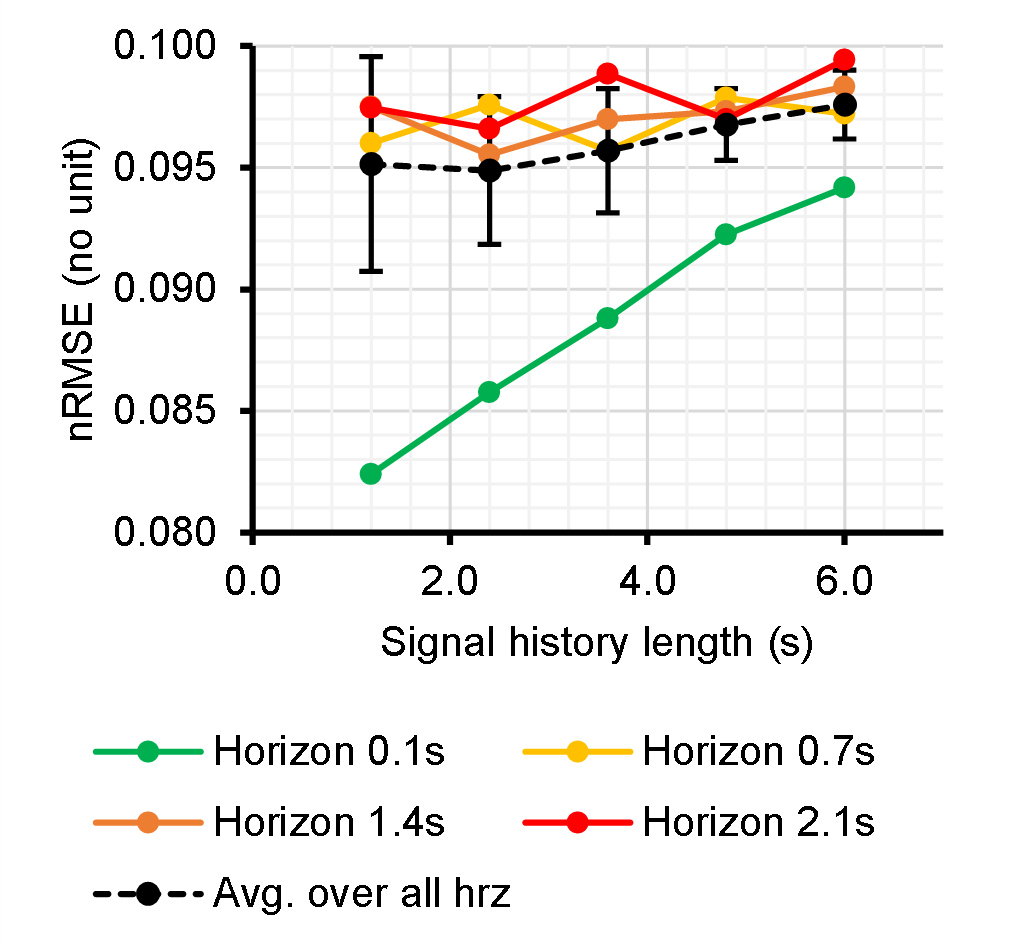} \label{fig:influence_of_SHL_SnAp-1_30Hz}}
    \quad
    \subfloat[DNI - 3.33Hz input sampling]{\includegraphics[width=.31\textwidth]{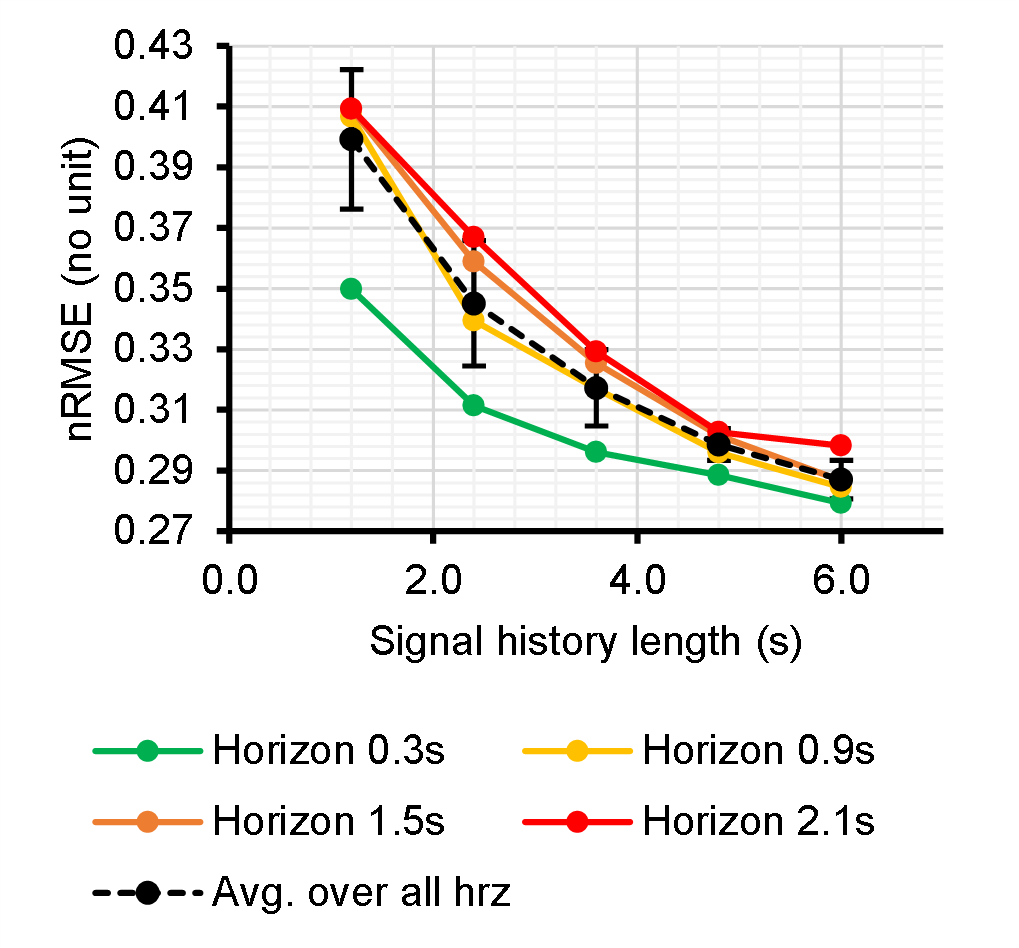} \label{fig:influence_of_SHL_DNI_3.33Hz}}
    \quad
    \subfloat[DNI - 10Hz input sampling]{\includegraphics[width=.31\textwidth]{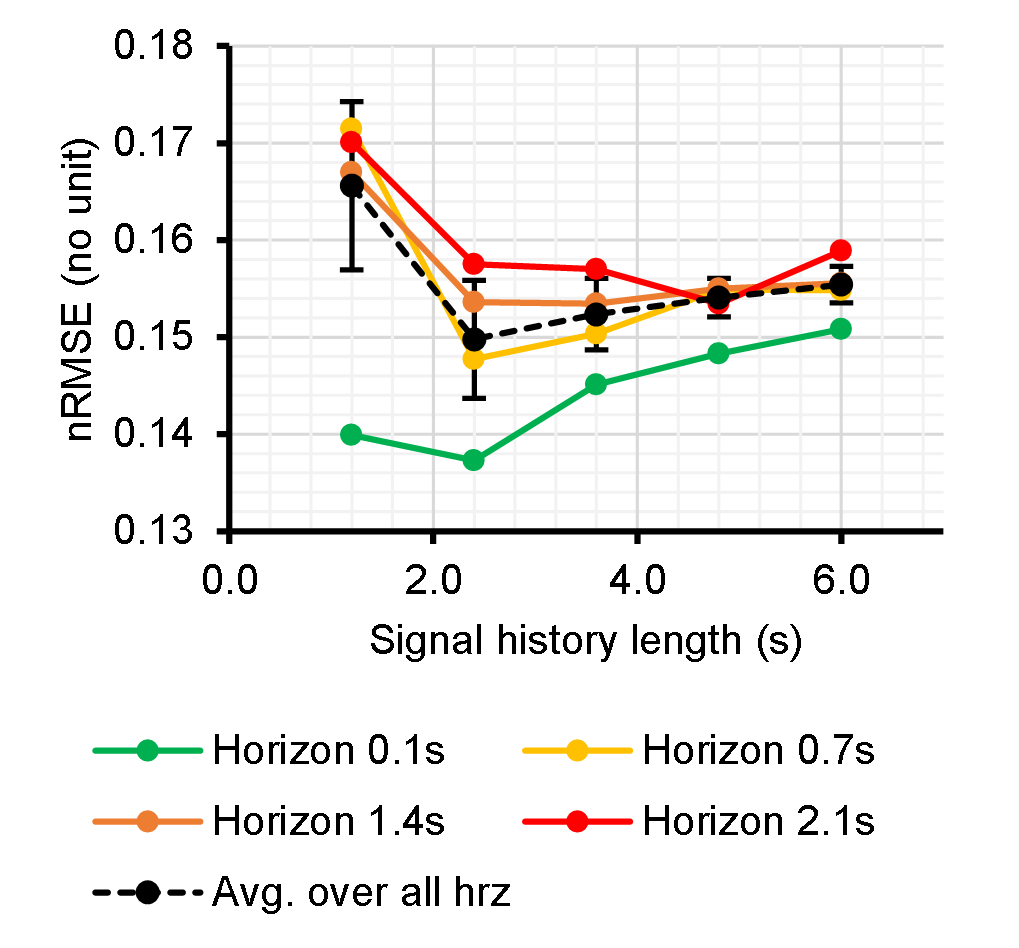} \label{fig:influence_of_SHL_DNI_10Hz}}
    \quad
    \subfloat[DNI - 30Hz input sampling]{\includegraphics[width=.31\textwidth]{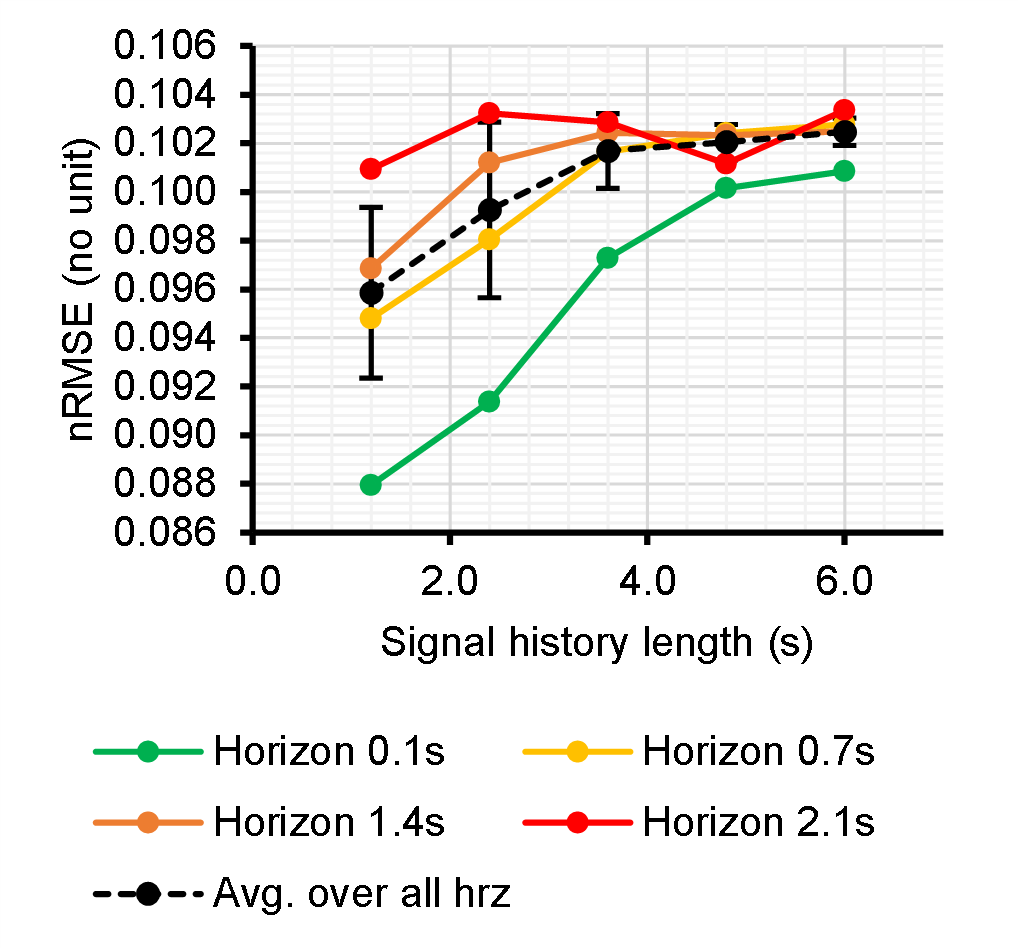} \label{fig:influence_of_SHL_DNI_30Hz}}
    \quad
    \caption{Forecasting nRMSE of UORO, SnAp-1, and DNI of the cross-validation set as a function of the signal history length $L$, for various response times $h$ and input signal sampling frequencies $f$. For each sequence and specific values of $L$ and $h$, we compute the nRMSE minimum over every possible combination of $\eta$ and $q$ within the cross-validation range (Table \ref{table:models comparison}); all errors in that grid are averaged over 50 runs to mitigate RNN stochasticity. Each colored point represents the average of these minimum errors over the nine records. The black dotted curves show the nRMSE minimum, averaged over both the nine respiratory traces and the response times considered, between 0.1s and 2.1s, or between 0.3s and 2.1s if $f=3.33\text{Hz}$. Error bars indicate its standard deviation over these values of $h$.}
    \label{fig:influence of the SHL}
\end{figure*}

The cross-validation nRMSE tended to increase as $h$ increased, as making predictions further in the future becomes more complex (Figs. \ref{fig:learning rate influence}, \ref{fig:influence of the nb of hidden units}, and \ref{fig:influence of the SHL}). On average, over the nine sequences and all the look-ahead values considered, learning rates of $\eta = 0.01$ and $\eta = 0.005$ led to the best cross-validation results at 10Hz and 30Hz, respectively (Fig. \ref{fig:learning rate influence}). $\eta = 0.01$ also led to the lowest cross-validation nRMSE at $f=3.33\text{Hz}$, except for SnAp-1, for which $\eta = 0.02$ was a slightly better choice. The decreasing trend of the nRMSE as $\eta$ decreases at $f=30\text{Hz}$ indicates that a lower nRMSE minimum could plausibly be attained at a value of $\eta$ lower than 0.005. Generally, the optimal learning rate decreases as $f$ increases due to the lower variations between successive marker positions at closer time points. Concerning SnAp-1, the nRMSE corresponding to $h = 2.1\text{s}$ was minimized at $\eta = 0.02$ and $\eta = 0.01$ for $f=10\text{Hz}$ and $f=30\text{Hz}$, respectively (Figs. \ref{fig:influence_of_learning_rate_SnAp-1_10Hz} and \ref{fig:influence_of_learning_rate_SnAp-1_30Hz}). Indeed, a higher learning rate might be necessary to adjust the synaptic weights more strongly when large forecasting errors occur with a relatively high horizon; that phenomenon was also observed in \cite{pohl2022prediction}. However, this should be nuanced, as the graphs corresponding to $h = 2.1\text{s}$ are noisier than those averaged over all values of $h$. This increased variability, along with the uncertainties inherent to the small dataset size, adds to the difficulty of drawing definitive conclusions.

The nRMSE either decreased with $q$ or tended to plateau when $q \geq 90$, for instance, for SnAp-1 at $f=3.33\text{Hz}$, or $q \geq 120$, for UORO at $f=10\text{Hz}$ (Fig. \ref{fig:influence of the nb of hidden units}). There was, however, an increasing trend of the nRMSE of UORO for $q \geq 120$ at $f=3.33\text{Hz}$, although the corresponding confidence intervals were overlapping. The nRMSE minimum was consistently reached at a value of $q$ greater than 90, except for UORO at $f=3.33\text{Hz}$ (Fig. \ref{fig:influence_of_n_hidden_units_UORO_3.33Hz}). 

The optimal SHL (expressed in seconds) for UORO, SnAp-1, and DNI decreased as $f$ increased (Fig. \ref{fig:influence of the SHL}). For DNI, the nRMSE was a decreasing function of the SHL at 3.33Hz, and its minimum was attained at $L=6.0\text{s}$, regardless of the horizon (Fig. \ref{fig:influence_of_SHL_DNI_3.33Hz}). The graph representing the nRMSE of DNI averaged over all horizon values as a function of the SHL at $f=10\text{Hz}$ is convex, and its minimum was attained at $L=2.4\text{s}$. At $f=30\text{Hz}$, the nRMSE of DNI averaged over all horizon values was an increasing function of the SHL; its minimum was achieved at $L=1.2\text{s}$. Remarkably, the errors corresponding to DNI for each horizon represented were also minimized at $L=1.2\text{s}$. The nRMSE of SnAp-1 averaged over all horizon values decreased with the SHL at 3.33Hz and 10Hz. There was also a decreasing error trend for the representative horizon values selected at 3.33Hz and 10Hz, except for $h=0.3\text{s}$. For the latter value of $h$, the nRMSE tended to increase with $L$, and its minimum was invariably attained at $L=1.2\text{s}$, regardless of the frequency. The overall slope of the graph representing the nRMSE of SnAp-1 as a function of the SHL increased between 3.33Hz and 10Hz. That suggests that the optimal SHL (corresponding to the average over all horizons), likely greater than 6.0s, could be closer to 6.0s at $f=10\text{Hz}$ than at $f=3.33\text{Hz}$. The nRMSE of SnAp-1 averaged over all response time values becomes a convex function of the SHL at 30Hz, and its minimum was attained at $L=2.4\text{s}$. Concerning UORO, the nRMSE averaged over all horizon values decreased with the SHL regardless of $f$, and the absolute value of its slope decreased with $f$. The corresponding minima were attained at $L=6.0\text{s}$, except in a few cases. Our results concerning hyperparameter tuning are summarized in Table \ref{table: hyperparameter tuning}.

\begin{table*}[htb!]
\setlength{\tabcolsep}{3pt}
\centering
\begin{tabular}{lll}
\hline
Parameter & Observations       & Recommended value   \\
\hline
\hline
Learning rate $\eta$        & The optimal value of $\eta$ decreased as $f$ increased.                                                                                         & \begin{tabular}[c]{@{}l@{}} $\eta = 0.01$ when $f \leq 10\text{Hz}$ \\$\eta = 0.005$ at $f = 30\text{Hz}$\end{tabular}  \\
Hidden layer size $q$ & The nRMSE decreased as $q$ increased or tended to stay flat when $q \geq 90$. & $q \geq 90$ \\
Signal history length $L$  & The optimal value of $L$ decreased as $f$ increased. & \multirow{2}{*}{$L = 6.0\text{s}$ at $f = 3.33\text{Hz}$} \\
(in s)                          & The nRMSE generally decreased with $L$ at $f = 3.33\text{Hz}$. & \\ 
\hline
\end{tabular}
\caption{Summary of our insights into hyperparameter tuning and selection when using UORO, SnAp-1, and DNI.}
\label{table: hyperparameter tuning}
\end{table*}

\subsection{Time performance}

\begin{figure*}[htb!]
    \centering
    \subfloat[\normalsize Sampling at 3.33Hz]{\includegraphics[width=.33\textwidth]{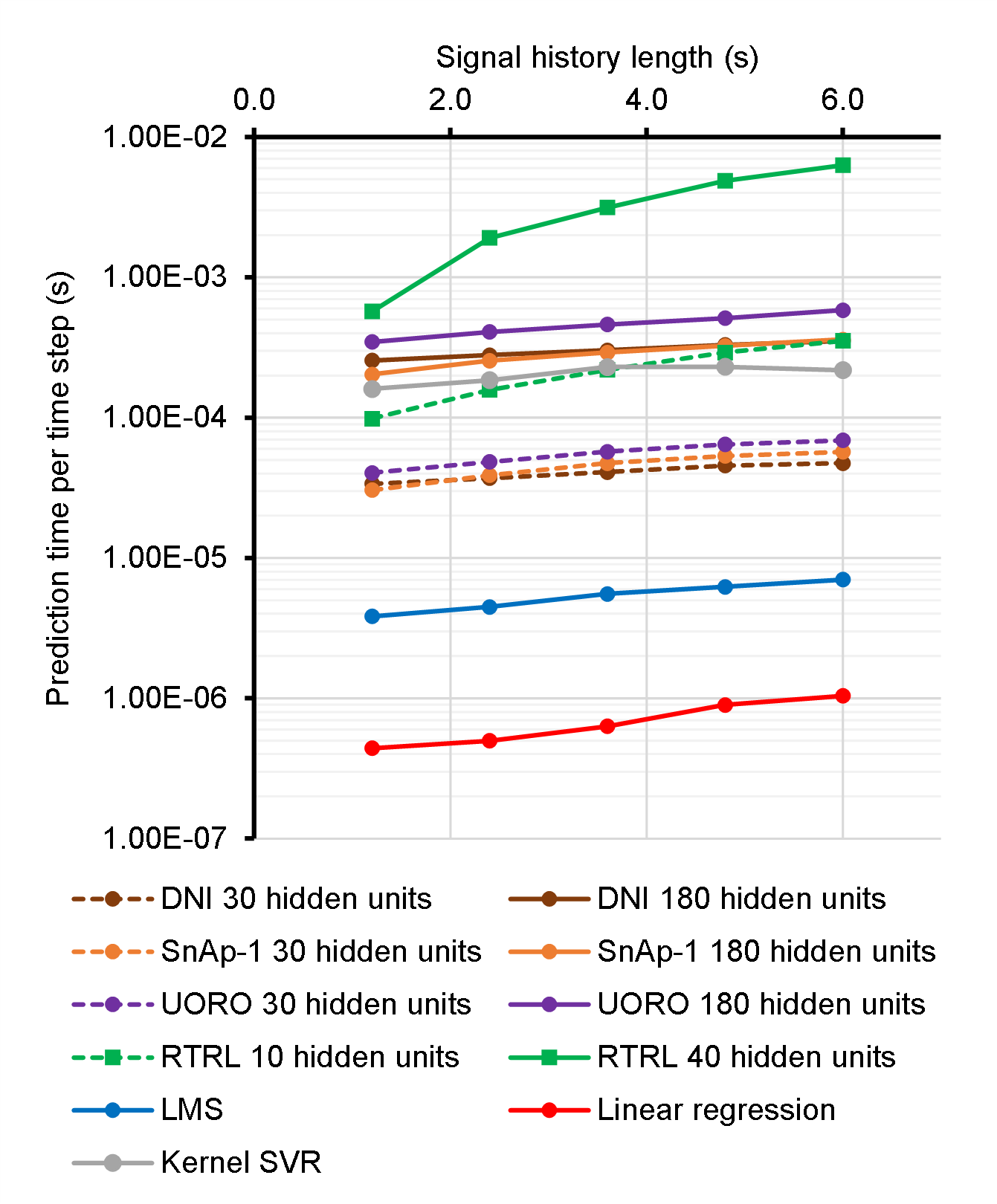}}%
    \subfloat[\normalsize Sampling at 10.0Hz]{\includegraphics[width=.33\textwidth]{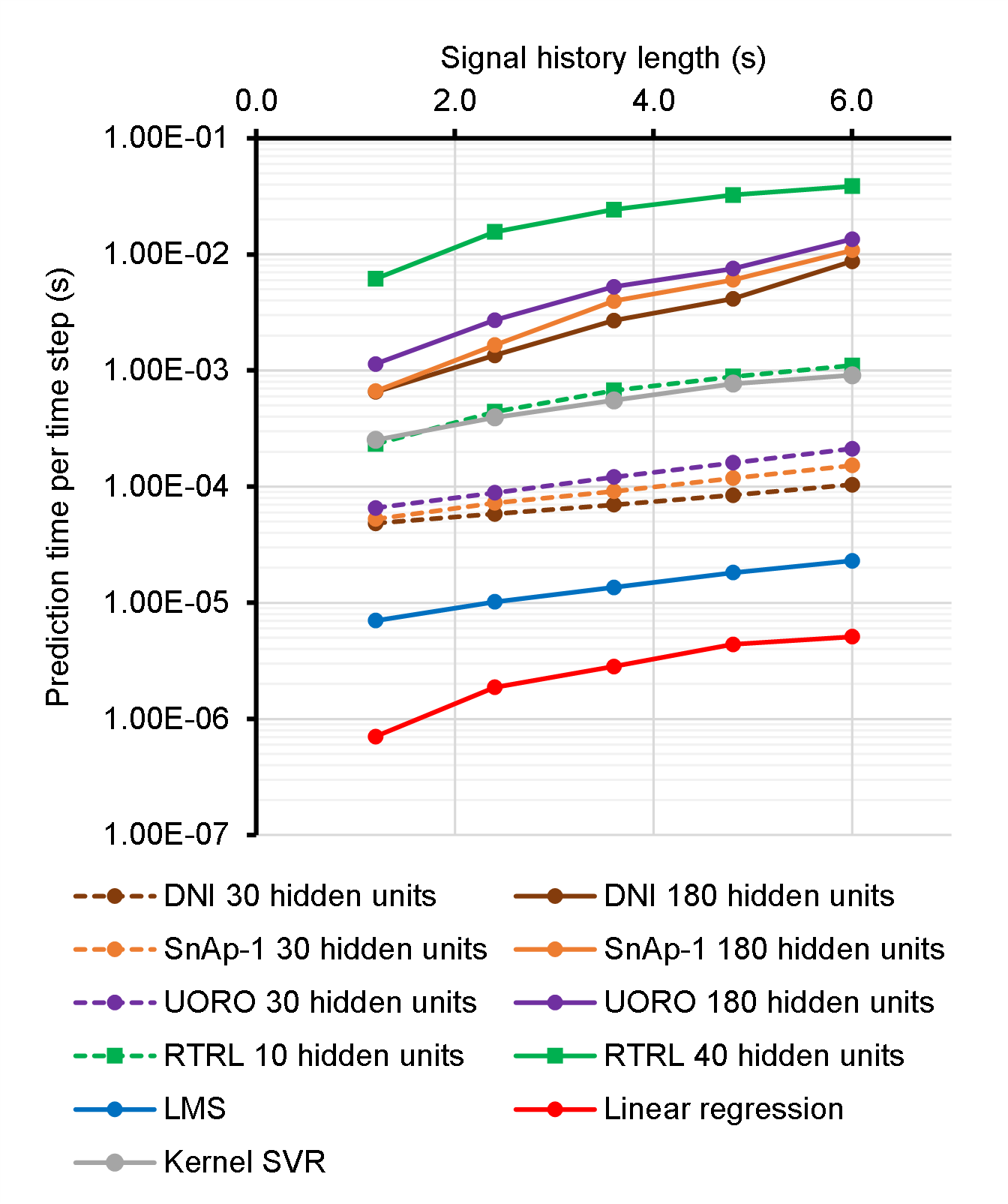} }%
    \subfloat[\normalsize Sampling at 30.0Hz]{\includegraphics[width=.33\textwidth]{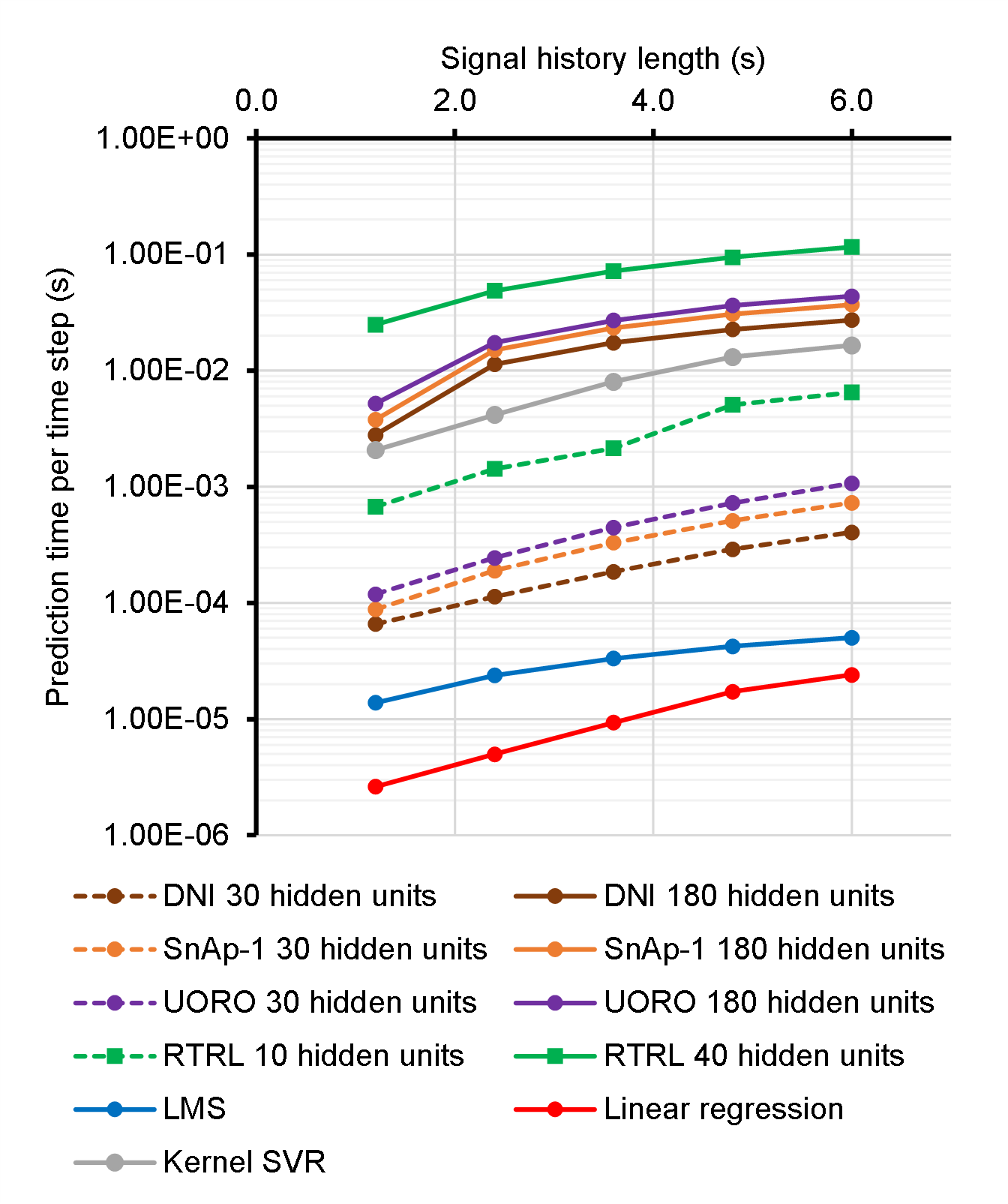} }%
    \caption{Calculation time per time step (Dell 13th Gen Intel Core i7-13700 2.10GHz CPU 16Gb RAM with MATLAB) as a function of the signal history length for different input signal sampling frequencies\protect\footnotemark. }%
    \label{fig:prediction time}%
\end{figure*}

\footnotetext{Each point in the graphs was obtained using parallel processing to loop over multiple horizon values; the computation time per time step for a single horizon without parallel processing may be lower than the one shown here.}

\begin{table}[tb!]
\setlength{\tabcolsep}{3.5pt}
\begin{center}
\begin{tabular}{llll}
\hline
Prediction        & Sampling                    & Sampling                   & Sampling \\
method            & at 3.33Hz                  & at 10Hz                   & at 30Hz \\
\hline \hline
RTRL          & 1.37                       & 10.1                     & 34.0  \\
UORO          & $2.24 \times 10^{-1}$      & 2.05                      & 11.6  \\
SnAp-1        & $1.47 \times 10^{-1}$      & 1.52                      & 9.69  \\
DNI           & $1.46 \times 10^{-1}$      & 1.04                      & 6.83  \\          
LMS               & $5.43 \times 10^{-3}$      & $1.44 \times 10^{-2}$     & $3.27 \times 10^{-2}$              \\
Linear regression   & $7.02 \times 10^{-4}$      & $2.98 \times 10^{-3}$     & $1.17 \times 10^{-2}$             \\
Kernel SVR & $ 2.05 \times 10^{-1}$      & $5.76 \times 10^{-1}$     & 8.79             \\
\hline
\end{tabular}
\end{center}
\caption{Mean inference time per time step in milliseconds (13th Gen Intel Core i7-13700 2.10GHz CPU 16Gb RAM with MATLAB). Each value in the table corresponds to the average over all the SHLs (between 1.2s and 6.0s) and hidden layer sizes considered (between 10 and 40 for RTRL and between 30 and 180 for the other RNN algorithms).}
\label{table:time perf}
\end{table}

The computations were performed using a 13th Gen. Intel Core i7-13700 CPU (2.10GHz), 16Gb RAM, and MATLAB as a programming environment. Linear regression and LMS were the fastest and second-fastest algorithms, respectively (Table \ref{table:time perf}). The inference time of kernel SVR was similar to that of DNI, UORO, and SnAp-1. RNN algorithms were more computationally expensive than LMS; for instance, DNI's inference time was approximately 210 times higher than that of LMS at 30Hz. DNI, UORO, and SnAp-1 had a similar time performance, with DNI being the most efficient and UORO the slowest in our current implementation. That empirical similarity arises from their shared theoretical asymptotic complexity $\mathcal{O}(q(m+p+q))$. RTRL had the worst time performance among all the algorithms considered; its computation time was roughly 10 times higher than that of DNI at $f=3.33\text{Hz}$ and $f=10\text{Hz}$ and 5 times higher at $f=30\text{Hz}$. That is due to the higher asymptotic complexity $\mathcal{O}(q^3 (m+p+q))$ of RTRL. The relatively low processing time of DNI, UORO, and SnAp-1, compared to computationally demanding online algorithms such as RTRL, coupled with their high accuracy, evidenced in Section \ref{section:accuracy and jitter}, makes them a strong candidate for clinical adoption in radiotherapy.

The computation time per time step increased with the sampling frequency, with a mean relative increase for UORO, SnAp-1, and DNI of approximately 8 times and 5 times when $f$ increased from 3.33Hz to 10Hz and from 10Hz to 30Hz, respectively. This is because the number of input units $m$ is proportional to $L$, the number of time steps to make one prediction, and the latter is the product of the sampling frequency $f$ (in Hz) and the signal history length $L_s$ expressed in seconds:
\begin{equation}
L = f L_s
\end{equation}

The computation time also increased with $L_s$. For instance, at $f=30\text{Hz}$, it increased by approximately 11 times, 14 times, and 15 times for UORO, SnAp-1, and DNI, respectively, as $L_s$ increased from 1.2s to 6.0s (Table \ref{table:computation time variation with SHL} in Appendix \ref{appendix: computation time}). However, its variation with $q$ was more significant, as evidenced by a relative inference time increase of 49 times, 59 times, and 76 times for UORO, SnAp-1, and DNI, respectively, at $f=30\text{Hz}$, as $q$ increased from 30 to 180 (Table \ref{table:computation time variation with q} in Appendix \ref{appendix: computation time} and Fig. \ref{fig:prediction time}). This is because the time complexity of these algorithms, $\mathcal{O}(q (m+p+q))$, is characterized by a quadratic variation with $q$ and a linear variation with $m = 3 n_{\text{M}} L$.

\section{Discussion}

\subsection{Comparison with our previous work on external marker position prediction}

Our work follows the general methodology of \cite{pohl2022prediction}, and our main contributions with regard to that previous study are the following:

\begin{enumerate}
\item We compare RTRL and UORO with other online learning algorithms for RNNs, namely SnAp-1 and DNI, and add new calculation elements that enhance the implementation of the two latter algorithms in the case of vanilla RNNs.
\item We study the influence of the respiratory signal sampling frequency on performance.
\item Hyperparameter selection was improved, leading to better accuracy at $f=10\text{Hz}$ in particular
(Table \ref{table:pred perf comparison with previous paper}).  
\end{enumerate}

\begin{table}[htb!]
\small 
\setlength{\tabcolsep}{4.5pt}
\begin{center}
\begin{tabular}{lllll}
\hline
Error     &  Prediction & Previous                            & Current    & Relative \\
type      &  method     & work \cite{pohl2022prediction}      & work       & decrease \\
\hline \hline
MAE       & RTRL        & 0.834mm & 0.653mm & 21.7\% \\
          & UORO        & 0.845mm & 0.535mm & 36.7\% \\[0.1cm]
RMSE      & RTRL        & 1.419mm & 0.926mm & 34.7\% \\
          & UORO        & 1.275mm & 0.755mm & 40.8\% \\[0.1cm]
nRMSE     & RTRL        & 0.303 & 0.195 & 35.6\% \\
          & UORO        & 0.282 & 0.166 & 41.2\% \\[0.1cm]   
Max       & RTRL        & 11.68mm & 5.93mm & 49.2\% \\
error     & UORO        & 8.81mm & 5.48mm & 37.8\% \\[0.1cm]
Jitter    & RTRL        & 0.753mm & 0.647mm & 14.2\% \\
          & UORO        & 0.967mm & 0.655mm & 32.3\% \\                          
\hline
\end{tabular}
\end{center}
\caption{Comparison between the forecasting performance of RTRL and UORO at $f=10\text{Hz}$ in this study (Table \ref{table:pred perf}) and in \cite{pohl2022prediction}. Each error value corresponds to the average of a given performance measure of the test set over the nine sequences and horizon values between 0.1s and 2.1s.}
\label{table:pred perf comparison with previous paper}
\end{table}

Regarding the last point, the overall enhancement in RNN performance was primarily due to better selection of the gradient threshold value, set to $\tau=2.0$ in the previous study and $\tau=100$ in the current one. This higher value allows network parameters to be updated more strongly when the loss gradient norm is high while still ensuring numerical stability. In contrast, the lower value of $\tau$ in the previous work limited RNN adaptation in the presence of loss gradients with high norms, thereby hindering performance. To account for the higher threshold $\tau$, we also modified the range of learning rates in this study, using lower values ranging from 0.005 to 0.02, compared to those selected in \cite{pohl2022prediction} (between 0.02 and 0.2). Indeed, it is likely that gradient clipping happened relatively frequently in the setting of the previous study, as the learning rates were relatively high, resulting in lower performance. The results reported in our current work could be improved in the 30Hz scenario by using even lower values for $\eta$, as suggested by Figs. \ref{fig:influence_of_learning_rate_UORO_30Hz}, \ref{fig:influence_of_learning_rate_SnAp-1_30Hz}, and \ref{fig:influence_of_learning_rate_DNI_30Hz}. Indeed, several studies on respiratory motion forecasting at a sampling rate close to 30Hz recommend using a learning rate between 0.001 and 0.005 \cite{samadi2023respiratory, lin2019towards, yu2020rapid}.

The RNN accuracy improvements can also be attributed to the inclusion of higher values for $q$ in the hyperparameter search grid. Indeed, the number of hidden units spans from $q = 10$ to $q = 90$ in \cite{pohl2022prediction} and from $q = 30$ to $q = 180$ in our current work. It was observed in \cite{POHL2021101941, pohl2022prediction} that a relatively high value of $q$ was preferable when predicting respiratory motion using a vanilla RNN with a single hidden layer, which is confirmed by our current study (Fig. \ref{fig:influence of the nb of hidden units}). Although we selected lower values of $q$ for RTRL in this work than in \cite{pohl2022prediction} to accelerate inference and grid search, performance also improved for RTRL, indicating that correctly setting the values of $\tau$ and $\eta$ is critical. 

Our findings support the observation in \cite{verma2010survey} that LMS surpasses linear regression at moderate and high horizons (Fig. \ref{fig:nRMSE_for_different_horizons}). Although using LMS at medium look-ahead times was recommended in \cite{pohl2022prediction}, our current work demonstrates that RNNs trained online can outperform it with appropriate hyperparameter selection. Cross-validation and inference with LMS are faster, but RNNs have better overall accuracy when correctly tuned, and LMS appears to be more unstable with regard to changes in $f$ (Section \ref{section: experimental design}). The superiority of RNNs concerning the latter point may result from the ability of the hidden layer to cope with variations in signal scale and provide robust signal representation to the output layer.

\subsection{Significance of our results relative to the dataset and literature}
\label{section: comments on dataset}

The pertinence and value of our dataset are discussed in \cite{pohl2022prediction}, which we can summarize as follows. On the one hand, it is publicly available online \cite{pohl_michel_2021_5506965} and includes a relatively large variety of respiratory patterns. On the other hand, its size is relatively small compared to other datasets used in some of the recent studies about respiratory motion forecasting. However, our results are still significant, as most of our observations, such as the superiority of linear methods and neural networks at low and high look-ahead times, respectively, align with the literature (Section \ref{section:intro pred in radiotherapy}), and ANNs trained online can learn from little data. 

Our work is one of the few that highlights the influence of both signal sampling frequency and response time on forecasting accuracy, with low frequencies around 3.33Hz being typical of image acquisition during MR-guided LINAC treatment and high frequencies more common in marker-based or externally tracked radiotherapy. The sampling rate had a high impact on performance. Still, there was no significant increase in the errors associated with RNNs when $h$ increased, except at 3.33Hz, even though we considered the most extensive range of values for $h$ within the literature, to the extent of our knowledge. We hypothesize that that was due to the relatively small size of our dataset, the robustness inherent to online learning algorithms, and judicious hyperparameter selection.

While we could ascertain the superiority of DNI, UORO, and SnAp-1 compared to the other algorithms in our study with a high degree of confidence (Table \ref{table:pred perf}), it was more challenging to draw firm conclusions regarding the relative performance of these three algorithms given their similar accuracy and the moderately low amount of data in our study. Regarding the latter point, \citeauthor{marschall2020unified} reported that DNI outperformed UORO on the "Mimic" task, corresponding to a few input units ($m = 1$) and a comparatively long time horizon of 10 steps \cite{marschall2020unified}. In contrast, UORO performed better than DNI on the "Add" task, which requires memorizing more information, with $m = 32$ and $h$ "likely shorter than 10 time steps." The authors hypothesized that "UORO [...] is effective at maintaining information over time, but the stochasticity in the updates places a limit on how much information can be retained. [...] Perhaps UORO [...] produces gradients with a limited amount of information that survives many updates, while DNI [...] has a larger information capacity but a limited time horizon." Our simulations are closer to the "Add" task, as they are characterized by a relatively high number of inputs, with $m \in \{324, ..., 1620\}$ and $h \in \{3, ..., 63\}$ at $f=30\text{Hz}$. Still, we could not demonstrate the superiority of DNI compared with UORO. However, our experiments differ from those in \cite{marschall2020unified}, as the values of $m$ and $h$ explored in our study are higher. Moreover, our implementation of DNI (with the full update rule for $A$) differs, as our expression for the gradient of $\|f(A)\|^2$ takes into account the $\tilde{x}_{n+1}^T f(A) D_n^T$ term neglected in \cite{marschall2020unified} (Eq. \ref{eq:final expression of grad(norm(f(A)))}). In addition, SnAp-1 was reported to surpass UORO on the WikiText103 language modeling task, in experiments involving dense GRUs with 128 recurrent units \cite{menick2020practical}. Nonetheless, the authors noted that "language modeling does not directly measure a model's ability to learn structure that spans long time horizons." However, that study and ours are difficult to put into perspective, as language modeling is a distinct task requiring more data, whereas our time-series dataset is limited in size. Furthermore, our implementation differs from that in \cite{menick2020practical}, as, for instance, the latter work employs GRUs instead of vanilla RNNs.

Forecasting the motion of markers associated with normal breathing resulted in higher accuracy than irregular breathing at $f=3.33\text{Hz}$ and $f=10\text{Hz}$ (Section \ref{section:accuracy and jitter} and Appendix \ref{appendix:regular vs irregular perf}). Still, these differences were less apparent at 30Hz. This may be because regular and irregular signals appear more similar locally (within a window of $h$ time steps) as $f$ increases. Specifically, the difficulty gap between forecasting regular and irregular signals narrows when $h$ becomes small relative to $f$. A significant yet less pronounced forecasting performance discrepancy may exist at 30Hz, but more data is required to confirm that. Regardless, this indicates good intrinsic robustness of RNNs to sudden changes in respiratory patterns at high sampling frequencies, similar to that of transformers at high horizons, observed in \cite{jeong2022clinical}. Noticeably, the latter study reported RMSE increases of 15\% and 17\% for LSTMs and transformers, respectively, between steady signals and unsteady ones featuring irregular periods and amplitudes, sampled at 20Hz. These are similar to the average 20\% increase in RMSE that we observed at 10Hz (Table \ref{table:regular vs irregular breathing} in Appendix \ref{appendix:regular vs irregular perf}). Furthermore, it has been highlighted that subjects breathing faster tend to have respiratory traces harder to forecast \cite{liang2023real}. Therefore, in our experiments, when comparing results for regular and irregular motion, we removed one of the sequences that features a lower breathing speed (cf footnote \ref{footnote: slow motion sequence removed} and Fig. \ref{fig:coordx_marker1_seq7}). The nRMSE associated with that sequence was approximately half of that averaged over the nine sequences, for all the RNN algorithms and sampling frequencies investigated.

In our work, the average cross-validation nRMSE decreased as $L$ increased at $f=3.33\text{Hz}$ for UORO, SnAp-1, and DNI (Fig. \ref{fig:influence of the SHL}). This is in disagreement with the observations in \cite{romaguera2021probabilistic} about chest image prediction at low sampling rates using MRI and ultrasound sequences with a temporal resolution of 450ms and 250ms, respectively. That study reported that performance generally increased together with the SHL. However, the research goal (predicting videos accurately) and the network designed to achieve that task, based on the combination of a conditional variational autoencoder and LSTMs, differ from those in our work. Likewise, \citeauthor{yao2022feature} found that for a signal sampled at a high frequency (30Hz $\sim$ 45Hz) and a low horizon ($h=150\text{ms}$), the forecasting accuracy increased with the SHL. Nevertheless, that was not the case for SnAp-1 and DNI in our experiments (Figs. \ref{fig:influence_of_SHL_SnAp-1_30Hz} and \ref{fig:influence_of_SHL_DNI_30Hz}); attention mechanisms might indeed help select more pertinent features when the value of $L$ is higher \cite{yao2022feature}. Alternatively, using architectures such as LSTMs, more suited for capturing long-range dependencies than standard RNNs \cite{hochreiter1997long}, which we selected in our work for their simplicity, may help achieve better performance with higher SHLs. Moreover, \citeauthor{samadi2023respiratory} claimed that "for a higher system latency, a larger input window is required" \cite{samadi2023respiratory}, but that was not consistently validated in our experiments, for instance, when considering the UORO validation curves for $h=0.3\text{s}$ and $h=2.1\text{s}$ in Fig. \ref{fig:influence_of_SHL_UORO_10Hz}. Generally, a low SHL may correspond to an amount of information fed to the network that is insufficient for accurate prediction. In contrast, higher values of $L$ may make the predictor less responsive to high-frequency signal components.

\subsection{Performance comparison with previous works}

\begin{table*}[htb]
\scriptsize
\setlength{\tabcolsep}{2.5pt}
\begin{center}
\begin{tabular}{llllllll}
\hline
Network                                      & Breathing       & Sampling     & Amount of              & Signal      & Response  & Prediction error      \\
                                             & data            & rate         & data                   & amplitude   & time      & and inference time            \\
\hline
\hline
1-layer MLP with                                & CyberKnife       & 7.5Hz      & 27 records             & 2mm         & 650ms    & MAE 0.65mm, RMSE 0.95mm,      \\                               
adaptive retraining \cite{teo2018feasibility}   & data             &             & of 1min               & to 16mm     &           & Max error 3.94mm \\[0.1cm]                                                              
3-layer LSTM with                            & Tumor 3D         & 25Hz       & 158 records            & 0.6mm       & 280ms     & RMSE 0.9mm       \\                                                              
adaptive retraining  \cite{yun2019deep}      & center of mass   &             & of 8min               & to 51.2mm   &           &                  \\[0.1cm]  
LSTM followed                             & RPM data         & 30Hz       & 550 records lasting           & 11.9mm      & 200ms     & RMSE 0.28mm      \\
by FCLs \cite{lee2021geometric}           & (Varian)         &             & between 91s and 488s            & to 25.9mm   &           &                  \\[0.1cm]                                 

5-layer TCN                                  & CyberKnife       & 25Hz       & First 3.5min of       & -           & 1) 400ms & 1) RMSE 0.67mm   \\                               
with residual                                & data             &             & 69 traces from         &             & 2) 560ms & 2) RMSE 0.81mm   \\
connections \cite{chang2021real}             &                  &             & 21 patients            &             &           &                  \\[0.1cm]                                                                

2-layer LSTM                                 & External markers & 20Hz       & 7 records lasting             & -           & 450ms     & z-coordinate errors: MAE 0.3mm, \\
\& 2 FCLs \cite{wang2021real}                 & (AccuTrack 250)  &             & between 5min and 6min           &             &           & RMSE$<$0.5mm, max error 1.5mm \\ [0.1cm]                               

3 or 5-layer                                 & Tumor centroid   & 4Hz        & 16.1h and 1.5h of data    & -           & 1) 250ms & 1) RMSEs 0.48mm \& 0.42mm \\                               
LSTM trained                                 & SI coordinate    &             & for 2 cohorts (88    &             & 2) 500ms & 2) RMSEs 1.20mm \& 1.00mm, \\
offline and                                  & from sagittal    &             & and 3 cancer patients, &             &           &    nRMSEs 0.086 \& 0.107  \\                                                              
retrained online \cite{lombardo2022offline}  & 2D cine-MRI      &             & respectively)          &             & 3) 750ms  & 3) RMSEs 2.20mm \& 1.77mm \\[0.1cm]                                                                
                                    
TCN followed by a                              & 2D target        & 30Hz       & 2min videos from      & -           & 1) 150ms & 1) MAE 0.88mm, \\                               
3-layer self-attention                       & trajectories     & to 45Hz    & 58 subjects            &             &           & RMSE 1.09mm, nRMSE 0.08 \\
module and linear                            & from liver       &             &                        &             & 2) 400ms & 2) MAE 2.08mm,  \\                                                              
autoregressive model \cite{yao2022feature}   & ultrasound       &             &                        &             &           & RMSE 2.63mm, nRMSE 0.18 \\[0.1cm]                                                                                  
  
2-layer LSTM, TCN,                           & CyberKnife       & 26Hz       & 304 traces from        & -           & 1) 231ms & 1) MAE 0.088mm, \\                               
external attention module,                   & data             &             & 31 patients            &             &           & nRMSE 0.028   \\
2 FCLs, and linear                           &                  &             & with a 71-min          &             & 2) 923ms & 2) MAE 0.31mm, \\                                                              
autoregressive model \cite{zhang2023lgeanet} &                  &             & average duration       &             &           & nRMSE 0.31  \\[0.1cm]                                                                                          
 
2-layer transformer                          & CyberKnife       & 26Hz       & 304 traces lasting        & -           & 1) 200ms & 1) MAE 0.24mm, RMSE 0.32mm \\                               
encoder module                               & data and         &             & from 6.5min to 132min         &             & 2) 400ms & 2) MAE 0.34mm, RMSE 0.45mm   \\
followed by a                                & augmentation     &             & Augmentation doubled   &             & 3) 600ms & 3) MAE 0.36mm, RMSE 0.50mm \\                                                              
2-layer LSTM \cite{tan2022lstformer}         & data             &             & the nb. of time steps.  &             &           & inference time from 22ms to 66ms \\[0.1cm]

\hline
1 \& 2) 1-layer RNN                          & 3 external       & 1) 3.33Hz  & 9 records              & 6mm         & 0.1s      & 1) MAE 1.09mm, RMSE 1.53mm,      \\
trained with SnAp-1                          & markers          &             & from 3 subjects            & to 40mm     & to 2.1s   & nRMSE 0.33, max error 8.45mm \\
                                             & (Polaris)        & 2) 10Hz    & lasting 73s to 222s                       & (SI         &           & 2) MAE 0.49mm, RMSE 0.70mm, \\
                                             &                  &             &                        & direction)  &           & nRMSE 0.16, max error 5.60mm       \\          
3) 1-layer RNN                               &                  & 3) 30Hz    &                        &             &           & 3) MAE 0.31mm, RMSE 0.40mm,      \\
trained with UORO                            &                  &             &                        &             &           & nRMSE 0.086, max error 3.29mm       \\                  
                                             &                  &             &                        &             &           & inference time of 12ms (at 30Hz)      \\          
\hline 
\end{tabular}
\end{center}
\caption{Comparison of the performance of RNNs in our study with results in the literature about respiratory motion prediction with ANNs for radiotherapy (cf Sections \ref{section:intro pred in radiotherapy}, \ref{section:intro RNN and transformers}, and \ref{section:intro parameter adaptation}). The term "RNN" refers here to a vanilla RNN, as opposed to LSTMs. A field with " - " indicates that the information is not available in the corresponding research article. The performance of the RNNs in our work is reported in the last rows\protect\footnotemark.}
\label{table:comparison_with literature}
\end{table*}

\footnotetext{FCL stands for "fully connected layer." By abuse of language, the number of layers reported actually corresponds to the number of hidden layers. For example, a "1-layer MLP" architecture designates an MLP with a single hidden layer. The accuracy metrics corresponding to our work, located in the last rows, are the averages over the horizon values between 0.1s and 2.1s in Table \ref{table:pred perf}, and the inference time reported is the mean computation time per time step over the values of $q$ and $L$ in the cross-validation range in Table \ref{table:time perf}.} 

In this section, we compare the performance of RNNs trained with UORO, SnAp-1, and DNI in our study with that of other ANNs in previous studies on breathing motion forecasting (summary in Table \ref{table:comparison_with literature}). This comparison is challenging, especially because the data utilized differ from study to study. Specifically, respiratory signals may be subject to varying degrees of irregularity, such as abnormal sudden motion, shifts, and drifts. They may be characterized by diverse distributions of breathing amplitudes and frequencies. Moreover, the procedure for partitioning the data into the training set and test set also differs, with distinct arbitrary choices regarding, for instance, the amount of training data relative to the testing data and whether some traces are entirely excluded from the training set. Some datasets comprise more data than others and are publicly available, which indicates potentially more generalizable results. This is, for example, the case of the CyberKnife data from Georgetown University \cite{Ernst_2013}, used for instance in \cite{tan2022lstformer} and \cite{zhang2023lgeanet}, among the studies in Table \ref{table:comparison_with literature}. In addition, the way performance metrics are defined may vary among previous works. For instance, normalization by the amplitude and standard deviation of the signal is conducted in \cite{lombardo2022offline} and \cite{zhang2023lgeanet}, respectively, to compute the nRMSE. Moreover, some previous studies reported metrics using data whose amplitude was rescaled from -1 to 1 \cite{lin2019towards, sun2017respiratory} \footnote{Nevertheless, this is not the case for any of the studies in Table \ref{table:comparison_with literature}.}. Last, many related studies focused on 1D respiratory signal forecasting, whereas we perform 3D signal prediction and report errors in the 3D Euclidean space. Despite those intricacies, a comparison is still valuable, as it provides a general idea about the performance of the algorithms in our research relative to the results reported in the literature.

Concerning prediction with low sampling frequencies, the deep LSTM trained with a 4Hz signal in \cite{lombardo2022offline} achieved lower RMSEs at $h=250\text{ms}$ and $h=500\text{ms}$ than the RNN trained with SnAp-1 at $f=3.33\text{Hz}$ in our work\footnote{Lower nRMSEs at $h=0.5\text{s}$ were also reported in \cite{lombardo2022offline}, but a different normalization factor was used.}. Indeed, the latter reached an average RMSE of 1.53mm over response times between 0.3s and 2.1s. However, that error is 14\% and 30\% lower than those corresponding to the same LSTM at $h=750\text{ms}$ \cite{lombardo2022offline}. Furthermore, \citeauthor{lombardo2022offline} preprocessed the data using future information, for instance, by normalizing it between -1 and 1 using the global extrema in each sequence. Other preprocessing steps, such as smoothing the data and excluding sequences with low-amplitude motion where noise is more prevalent, might also have led to a potentially overestimated accuracy. Regarding the prediction of a 7.5Hz breathing signal with an MLP, \citeauthor{teo2018feasibility} reported an MAE and RMSE equal to 0.65mm and 0.95mm, respectively, which are between those that we obtained at 3.33Hz and 10Hz with SnAp-1\footnote{The maximum error attained in \cite{teo2018feasibility} was 30\% lower than that achieved by SnAp-1 at $f=10\text{Hz}$ in our work, but maximum errors might be less indicative of the general performance of an algorithm due to higher variance.} \cite{teo2018feasibility}. Nonetheless, the response time in that study was relatively low compared to those that we considered, and the signal amplitudes were also roughly 2 to 3 times lower than in our work, which suggests that SnAp-1 may perform better on that dataset. Similarly, \citeauthor{wang2021real} reported an RMSE below 0.5mm using a deep LSTM predicting data from the AccuTrack 250 system sampled at 20Hz \cite{wang2021real}. That error falls between those achieved by SnAp-1 at 10Hz (0.70mm) and 30Hz (0.41mm) in our research. Moreover, that LSTM attained a maximum error and an MAE lower than those corresponding to UORO at 30Hz in our work. Still, those were coordinate-wise errors, and the associated look-ahead time was relatively low compared to those that we investigated. 

Regarding prediction at high sampling rates, a deep LSTM and a TCN were proposed in \cite{yun2019deep} and \cite{chang2021real}, respectively, to predict respiratory motion at 25Hz. These networks led to higher RMSEs---0.9mm for the LSTM and 0.68mm for the TCN---than that of UORO at 30Hz in our study (0.40mm). This was despite relatively shorter response times (280ms for the LSTM and 400ms for the TCN) and the similarity between the signal amplitude in \cite{yun2019deep} (0.6mm to 51.2mm) and our study (6mm to 40mm). Likewise, the nRMSE corresponding to prediction at $f=26\text{Hz}$ using an architecture combining LSTMs, TCNs, external attention, and a linear autoregressive model in \cite{zhang2023lgeanet}, equal to 0.31, was approximately 3 times higher than that of UORO at $f=30\text{Hz}$, despite the low horizon $h=231\text{ms}$ in that work. Similarly, \citeauthor{tan2022lstformer} forecast 26Hz CyberKnife data with a network comprised of a transformer encoder and LSTM layers and reported MAEs and RMSEs at $h \geq 400 \text{ms}$ higher than those of UORO at $f=30\text{Hz}$ in our research \cite{tan2022lstformer}. In addition, the associated inference time was twice as high as that of UORO due to the computational burden introduced by the transformer module. \citeauthor{lee2021geometric} predicted 30Hz real-time position management (RPM) data using an LSTM network and achieved an RMSE of 0.28mm, lower than that associated with UORO at 30Hz \cite{lee2021geometric}. Still, the time series in that study had lower amplitudes, and the response time considered (200ms) was short. Last, an architecture combining a linear autoregressive model and TCN with self-attention was proposed in \cite{yao2022feature} to predict 2D target trajectories from liver ultrasound imaging. It led to MAEs and RMSEs higher than those corresponding to UORO despite the higher sampling rate (up to 45Hz) and relatively low response time (up to 400ms) considered in that work.

We need to nuance the relatively high accuracy of online learning algorithms for RNNs in Table \ref{table:comparison_with literature} by mentioning two studies that seem to indicate higher performance of deep learning approaches. First, \citeauthor{jeong2022clinical} achieved an RMSE of 0.15mm at $h = 500\text{ms}$ with a transformer architecture (comprised of 6 encoder and decoder layers) predicting a respiration gating signal consisting of the distance from a laser source to the body surface of cancer patients, using a dataset of 540 respiratory traces from 442 subjects sampled at 20Hz \cite{jeong2022clinical}. These lasted from 84s to 273s, with an average recording time of 145s, and were characterized by a mean amplitude in the SI direction of 11mm $\pm$ 8mm (standard deviation). Likewise, \citeauthor{samadi2023respiratory} also achieved higher performance using a GRU trained with 26Hz CyberKnife VSI data comprising 800 records between 23min and 60min from 30 lung and abdominal cancer patients. The associated MAE, RMSE, and nRMSE\footnote{Min-max amplitude normalization was used to compute the nRMSE in \cite{samadi2023respiratory}, which makes comparison with our work difficult.} at $h = 115\text{ms}$ were equal to 0.086mm, 0.108mm, and 0.031, respectively \cite{samadi2023respiratory}. However, the accuracy corresponding to $f=30\text{Hz}$ in our study might seem lower because we report 3D errors, and irregular breathing sequences constitute almost half of our entire dataset. Also, as suggested in Fig. \ref{fig:learning rate influence}, we may achieve better performance by selecting lower learning rates at 30Hz. More importantly, rather than learning general respiratory motion characteristics from a large dataset, our complementary approach extracts a meaningful representation from the limited information of a single subject's breathing trace. With that approach, we achieved better or similar performance than most recent methods relying on complex architectures and much training data (Table \ref{table:comparison_with literature}). Beyond being more privacy-friendly, our method requires only a one-minute acquisition of marker trajectories before treatment, which should not be a clinical burden. However, cross-validation might be computationally expensive and could delay the start of treatment. One could also use online learning algorithms to fine-tune in real time the weights of an RNN model previously trained with a large database, allowing it to specialize on a single patient and thereby achieve higher performance during treatment.

\citeauthor{johl2020performance} and \citeauthor{li2023online} claimed that linear regression was better suited than neural networks for predicting breathing movements \cite{johl2020performance, li2023online}. This may be due to their experimental setup, where they selected low horizon values relative to the signal sampling frequency, namely $h = 160\text{ms}$ for $f = 25\text{Hz}$ and $h = 400\text{ms}$ for $f = 5\text{Hz}$, respectively. Even though we found RNNs to be more effective overall, linear regression performed comparably or better when $h$ is low relative to $f$ (see, for instance, Fig. \ref{fig:nRMSE_for_different_horizons}). In addition, we observed that RNNs trained online were quite robust at high horizon values. By contrast, most previous studies reported a general performance decrease as $h$ increased. This was not very apparent in our study, except for $f = 3.33\text{Hz}$, which might come from a variety of reasons: the low amount of data might introduce significant noise when measuring performance, the horizon values examined might be low relative to the sampling frequency when $f \geq 10\text{Hz}$, and cross-validation is relatively extensive in our work. However, we have already considered some of the highest values of $h$ within the literature on respiratory motion forecasting. Instead, we hypothesize that RNNs trained online are inherently capable of achieving accurate predictions for high-latency systems, even with a moderate amount of data.

\subsection{Future works}

In subsequent studies, LSTM or GRU networks may be employed in lieu of a basic RNN structure to enhance forecasting accuracy. Additionally, fast online learning algorithms such as those examined in this work could dynamically retrain in real time the final hidden layer of a deep RNN predicting respiratory waveform signals, thereby enhancing its robustness to unforeseen instances of irregular breathing patterns. Generally, the advancement of efficient online learning algorithms for RNNs will positively impact tumor position forecasting in lung radiotherapy. It could be worth examining other algorithms in that space, such as random feedback local online (RFLO) learning \cite{roth2018kernel}, which demonstrated good empirical results on simple tasks \cite{marschall2020unified}. One could also investigate sparse RNNs trained with SnAp-n; only SnAp-1 was considered in the current study, as we restricted the latter's scope to dense networks. Proper hyperparameter selection is critical for performance, but grid search is relatively slow, and future studies will benefit from faster and more sophisticated optimization schemes to enhance clinical applicability. SVR with an RBF kernel, which we selected as a classical non-ANN baseline, demonstrated relatively poor performance, possibly due to the associated offline learning setting and independent prediction of outputs. Future studies may benefit from comparison with a stronger benchmark, such as multi-output SVR \cite{tran2024critical}, modeling the correlation between future marker positions, or an online version of SVR \cite{ma2003accurate}. The relatively small size of our dataset was one of the limitations of our research; using larger ones from other institutions \cite{Ernst_2013} or synthesizing breathing motion via generative models \cite{pastor2021semi} will help improve the reliability and generalizability of subsequent works. We restricted ourselves to one minute of training because the shortest time series in our dataset lasts 72s and arbitrarily fixed the cross-validation period; future studies would benefit from assessing how varying the warm-up and cross-validation periods impacts accuracy and robustness to irregular motion. Enhancing the sharp prediction of sudden changes \cite{le2022deep} and tackling prediction interpretability issues \cite{baric2021benchmarking} are other promising avenues in this field. In addition, further research is needed to evaluate the combined tumor tracking error, which arises from both forecasting the surrogate signal and inferring the tumor position from marker locations via a correspondence model. In this study, we could only assess the first type of error. Finally, investigating the resulting decrease in the dose delivered to healthy tissues surrounding the target would help fully assess the clinical impact of state-of-the-art forecasting algorithms in respiratory motion management.

\section{Conclusions} 

In this work, we assessed the capabilities of several online learning algorithms for RNNs to forecast the positions of external markers on the chest and abdomen for lung cancer robotic radiosurgery. Our study is the first to evaluate the performance of SnAp-1 and DNI in that context, to the best of our knowledge. Such prediction methods can compensate for the latency of radiotherapy treatment systems caused by image acquisition, data processing, and radiation beam delivery, thereby decreasing irradiation to healthy tissues. That will, in turn, reduce the risk of side effects, such as radiation pneumonitis or pulmonary fibrosis, induced by the treatment. Although performance comparison with the literature is complex due to the variety of datasets and training settings in previous works, we found that RNNs trained online had a similar or better accuracy than most neural networks previously investigated. Indeed, SnAp-1 achieved mean nRMSEs equal to 0.335 and 0.157 when forecasting respiratory traces sampled at 3.33Hz and 10Hz, respectively, and UORO reached a mean nRMSE of 0.086 with 30Hz signals. Linear regression attained similar or better performance than RNNs when $h$ was low relative to $f$, as evidenced, for instance, by its low nRMSE, equal to 0.098, at $h=0.1\text{s}$ and $f=10\text{Hz}$. Those values correspond to averages over the selected horizons $h \leq 2.1\text{s}$ and the nine time series in our dataset, each comprised of the 3D positions of three external markers with amplitudes from 6mm to 40mm in the SI direction and lasting from 73s to 222s. These relatively low errors were attained despite the relatively high prevalence of irregular respiratory records within our dataset and the low amount of training data that we used: only one minute from a single subject. By contrast, previous works have typically employed a large database to train algorithms offline.

RNNs trained online can efficiently learn from the most recent incoming data instead of discarding it. In the context of respiratory motion forecasting, these algorithms can capture the latest characteristics of the breathing movements of a particular patient and adapt to unseen irregularities, leading to improved accuracy compared to offline learning approaches. RTRL and UORO have been investigated with that clinical application in mind \cite{mafi2020real, POHL2021101941, pohl2022prediction}, and in this study, we compare them with SnAp-1 and DNI. The latter are alternatives to RTRL with a lower computational cost of $\mathcal{O}(q^2)$, where $q$ is the number of hidden units, equal to that of UORO. In this work, we derive efficient implementations for SnAp-1 and DNI in the case of vanilla RNNs. Specifically, we introduce "compressed" influence and immediate Jacobian matrices without zero entries to reduce the memory requirements and computation time of SnAp-1. Concerning DNI, we propose an improved formula for updating the coefficient matrix $A$ in credit assignment estimation that overcomes the implicit assumptions made in \cite{marschall2020unified} when fitting the synthetic gradient to the true gradient. In general, UORO, SnAp-1, and DNI achieved higher accuracy and time performance than RTRL. DNI's inference time was the lowest among all the RNN algorithms compared; it was equal to 6.8ms per time step at 30Hz, which is approximately 5 times lower than that of RTRL. This is despite RTRL being trained with fewer neurons (up to $q=40$) to compensate for its higher complexity, $\mathcal{O}(q^4)$, whereas we considered values of $q$ up to 180 for DNI in the grid search process. Some previous works examined dynamic retraining of ANNs as a method to adjust to the most recent inputs \cite{teo2018feasibility, yun2019deep, lombardo2022offline}. However, such a strategy involves arbitrarily selecting additional hyperparameters (e.g., the window size and number of iterations) and results in forgotten information. In contrast, online learning algorithms leverage the latest data points while retaining knowledge of the past. Future research directions include exploring other fast online learning algorithms for RNNs, selecting hyperparameters more efficiently to reduce the cross-validation computing time, examining online learning specialization of population models trained offline to enhance accuracy, reliability, and robustness to unsteady breathing patterns, and validating the proposed method with more clinical data.

\section*{Acknowledgments}

We thank Prof. Masaki Sekino, Prof. Ichiro Sakuma, and Prof. Hitoshi Tabata (The University of Tokyo, Graduate School of Engineering) for their insightful comments that helped improve the quality of this research. We also thank Dr. Christian Le Minh (Max Planck Institute) and Mr. Suryanarayanan N.A.V. (The University of Tokyo, Graduate School of Engineering), who provided help regarding software. We also thank Dr. Jonathan Cullen (Brainomix Limited) and Dr. Stephen Wells (Nikon), who helped proofread the article.

\section*{Ethical approval}

The authors did not perform experiments involving human participants or animals.

\section*{Funding} 

This research has not received any specific grant from public, commercial, or not-for-profit funding agencies.

\section*{Declaration of competing interests}

The authors declare that they have no conflict of interest.

\section*{Code and data availability}

The code and dataset used are both publicly available \cite{pohl2024MRforecastingcode}.

\bibliographystyle{spbasic}      
\bibliography{bibliography}

\appendix

\section{Appendix: Notes on the derivation of SnAp-1 for standard RNNs}%
\label{appendix: SnAp-1}

The general derivation of SnAp-1 is outlined in \cite{menick2020practical}. In this section, we explain in detail how "compressed" immediate Jacobian and influence matrices can be introduced in the implementation of SnAp-1 for standard RNNs defined in Eqs. \ref{eq:state_vanilla} and \ref{eq:measurement_vanilla}, leading to a reduction of its complexity down to $\mathcal{O}(q^2)$. Furthermore, we delve into specifics regarding various quantities appearing in the computation of the loss gradient $\nabla_{\theta} L_n$ in SnAp-1. Notably, the update of the parameters $W_{c,n}$ in line 21 in Algorithm \ref{alg:RNN-SnAp-1} is the same as in UORO and is described in Appendix A.2. in \cite{pohl2022prediction}.

\subsection{Influence matrix update}

In SnAp-1, it is hypothesized that the influence matrix update is governed primarily by the diagonal of the dynamic matrix $D_n = ({\partial{F_{\text{st}}}}/{\partial x})(x_n, u_n, \theta_n)$. Therefore, the latter is replaced with the matrix $\overline{D_n}$ containing its diagonal elements only, which makes the recursive computation of the influence matrix faster (Eq. \ref{eq:influence update SnAp-1}).

We define the following matrix for $j \in \{ 1, ..., q \}$:
\begin{equation} \label{eq: def partial F_st over partial W_a,n^j}
\frac{\partial F_{\text{st}}}{\partial W_{a,n}^j} = 
\left[ \frac{\partial F_{\text{st}}}{\partial W_{a,n}^{1,j}}, ..., \frac{\partial F_{\text{st}}}{\partial W_{a,n}^{q,j}} \right]
\end{equation}
and similarly, for $j \in \{ 1, ..., m+1 \}$:
\begin{equation} \label{eq: def partial F_st over partial W_b,n^j}
\frac{\partial F_{\text{st}}}{\partial W_{b,n}^j} = 
\left[ \frac{\partial F_{\text{st}}}{\partial W_{b,n}^{1,j}}, ..., \frac{\partial F_{\text{st}}}{\partial W_{b,n}^{q,j}} \right]
\end{equation}
Eq. 48 in Appendix A.3. of \cite{pohl2022prediction} can be rewritten, for $j \in \{ 1, ..., q \}$, as:
\begin{equation}
\frac{\partial F_{\text{st}}}{\partial W_{a,n}^j} = x_{n,j} \text{Diag}(\Phi'(z_n))
\end{equation}
Similarly, for $j \in \{ 1, ..., m+1 \}$, we also have:
\begin{equation}
\frac{\partial F_{\text{st}}}{\partial W_{b,n}^j} = u_{n,j} \text{Diag}(\Phi'(z_n))
\end{equation}

The parameter vector can be decomposed in the following way:
\begin{equation}
\theta_n = [W_{a,n}^{\text{unrolled}}, W_{b,n}^{\text{unrolled}}, W_{c,n}^{\text{unrolled}}]
\end{equation}
where $W_{a,n}^{\text{unrolled}}$, $W_{b,n}^{\text{unrolled}}$, and $W_{c,n}^{\text{unrolled}}$ are line vectors containing the elements of $W_{a,n}$, $W_{b,n}$, and $W_{c,n}$, respectively.
We can thus rewrite the immediate Jacobian matrix as follows:
\begin{align}
\frac{\partial F_{\text{st}}}{\partial \theta}
&= \left[ \frac{\partial F_{\text{st}}}{\partial W_{a,n}^{\text{unrolled}}}, \frac{\partial F_{\text{st}}}{\partial W_{b,n}^{\text{unrolled}}}, 0_{q \times pq} \right]  \\
&= \left[ \frac{\partial F_{\text{st}}}{\partial W_{a,n}^1}, ..., \frac{\partial F_{\text{st}}}{\partial W_{a,n}^q}, 
\frac{\partial F_{\text{st}}}{\partial W_{b,n}^1}, ..., \frac{\partial F_{\text{st}}}{\partial W_{b,n}^q}, 0_{q \times pq} \right] \\
&= \Big[ x_{n,1} \text{Diag}(\Phi'(z_n)), ..., u_{n,m+1} \text{Diag}(\Phi'(z_n)), 0_{q \times pq} \Big] \label{eq: standard RNN immediate Jacobian}
\end{align}
We have just proved Eq. \ref{eq: standard RNN immediate Jacobian short version}. Since the influence matrix is initialized to $0_{q \times |W|}$, we can show by recursion, using the latter equation and Eq. \ref{eq:influence update SnAp-1}, that it has the form:
\begin{equation} \label{eq:influence matrix shape}
\frac{\partial x_n}{\partial \theta} = \Big[ \text{Diag} (j_{n,1}), ..., \text{Diag} (j_{n,m+q+1}), 0_{q \times pq} \Big]
\end{equation}
where for $k \in \{1, ..., m+q+1 \}$, $j_{n,k}$ is a column vector of size $q$.
We then respectively define the compressed influence and immediate Jacobian matrices, $J_n$ and $I_n$, both of size $q \times (m+q+1)$, as follows: 
\begin{equation}
J_n = \Big[ j_{n,1}, ..., j_{n,m+q+1} \Big]
\end{equation}
\begin{align}
I_n 
&= \Big[ x_{n,1} \Phi'(z_n), ..., x_{n,q} \Phi'(z_n), u_{n,1} \Phi'(z_n), ..., u_{n,m+1} \Phi'(z_n) \Big] \\
&= \Phi'(z_n) [x_n^T, u_n^T] \label{compact immediate jacobian definition}
\end{align}

Under the assumption of SnAp-1 and the standard RNN setting, the formula governing the recursive update of the influence matrix (Eq. \ref{eq:influence update SnAp-1}) involves matrices that contain at most one non-zero element per column. In this work, to improve computational efficiency, we rewrite that equation using the non-sparse matrices $I_n$ and $J_n$ defined above:
\begin{equation}
J_{n+1} = \overline{D_n} J_n + I_n
\end{equation}
We have just proved Eq. \ref{eq: reduced influence update}. The recursive update of the influence matrix is used in the latter form in line 27 of Algorithm \ref{alg:RNN-SnAp-1}.

\subsection{Simplified dynamic matrix}

In this section, we focus on the explicit formulation of $\overline{D_n}$. Using Eq. \ref{eq:state_vanilla}, we can write:
\begin{equation} \label{eq:chain rule with partial F_st over partial x}
\frac{\partial F_{\text{st}}}{\partial x} = \frac{\partial \Phi}{\partial z} \frac{\partial z_n}{\partial x}
\end{equation}
The left and right factors can be directly calculated using the definition of $\Phi$ in Eq. \ref{eq:non_linearity}:
\begin{equation}
\frac{\partial F_{\text{st}}}{\partial x} 
= \begin{bmatrix}
   \phi'(z_{n,1}) & & 0\\ 
   & \ddots & \\
  0 &  &  \phi'(z_{n,q}) 
 \end{bmatrix} W_{a,n} \label{eq:partial F_st over partial x}
\end{equation} 
Consequently:
\begin{align}
\overline{D_n} 
&= \text{Diag}\left( \frac{\partial F_{\text{st}}}{\partial x} \right) \\
&= \begin{bmatrix}
   \phi'(z_{n,1}) & & 0\\ 
   & \ddots & \\
  0 &  &  \phi'(z_{n,q}) 
 \end{bmatrix}
\text{Diag} \Big( W_{a,n} \Big) \\
&= \begin{bmatrix}
    \phi'(z_{n,1}) W_{a,n}^{1,1}& & 0\\
    & \ddots & \\
    0 & &\phi'(z_{n,q}) W_{a,n}^{q,q}
  \end{bmatrix}
\end{align}
The latter equation corresponds to line 25 in Algorithm \ref{alg:RNN-SnAp-1}. In addition, line 17 in Algorithm \ref{alg:RNN-DNI} directly comes from Eq. \ref{eq:partial F_st over partial x}.

\subsection{Loss gradient calculation}

Here, we focus on calculating the loss gradient with respect to the parameters $W_{a,n}$ and $W_{b,n}$. The loss gradient can be calculated as:
\begin{align}
\frac{\partial L_{n+1}}{\partial \theta} &= \frac{\partial L_{n+1}}{\partial x} \frac{\partial x_{n+1}}{\partial \theta} \\
&= \frac{\partial L_{n+1}}{\partial x} [ \text{Diag} (j_{n+1,1}), ..., \text{Diag} (j_{n+1,m+q+1}), 0_{q \times pq} ] 
\end{align}
where we used Eq. \ref{eq:influence matrix shape} to replace ${\partial x_{n+1}}/{\partial \theta}$ within the second line.
We define:
\begin{equation}\label{eq: theta_n^ab definition}
\theta_n^{ab} = [W_{a,n}^{\text{unrolled}}, W_{b,n}^{\text{unrolled}}] = [(\theta_n)_1, ..., (\theta_n)_{|W_a|+|W_b|}]
\end{equation} 
The loss gradient with respect to $W_{a,n}$ and $W_{b,n}$ can then be expressed as:
\begin{align}
\frac{\partial L_{n+1}}{\partial \theta^{ab}} &= \frac{\partial L_{n+1}}{\partial x} \Big[ \text{Diag} (j_{n+1,1}), ..., \text{Diag} (j_{n+1,m+q+1}) \Big]
\end{align}
The right factor in the right-hand side of the latter equation is a matrix containing many zeros; we can rewrite the product above using a non-sparse matrix instead, to improve time performance, as follows:
\begin{equation}
\frac{\partial L_{n+1}}{\partial \theta^{ab}} = \text{reshape} \Big( \nabla_x L_{n+1} * J_{n+1}, 1 \times q(m+q+1) \Big)
\end{equation}
That equation corresponds to line 28 in Algorithm \ref{alg:RNN-SnAp-1}. In that formula, the element-wise multiplication operator $*$ was extended to the product of a column vector $v$ of size $q$ and a matrix $J$ of size $q \times (m+q+1)$ by defining $v * J = [v,...,v] * J$ (i.e., $v$ is repeated $m+q+1$ times). It is shown in Appendix A.1. in \cite{pohl2022prediction} that $\nabla_x L_{n+1} = - W_{c,n}^T e_{n+1}$, which corresponds to line 22 in Algorithm \ref{alg:RNN-SnAp-1}.

\section{Appendix: Notes on the derivation of DNI for standard RNNs}%
\label{appendix:DNI_alg}

The theoretical background underlying DNI and its implementation for general neural networks are laid out in \cite{jaderberg2017decoupled}. Further explanations concerning the case of standard RNNs can be found in \cite{marschall2020unified}. This section complements the description of \citeauthor{marschall2020unified} by providing an improved expression for the gradient of $\|f(A)\|^2$, where $A$ is the coefficient matrix intervening in credit assignment prediction (Eq. \ref{eq: credit assignment proportionality relationship}). Furthermore, we derive here some of the formulas appearing in Algorithm \ref{alg:RNN-DNI} and discuss aspects related to time complexity. 

\subsection{Derivation of the gradient of $\|f(A)\|^2$}
\label{appendix: derivation of the gradient of norm(f(A)) squared in DNI}

We seek to compute $\partial{\|f(A)\|^2} / \partial{A}$ where:
\begin{align*}
  f \colon & \mathbb{R}^{p+q+1} \times \mathbb{R}^q \to \mathbb{R}^q \\
           & A \mapsto \tilde{x}_n A - \nabla_x L_{n+1}^T - \tilde{x}_{n+1} A D_n
\end{align*}

We select $(i,j) \in \{ 1, \ldots, p+q+1 \} \times \{ 1, \ldots, q \}$. We fix all the elements of $A$, except that with indices $(i,j)$, and consider the function $f_{i,j} \colon A_{i,j} \in \mathbb{R} \mapsto f(A)$. We have:
\begin{align}
\frac{1}{2}\frac{\partial \| f(A) \|^2}{\partial A_{i,j}}  &= \frac{1}{2} \left( \| \cdot \|^2 \circ f_{i,j} \right)' (A_{i,j})\\
&= \frac{1}{2} \left\langle \nabla (\| \cdot \|^2) (f_{i,j}(A_{i,j})) \,,\, f_{i,j}'(A_{i,j}) \right\rangle \\
& = \left\langle f_{i,j}(A_{i,j}) \,,\, f_{i,j}'(A_{i,j}) \right\rangle \\
& = \left\langle f(A) \,,\, f_{i,j}'(A_{i,j}) \right\rangle \label{eq: gradient norm squared of f(A)}
\end{align}
where $\left\langle \cdot \,,\, \cdot \right\rangle$ denotes the inner product operator.
We consider $k \in \{ 1, \ldots, q \}$. The $k^{\text{th}}$ component of $f_{i,j}(A_{i,j}) = f(A)$ is:
\begin{align}
f_{i,j}(A_{i,j})_k &= \sum_{u=1}^{p+q+1} (\tilde{x}_n)_u A_{u,k} - \left(\nabla_x L_{n+1}^T\right)_k \nonumber \\
& \phantomrel{=} \hphantom{f_{i,j}(A_{i,j})} - \sum_{u=1}^{p+q+1} \sum_{v=1}^{q}  (\tilde{x}_{n+1})_u A_{u,v} (D_n)_{v,k} 
\end{align}
Applying differentiation, we obtain:
\begin{align}
f_{i,j}'(A_{i,j})_k &= 1(k = j) (\tilde{x}_n)_i - (\tilde{x}_{n+1})_i (D_n)_{j,k} 
\end{align}
Therefore:
\begin{align}
f_{i,j}'(A_{i,j}) &= \left[ 0, ..., 0,(\tilde{x}_n)_i, 0, ..., 0 \right] - (\tilde{x}_{n+1})_i (D_n)_{j,\cdot} \label{eq:f_{i,j}'(A_{i,j})}
\end{align}
where $(\tilde{x}_n)_i$, the only non-zero element of the left (vector) term, is located at its $j^{\text{th}}$ position, and $(D_n)_{j,\cdot}$ denotes the $j^{\text{th}}$ row of the dynamic matrix $D_n$.
We obtain the following by replacing $f_{i,j}'(A_{i,j})$ in Eq. \ref{eq: gradient norm squared of f(A)} with its expression in Eq. \ref{eq:f_{i,j}'(A_{i,j})}:
\begin{equation}\label{eq: gradient norm squared of f(A) with f'i,j replaced} 
\frac{1}{2}\frac{\partial \| f(A) \|^2}{\partial A_{i,j}} 
= \left\langle f(A) \,,\, 
\left[ 0, ..., 0,(\tilde{x}_n)_i, 0, ..., 0 \right] - (\tilde{x}_{n+1})_i (D_n)_{j,\cdot} 
\right\rangle 
\end{equation}
The latter equation corresponds to Eq. 26 in \cite{marschall2020unified}, where it was implicitly assumed that the contribution of $D_n$ as a second term on the right side of the inner product was equal to zero. We can develop the right-hand side of Eq. \ref{eq: gradient norm squared of f(A) with f'i,j replaced} as follows:
\begin{align}
\frac{1}{2}\frac{\partial \| f(A) \|^2}{\partial A_{i,j}} 
&= \left\langle f(A) \,,\, \left[ 0, ..., 0, (\tilde{x}_n)_i, 0, ..., 0 \right] \right\rangle \nonumber\\
& \phantomrel{=} \hphantom{\frac{1}{2}\frac{\partial \| f(A) \|^2}{\partial A_{i,j}} } - \left\langle f(A) \,,\,  (\tilde{x}_{n+1})_i (D_n)_{j,\cdot}  \right\rangle \\
&= (\tilde{x}_n)_i f(A)_j - (\tilde{x}_{n+1})_i f(A) (D_n^T)_{\cdot, j}
\end{align}
The latter equation is the same as Eq. \ref{eq:final expression of grad(norm(f(A)))}, which we have just proved.

\subsection{Efficient computation of $\Delta_{\theta^{ab}} L_{n+1}$}
\label{appendix: derivative of loss with respect to non-output weights}

Eq. \ref{eq: DNI gradient loss} can be rewritten as:
\begin{align}
\frac{\partial L_{n+1}}{\partial \theta^{ab}} &= c_n \frac{\partial F_{\text{st}}}{\partial \theta^{ab}}(x_n, u_n, \theta_n) 
\end{align}
where $\theta_n^{ab}$ is defined in Eq. \ref{eq: theta_n^ab definition}. The computation of this product takes $q (m+q+1)$ multiplications. In other words, its complexity is the same as that of DNI, $\mathcal{O}(q^2)$. However, computational speed can be further improved in practice by rewriting that equation using non-sparse matrices. Indeed, using Eq. \ref{eq: standard RNN immediate Jacobian}, we can write:
\begin{align}
\frac{\partial L_{n+1}}{\partial \theta^{ab}} &= c_n \Big[ x_{n,1} \text{Diag}(\Phi'(z_n)), ..., u_{n,m+1} \text{Diag}(\Phi'(z_n)) \Big] \\
&= \Big[ x_{n,1} c_n \text{Diag}(\Phi'(z_n)), ..., u_{n,m+1} c_n \text{Diag}(\Phi'(z_n)) \Big] \label{eq: gradient loss derivative with respect to thetaab DNI}
\end{align}
The common factor in each block can be rewritten as follows:
\begin{align}
c_n \text{Diag}(\Phi'(z_n)) &= \Big[ (c_n)_1 \Phi'(z_n)_1, ..., (c_n)_q \Phi'(z_n)_q \Big] \\
&= c_n * \Phi'(z_n)^T \\
&= \varphi_n^T
\end{align}
where we defined the following auxiliary column vector:
\begin{equation}
\varphi_n = c_n^T * \Phi'(z_n) \in \mathbb{R}^q
\end{equation}
Therefore, we can rewrite Eq. \ref{eq: gradient loss derivative with respect to thetaab DNI} as follows:
\begin{align}
\frac{\partial L_{n+1}}{\partial \theta^{ab}} &= \Big[ x_{n,1} \varphi_n^T, ..., x_{n,q} \varphi_n^T, u_{n,1} \varphi_n^T, ..., u_{n,m+1} \varphi_n^T \Big] \\
&= \text{reshape($\varphi_n [x_n^T, u_n^T]$, $1 \times q(m+q+1)$)}
\end{align}
which corresponds to line 24 in Algorithm \ref{alg:RNN-DNI}.

\subsection{Influence of matrix multiplication order on time complexity}

In our implementation of DNI, the matrix multiplications in the expressions of $f(A)$ and $\Delta A$ in lines 19 and 20 of Algorithm \ref{alg:RNN-DNI} need to be computed in the order indicated by the brackets in the formulas below:
\begin{equation}
f(A_n) = \tilde{x}_n A_n - \nabla_x L_{n+1}^T - \left[\tilde{x}_{n+1} A_n\right] D_n
\end{equation}
\begin{equation}
\Delta{A} = \tilde{x}_n^T f(A_n) - \tilde{x}_{n+1}^T \left[f(A_n) D_n^T\right]
\end{equation}
Indeed, using an alternative multiplication order for the successive products (i.e., attempting to compute $\tilde{x}_{n+1} [A_n D_n]$ or $[\tilde{x}_{n+1}^T f(A_n)] D_n^T$) would lead to an overall higher time complexity $\mathcal{O}(q^3)$.

\flushcolsend
\onecolumn

\section{Appendix: Resampling the original 10Hz signal}\label{appendix:resampling process}

\begin{figure*}[hb!]
	\captionsetup[subfloat]{justification=raggedright} 
    \centering
    \subfloat[Original signal (10 Hz)]{\includegraphics[width=.23\textwidth]{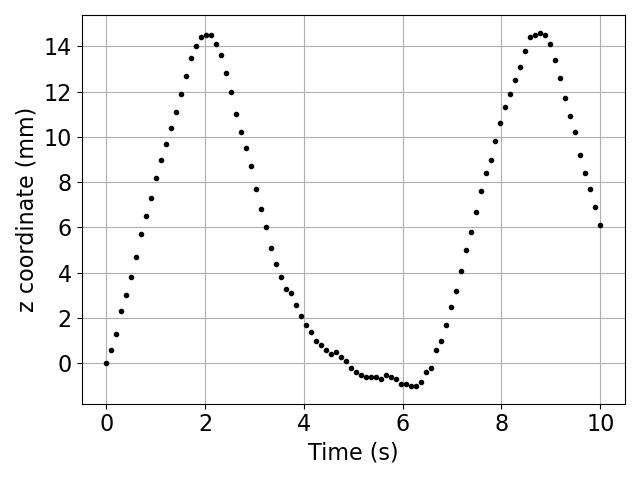}}%
    \:
    \subfloat[Downsampling to 3.33Hz]{\includegraphics[width=.23\textwidth]{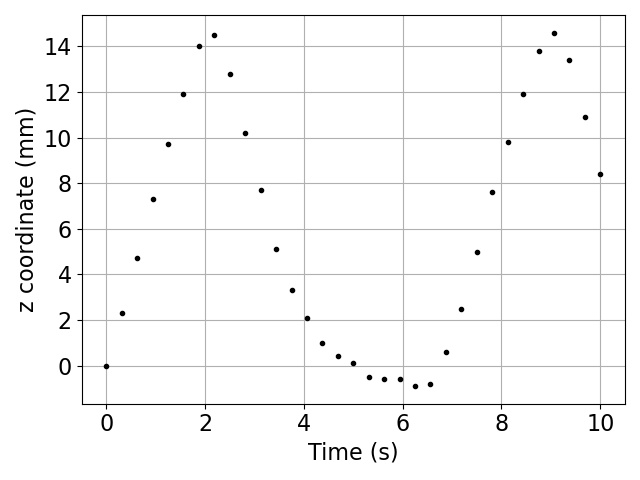} }%
	\:    
    \subfloat[Upsampling to 30Hz (first step): cubic spline interpolation]{\includegraphics[width=.23\textwidth]{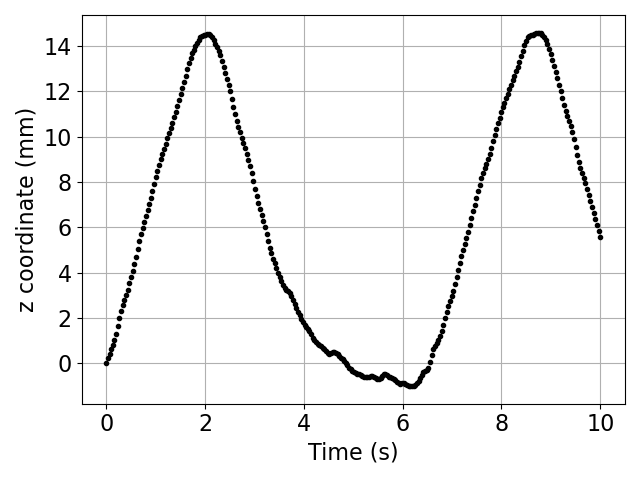} }%
	\:    
    \subfloat[Upsampling to 30Hz (second step): Gaussian noise addition and truncation of values to 1 decimal place]{\includegraphics[width=.23\textwidth]{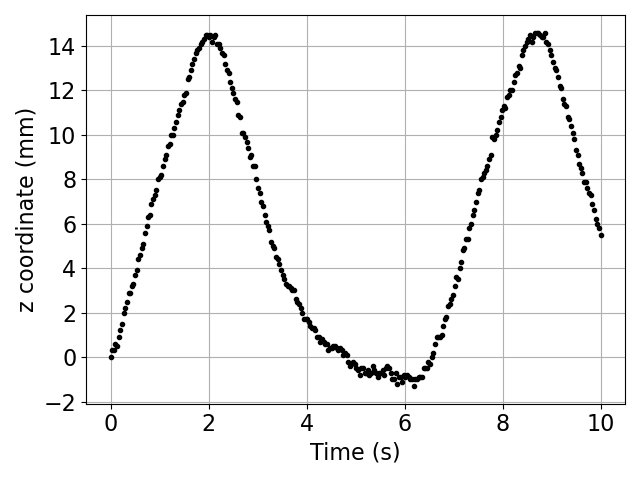} }%
    \vspace{-0.5cm} 
    \caption{Visualization of the resampling process, using the first 10s of the z-coordinate trajectory (axial direction) of marker 1 in sequence 2 as an example. Upsampling the original 10Hz time series involves two steps: interpolation and Gaussian noise addition. The latter simulates sensor noise and local breathing irregularities.}%
    \label{fig:resampling process}%
\end{figure*}

\vspace{-1cm}

\section{Appendix: Influence of the SHL and hidden layer size on computation time}\label{appendix: computation time}

\begin{table*}[htb!]
\small 
\setlength{\tabcolsep}{3pt}
\begin{center}
\begin{tabular}{llllllllll}
\hline
                & \multicolumn{3}{l}{Sampling at 3.33Hz}                 & \multicolumn{3}{l}{Sampling at 10Hz}                   & \multicolumn{3}{l}{Sampling at 30Hz}                       \\[1mm] 
Prediction      & 1.2s SHL              & 6.0s SHL              & Relative& 1.2s SHL              & 6.0s SHL              & Relative& 1.2s SHL              & 6.0s SHL              & Relative  \\
method          &                       &                       & increase&                       &                       & increase&                       &                       & increase  \\
\hline \hline
RTRL            & $2.98 \times 10^{-1}$ & 2.51                  & 7.41    & 2.33                  & 17.5                  & 6.52    & 10.5                  & 56.8                  & 4.40    \\
UORO            & $1.60 \times 10^{-1}$ & $2.81 \times 10^{-1}$ & 0.76    & $3.78 \times 10^{-1}$ & 4.33                  & 10.4    & 1.78                  & 22.1                  & 11.5 \\
SnAp-1          & $9.89 \times 10^{-2}$ & $1.90 \times 10^{-1}$ & 0.93    & $2.37 \times 10^{-1}$ & 3.41                  & 13.4    & 1.24                  & 18.7                  & 14.1    \\
DNI             & $1.19 \times 10^{-1}$ & $1.74 \times 10^{-1}$ & 0.47    & $2.30 \times 10^{-1}$ & 2.39                  & 9.37    & $8.51 \times 10^{-1}$ & 13.4                  & 14.7    \\          
LMS             & $3.83 \times 10^{-3}$ & $7.02 \times 10^{-3}$ & 0.83    & $7.03 \times 10^{-3}$ & $2.30 \times 10^{-2}$ & 2.27    & $1.39 \times 10^{-2}$ & $5.03 \times 10^{-2}$ & 2.63    \\
Linear regression & $4.41 \times 10^{-4}$ & $1.04 \times 10^{-3}$ & 1.36    & $7.04 \times 10^{-4}$ & $5.10 \times 10^{-3}$ & 6.24    & $2.62 \times 10^{-3}$ & $2.41 \times 10^{-2}$ & 8.22    \\                   
Kernel SVR      & $1.61 \times 10^{-1}$ & $2.18 \times 10^{-1}$ & 0.36    & $2.53 \times 10^{-1}$ & $9.11 \times 10^{-1}$ & 2.60    & 2.08                  & 16.5                  & 6.96    \\                   
\hline
\end{tabular}
\caption{Mean calculation time per time step in milliseconds (13th Gen Intel Core i7-13700 2.10GHz CPU, 16Gb RAM, using MATLAB) for all forecasting algorithms, input signal sampling frequencies, and the two boundary SHLs (1.2s and 6.0s) considered in this study. The relative increase of the computation time, as the SHL increases between those two values, is also provided (as a ratio). Each time period in the table associated with an RNN algorithm represents the inference time averaged over the hidden layer sizes explored during cross-validation, ranging from $q=10$ to $q=40$ for RTRL and from $q=30$ to $q=180$ for the other training methods.}
\label{table:computation time variation with SHL}

\end{center}
\end{table*}

\vspace{-1cm}

\begin{table*}[htb!]
\small 
\setlength{\tabcolsep}{3pt}
\begin{center}
\begin{tabular}{llllllllll}
\hline
           & \multicolumn{3}{l}{Sampling at 3.33Hz}                   & \multicolumn{3}{l}{Sampling at 10Hz}                     & \multicolumn{3}{l}{Sampling at 30Hz}                      \\[1mm] 
Prediction & Few hidden            & Many hidden           & Relative & Few hidden           & Many hidden            & Relative & Few hidden            & Many hidden           & Relative  \\
method     & units                 & units                 & increase & units                & units                  & increase & units                 & units                 & increase  \\
\hline \hline
RTRL       & $2.24 \times 10^{-1}$ & 3.36                  & 14.0     & $6.68 \times 10^{-1}$ & 23.4                  & 34.1     & 3.17                  & 71.4                  & 21.5      \\
UORO       & $5.59 \times 10^{-2}$ & $4.63 \times 10^{-1}$ & 7.28     & $1.30 \times 10^{-1}$ & 6.03                  & 45.6     & $5.22 \times 10^{-1}$ & 26.0                  & 48.8      \\
SnAp-1     & $4.54 \times 10^{-2}$ & $2.87 \times 10^{-1}$ & 5.32     & $9.74 \times 10^{-2}$ & 4.63                  & 46.6     & $3.70 \times 10^{-1}$ & 22.0                  & 58.5      \\
DNI        & $4.10 \times 10^{-2}$ & $3.04 \times 10^{-1}$ & 6.42     & $7.29 \times 10^{-2}$ & 3.51                  & 47.2     & $2.12 \times 10^{-1}$ & 16.3                  & 75.9      \\          
\hline
\end{tabular}
\caption{Mean calculation time per time step in milliseconds (13th Gen Intel Core i7-13700 2.10GHz CPU, 16Gb RAM, using MATLAB) for all RNN algorithms, input signal sampling frequencies, and the two boundary hidden layer sizes considered in this study. "Few hidden units" refers to $q=10$ for RTRL and $q=30$ for the other algorithms, while "many hidden units" refers to $q=40$ for RTRL and $q=180$ for the other algorithms. The relative increase of the computation time, as $q$ increases between those two values, is also provided (as a ratio). Each time period in the table represents the inference time averaged over the SHLs explored during cross-validation, between 1.2s and 6.0s.}
\label{table:computation time variation with q}

\end{center}
\end{table*}

\clearpage
\afterpage{
\scriptsize
\begin{landscape}
\section{Appendix: Comparison of the prediction performance with regular and irregular breathing sequences}\label{appendix:regular vs irregular perf}
\begin{center}
\hvFloat[nonFloat=true, capPos=t, rotAngle=0, objectPos=c]
{table}%
{\begin{tabular}{llllllll}
\hline
          &                   & Sampling              &                       & Sampling              &                       & Sampling              &                     \\
          &                   & at 3.33Hz            &                       & at 10Hz              &                       & at 30Hz              &                     \\
\hline
Error     &  Prediction       & Regular               & Irregular             & Regular               & Irregular             & Regular               & Irregular           \\
type      &  method           & breathing             & breathing             & breathing             & breathing             & breathing             & breathing           \\
\hline \hline
MAE       & RTRL          & $1.2975 \pm 0.0008$   & $1.2421 \pm 0.0007$   & $0.6256 \pm 0.0002$   & $0.5671 \pm 0.0002$   & $0.3555 \pm 0.0001$   & $0.3113 \pm 0.0001$\\
(in mm)   & UORO          & $1.0738 \pm 0.0009$   & $1.2898 \pm 0.0018$   & $0.4649 \pm 0.0002$   & $0.5909 \pm 0.0006$   & $0.2486 \pm 0.0001$   & $0.3342 \pm 0.0001$\\
          & SnAp-1        & $0.9980 \pm 0.0005$   & $1.2090 \pm 0.0012$   & $0.4822 \pm 0.0001$   & $0.4931 \pm 0.0001$   & $0.2862 \pm 0.0001$   & $0.2954 \pm 0.0001$\\
          & DNI           & $1.0296 \pm 0.0018$   & $1.1882 \pm 0.0012$   & $0.5260 \pm 0.0004$   & $0.5572 \pm 0.0004$   & $0.2807 \pm 0.0001$   & $0.3160 \pm 0.0001$\\          
          & LMS               & 1.4491                & 1.7463                & 1.0329                & 1.0336                & 0.6378                & 0.5339\\
          & Linear regression   & 4.2464                & 6.7573                & 3.5464                & 6.8746                & 4.4954                & 3.5473\\
          & No prediction     & 3.2131                & 3.7826                & 2.9874                & 3.5351                & 2.9960                & 0.5931\\
          & RNN with a frozen layer & $1.0122 \pm 0.0038$   & $1.8518 \pm 0.0054$   & $3.2728 \pm 0.0136$   & $2.0693 \pm 0.0066$   & $3.0255 \pm 0.0077$   & $1.3118 \pm 0.0033$\\
          & Kernel SVR    & 2.4862                & 2.9836                & 3.4113                & 2.9765                & 4.10356               & 3.1881             \\
\hline               
RMSE      & RTRL          & $1.7915 \pm 0.0011$   & $1.8015 \pm 0.0009$   & $0.8670 \pm 0.0003$   & $0.8384 \pm 0.0003$   & $0.4739 \pm 0.0001$   & $0.4075 \pm 0.0001$\\
(in mm)   & UORO          & $1.5098 \pm 0.0015$   & $1.9580 \pm 0.0036$   & $0.6311 \pm 0.0005$   & $0.9019 \pm 0.0014$   & $0.3246 \pm 0.0001$   & $0.4323 \pm 0.0002$\\
          & SnAp-1        & $1.4019 \pm 0.0007$   & $1.7923 \pm 0.0023$   & $0.6978 \pm 0.0002$   & $0.7232 \pm 0.0002$   & $0.4007 \pm 0.0001$   & $0.3713 \pm 0.0001$\\
          & DNI           & $1.4415 \pm 0.0027$   & $1.6773 \pm 0.0014$   & $0.7181 \pm 0.0007$   & $0.8104 \pm 0.0007$   & $0.3756 \pm 0.0002$   & $0.3953 \pm 0.0002$\\          
          & LMS               & 2.0329                & 2.4670                & 1.4452                & 1.4630                & 0.8809                & 0.7135 \\
          & Linear regression   & 5.7863                & 10.0626               & 4.8163                & 10.0242               & 6.2560                & 10.1977 \\
          & No prediction     & 4.3707                & 4.8882                & 4.0726                & 4.5813                & 4.0777                & 4.5866 \\
          & RNN with a frozen layer & $1.4816 \pm 0.0067$   & $2.5682 \pm 0.0084$   & $4.3835 \pm 0.0203$   & $2.9608 \pm 0.0104$   & $4.2954 \pm 0.0122$   & $1.7544 \pm 0.0050$\\
          & Kernel SVR    & 3.2524                & 4.0006                & 4.4461                & 3.9904                & 5.3185                & 4.2630             \\
\hline
nRMSE     & RTRL          & $0.44892 \pm 0.00029$ & $0.39203 \pm 0.00018$ & $0.21804 \pm 0.00008$ & $0.18209 \pm 0.00005$ & $0.11728 \pm 0.00003$ & $0.08907 \pm 0.00002$\\
(no unit) & UORO          & $0.40215 \pm 0.00048$ & $0.42236 \pm 0.00068$ & $0.16883 \pm 0.00015$ & $0.19187 \pm 0.00025$ & $0.08706 \pm 0.00004$ & $0.09374 \pm 0.00004$\\
          & SnAp-1        & $0.35136 \pm 0.00017$ & $0.38091 \pm 0.00042$ & $0.17868 \pm 0.00005$ & $0.15549 \pm 0.00004$ & $0.10339 \pm 0.00002$ & $0.08049 \pm 0.00002$\\
          & DNI           & $0.35993 \pm 0.00065$ & $0.36464 \pm 0.00029$ & $0.18075 \pm 0.00016$ & $0.17284 \pm 0.00013$ & $0.09764 \pm 0.00005$ & $0.08618 \pm 0.00004$\\          
          & LMS               & 0.51221               & 0.55174               & 0.34610               & 0.33496               & 0.20570               & 0.16447 \\
          & Linear regression   & 1.48894               & 2.41659               & 1.28496               & 2.38513               & 1.72509               & 2.44359 \\
          & No prediction     & 1.11088               & 1.07891               & 1.03644               & 1.01082               & 1.03736               & 1.01147 \\
          & RNN with a frozen layer & $0.38941 \pm 0.00122$ & $0.58516 \pm 0.00227$ & $0.99497 \pm 0.00425$ & $0.71884 \pm 0.00264$ & $0.92751 \pm 0.00255$ & $0.45238 \pm 0.00134$\\   
          & Kernel SVR    & 0.85002                & 0.87472                & 1.13796                & 0.87857                & 1.35082                & 0.94697             \\
\hline
Max error & RTRL          & $8.219 \pm 0.016$     & $12.191 \pm 0.019$    & $5.081 \pm 0.010$     & $6.823 \pm 0.009$     & $3.364 \pm 0.007$     & $3.603 \pm 0.006$\\
(in mm)   & UORO          & $7.691 \pm 0.025$     & $14.004 \pm 0.043$    & $4.195 \pm 0.010$     & $7.883 \pm 0.023$     & $2.607 \pm 0.005$     & $3.855 \pm 0.011$\\
          & SnAp-1        & $7.306 \pm 0.018$     & $11.849 \pm 0.030$    & $5.818 \pm 0.009$     & $6.460 \pm 0.009$     & $4.043 \pm 0.007$     & $3.136 \pm 0.008$\\
          & DNI           & $7.641 \pm 0.025$     & $11.575 \pm 0.025$    & $5.154 \pm 0.012$     & $6.883 \pm 0.014$     & $3.123 \pm 0.007$     & $2.863 \pm 0.006$\\   
          & LMS               & 10.588                & 13.797                & 9.375                 & 9.231                 & 6.966                 & 5.330 \\
          & Linear regression   & 28.336                & 55.589                & 23.469                & 55.550                & 31.715                & 54.237 \\
          & No prediction     & 15.050                & 18.727                & 14.294                & 18.363                & 14.534                & 18.563 \\
          & RNN with a frozen layer & $7.931 \pm 0.037$     & $13.106 \pm 0.050$    & $16.096 \pm 0.078$    & $14.353 \pm 0.060$    & $19.542 \pm 0.068$    & $8.919 \pm 0.041$\\  
          & Kernel SVR    & 13.878                & 20.092                & 16.630                & 19.678                & 19.501                & 20.875             \\
\hline
Jitter    & RTRL          & $1.0565 \pm 0.0006$   & $1.3502 \pm 0.0008$   & $0.5052 \pm 0.0002$   & $0.6051 \pm 0.0003$   & $0.2324 \pm 0.0001$   & $0.2968 \pm 0.0002$\\
(in mm)   & UORO          & $1.2010 \pm 0.0008$   & $1.4784 \pm 0.0014$   & $0.5253 \pm 0.0002$   & $0.7929 \pm 0.0006$   & $0.2522 \pm 0.0001$   & $0.3505 \pm 0.0002$\\
          & SnAp-1        & $1.4746 \pm 0.0010$   & $1.8900 \pm 0.0023$   & $0.7744 \pm 0.0002$   & $0.7058 \pm 0.0002$   & $0.3944 \pm 0.0001$   & $0.3431 \pm 0.0001$\\
          & DNI           & $1.9130 \pm 0.0035$   & $1.8192 \pm 0.0012$   & $0.8588 \pm 0.0008$   & $0.8867 \pm 0.0006$   & $0.3237 \pm 0.0002$   & $0.2854 \pm 0.0001$\\          
          & LMS               & 2.2535                & 1.9933                & 1.7437                & 1.2848                & 1.1017                & 0.6807 \\
          & Linear regression   & 1.5236                & 2.1768                & 0.6314                & 1.1183                & 0.3345                & 0.5165 \\
          & No prediction     & 1.0113                & 1.3032                & 0.3877                & 0.5043                & 0.2017                & 0.2697 \\
          & RNN with a frozen layer & $1.3514 \pm 0.0078$   & $2.3801 \pm 0.0110$   & $6.3629 \pm 0.0272$   & $3.7041 \pm 0.0134$   & $5.9569 \pm 0.0154$   & $2.4106 \pm 0.0066$\\                            
          & Kernel SVR    & 0.8630                & 1.0831                & 0.3696                & 0.4072                & 0.1566                & 0.1535             \\
\hline
\end{tabular}}
{Performance of each forecasting algorithm for different input signal sampling rates and levels of breathing regularity\protect\footnotemark. Each measure in the table represents the average of a given performance metric of the test set over the nine records and response times $h$ between 0.1s and 2.1s, using the best hyperparameters for each individual sequence and value of $h$. The 95\% confidence intervals for the mean metrics corresponding to the RNNs are computed assuming a Gaussian distribution.} 
{table:regular vs irregular breathing} 
\end{center}

\footnotetext{Same as footnote \ref{footnote: slow motion sequence removed}}

\begin{figure*}[htb!]
	\captionsetup[subfigure]{justification=centering}
    \centering
    \subfloat[RMSE as a function of $h$ \\ (regular breathing, $f=3.33\text{Hz}$)]{\includegraphics[width=.31\textwidth]{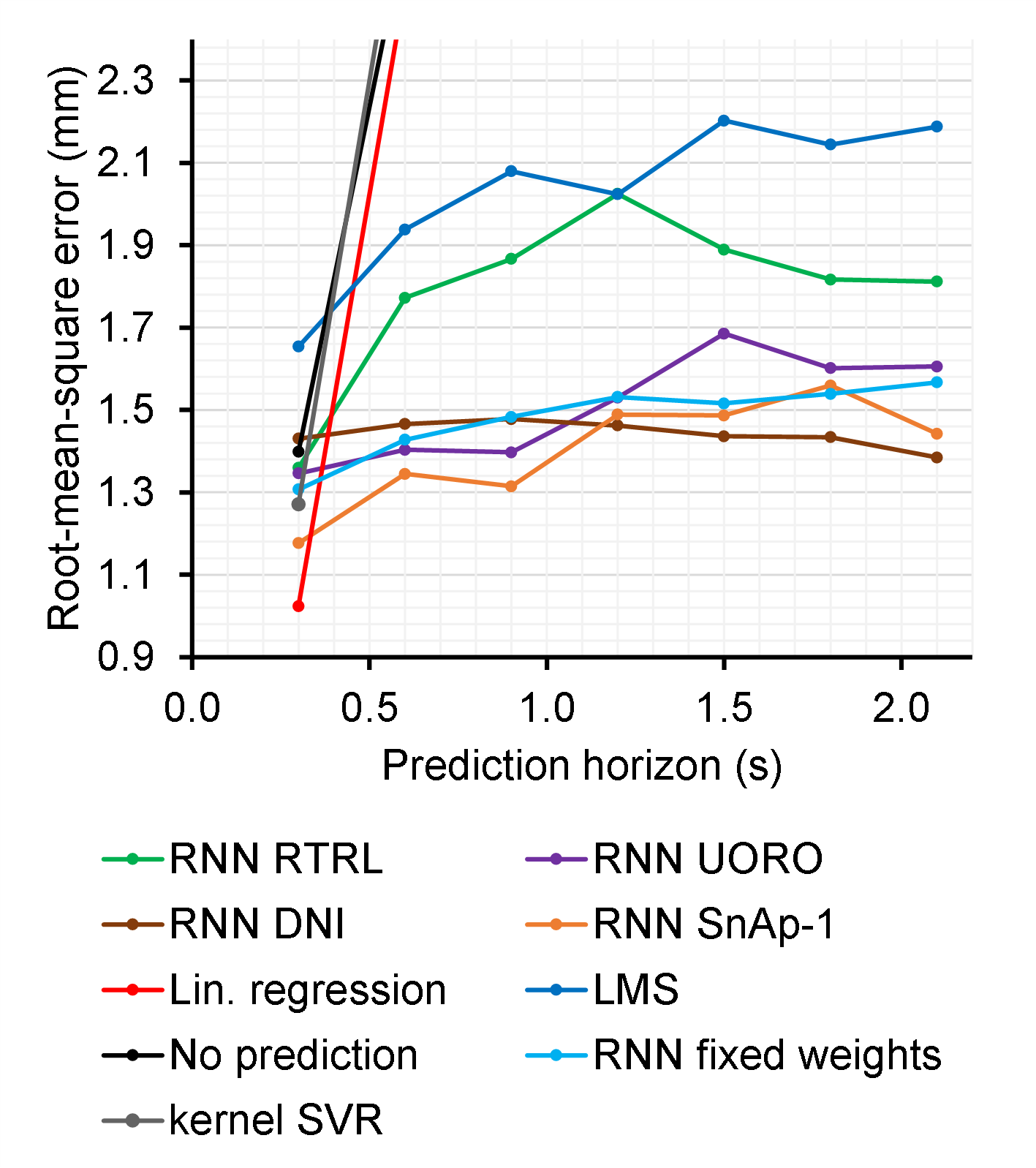}}%
    \subfloat[RMSE as a function of $h$ \\ (irregular breathing, $f=3.33\text{Hz}$)]{\includegraphics[width=.31\textwidth]{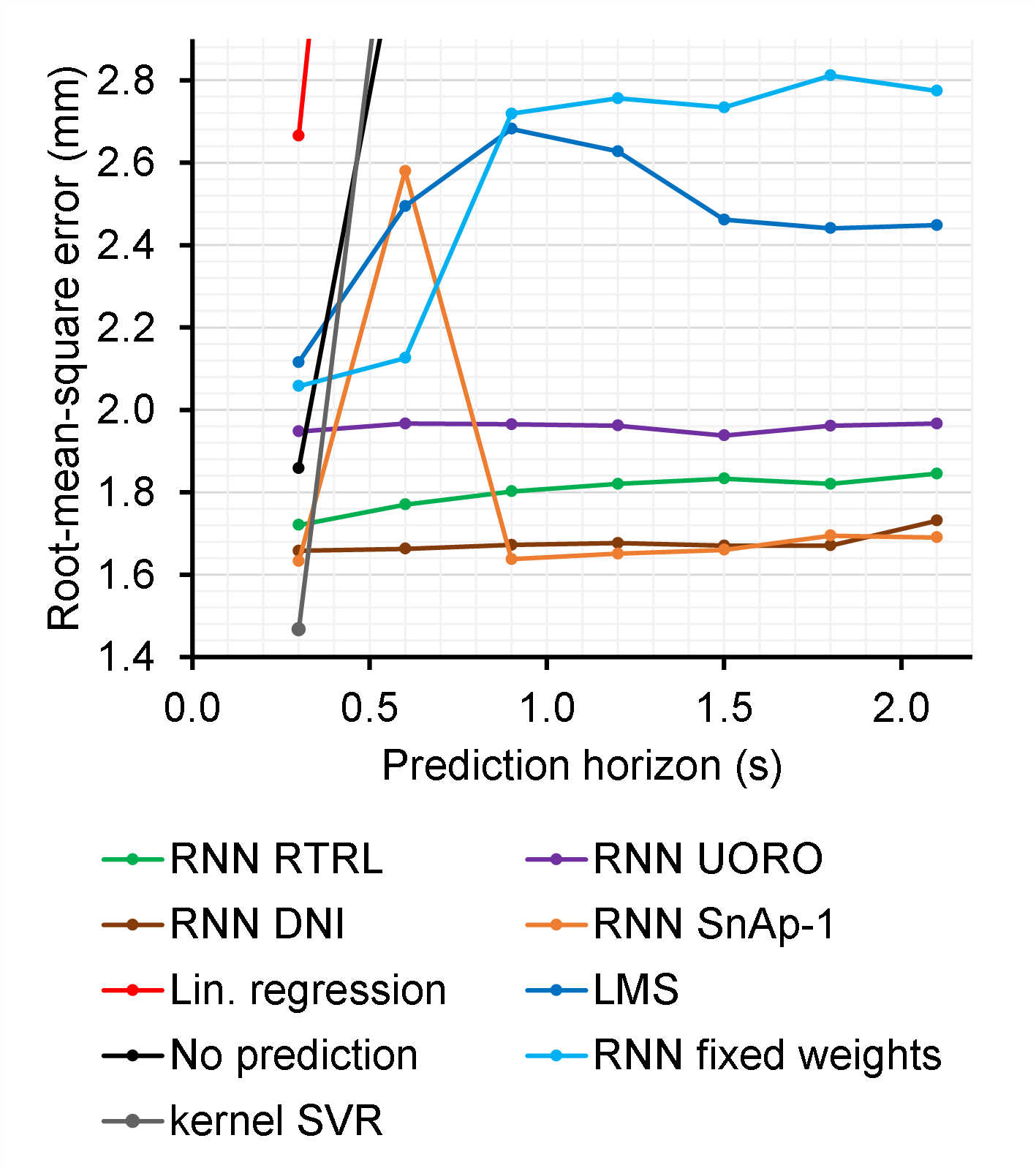} }%
    \subfloat[RMSE as a function of $h$ \\ (regular breathing, $f=10\text{Hz}$)]{\includegraphics[width=.31\textwidth]{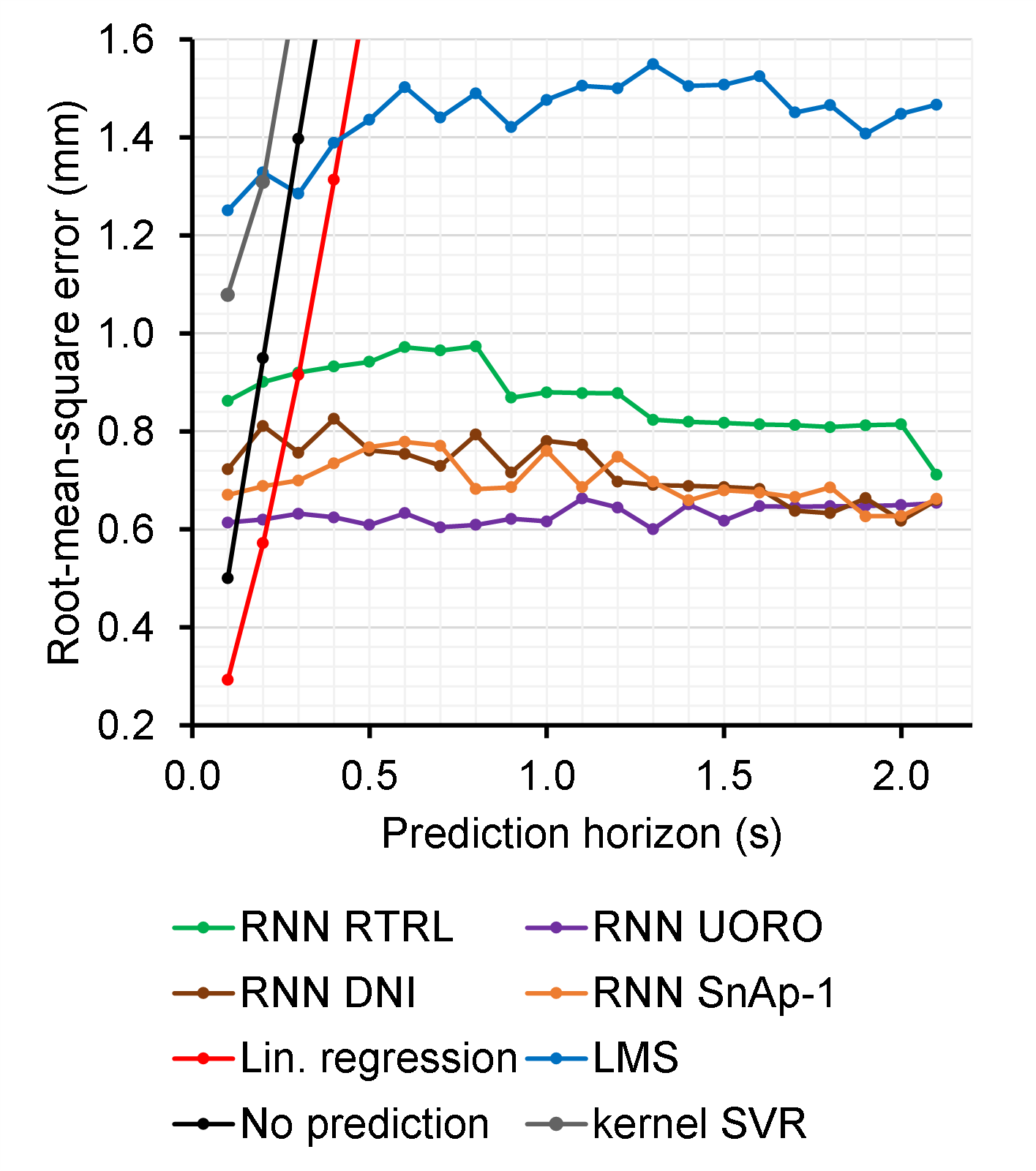} }%
    \subfloat[RMSE as a function of $h$ \\ (irregular breathing, $f=10\text{Hz}$)]{\includegraphics[width=.31\textwidth]{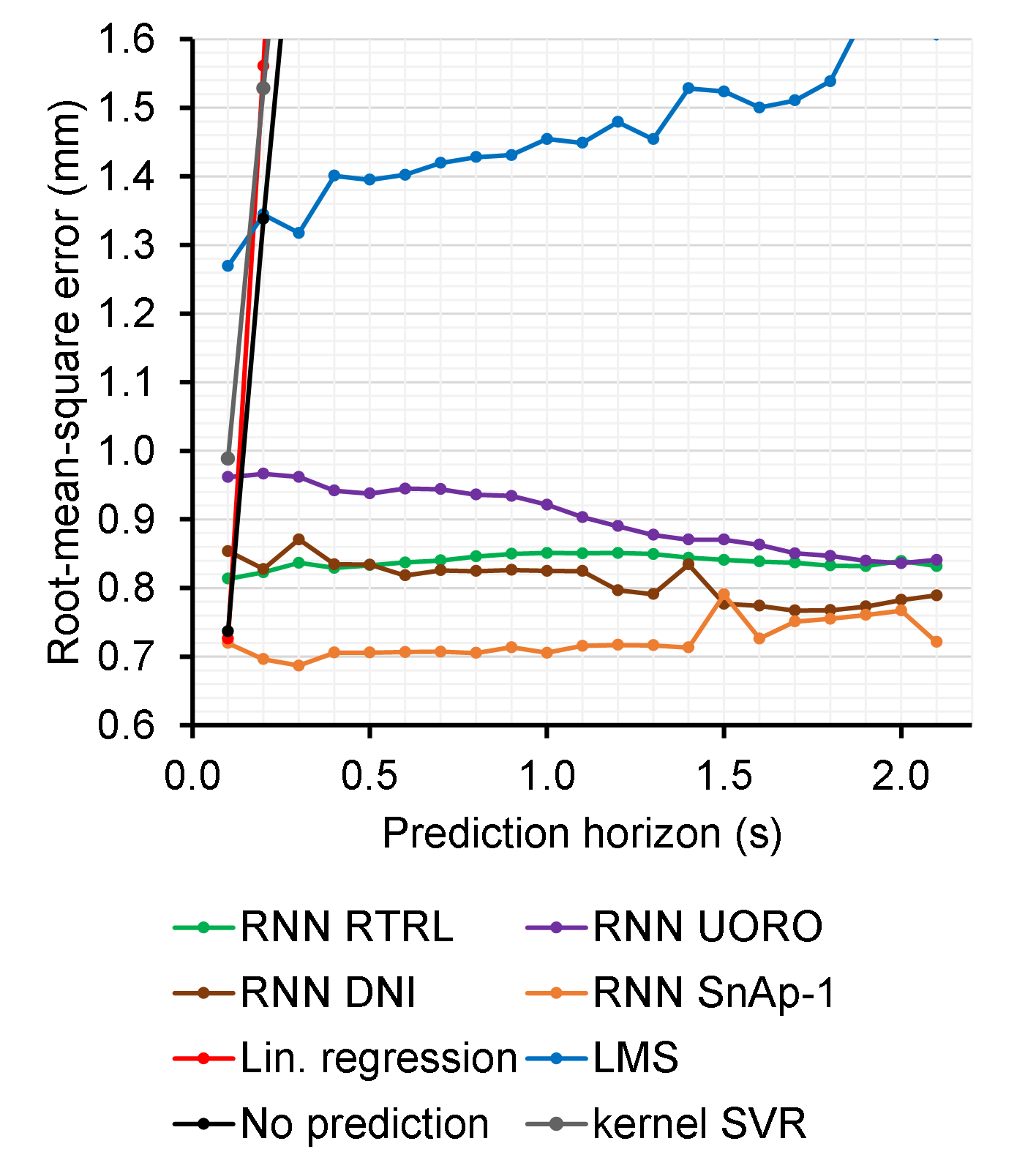} }%
    \quad 
    \subfloat[Maximum error as a function of $h$ \\ (regular breathing, $f=3.33\text{Hz}$)]{\includegraphics[width=.31\textwidth]{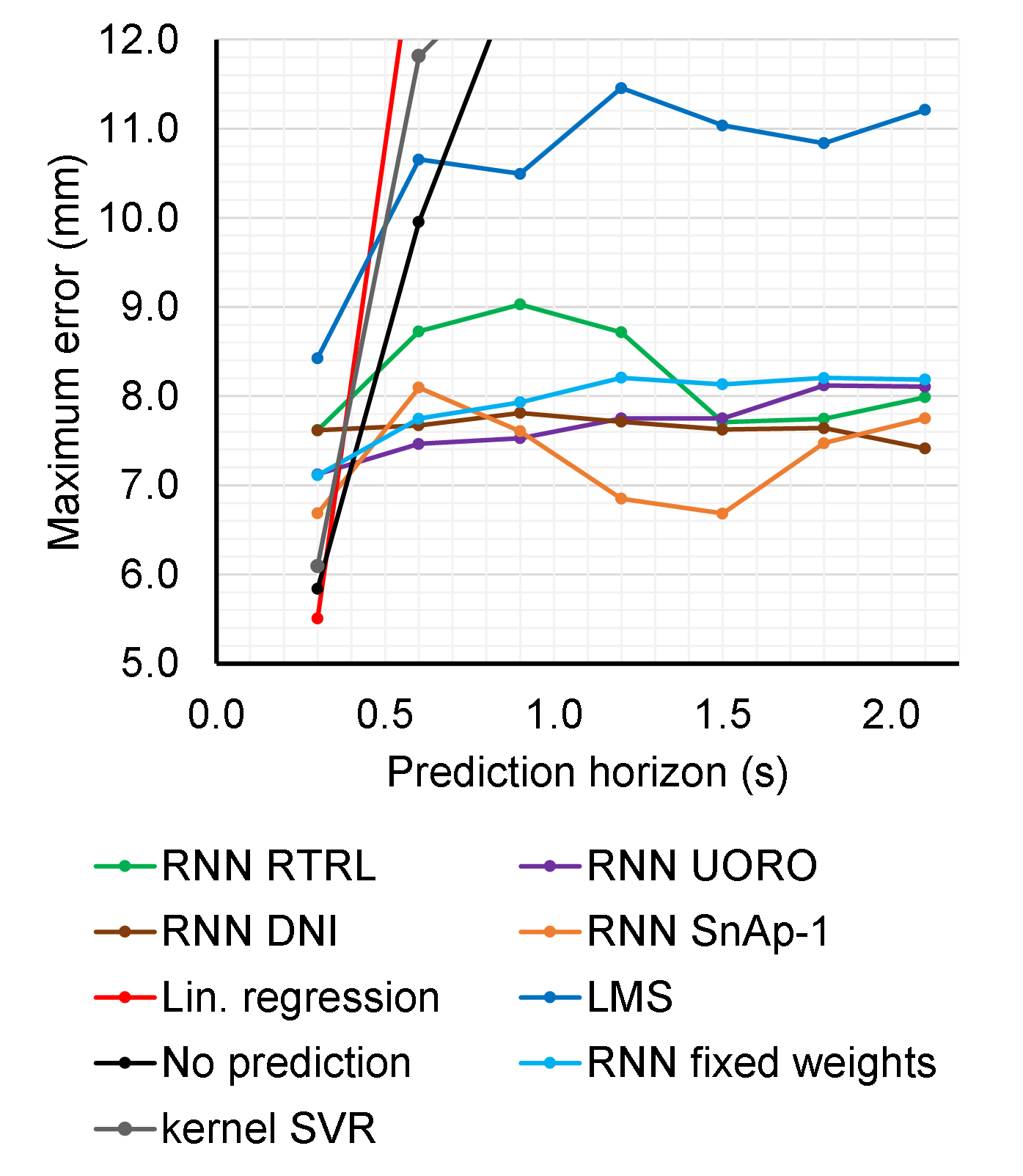}}%
    \subfloat[Maximum error as a function of $h$ \\ (irregular breathing, $f=3.33\text{Hz}$)]{\includegraphics[width=.31\textwidth]{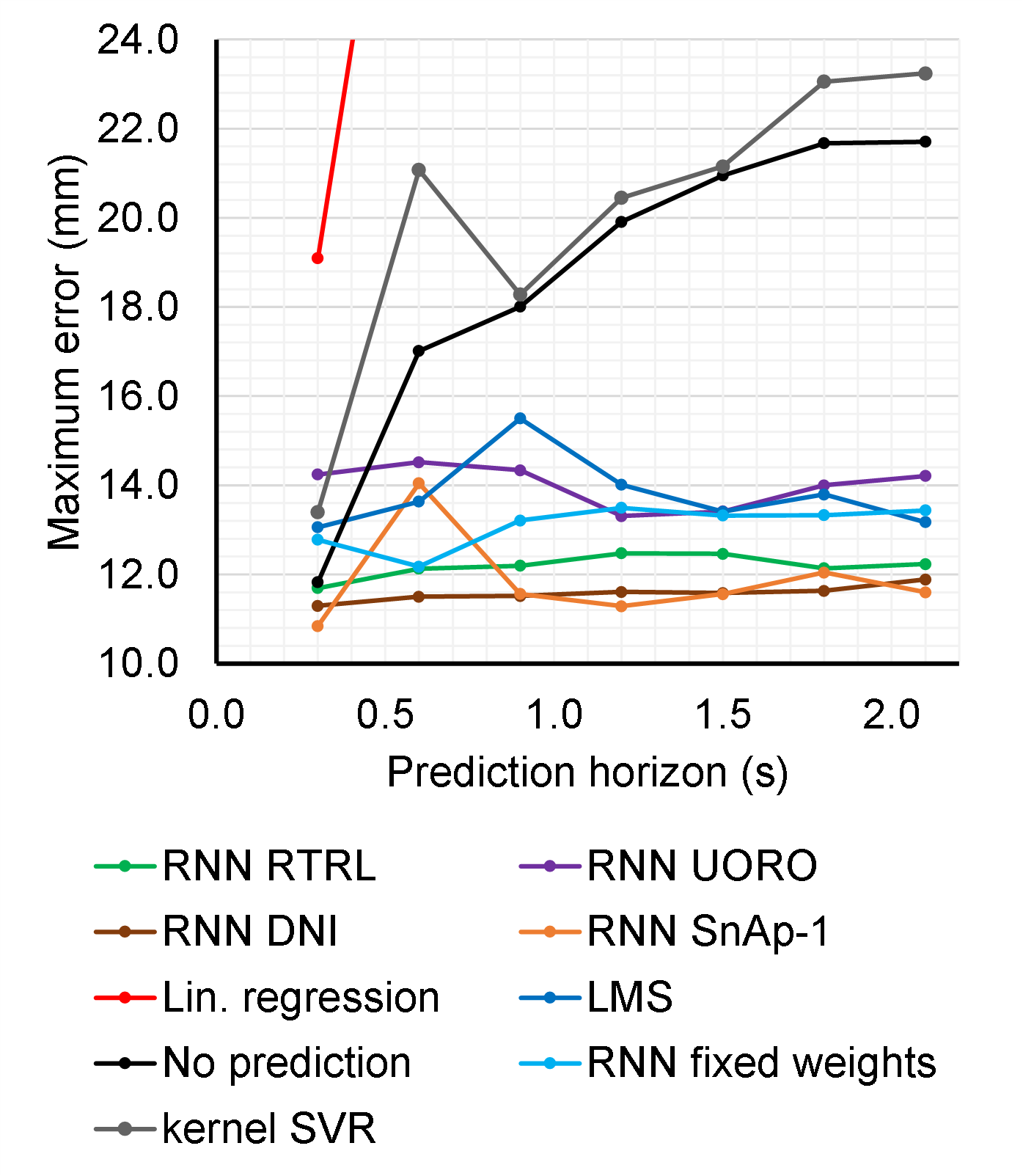} }%
    \subfloat[Maximum error as a function of $h$ \\ (regular breathing, $f=10\text{Hz}$)]{\includegraphics[width=.31\textwidth]{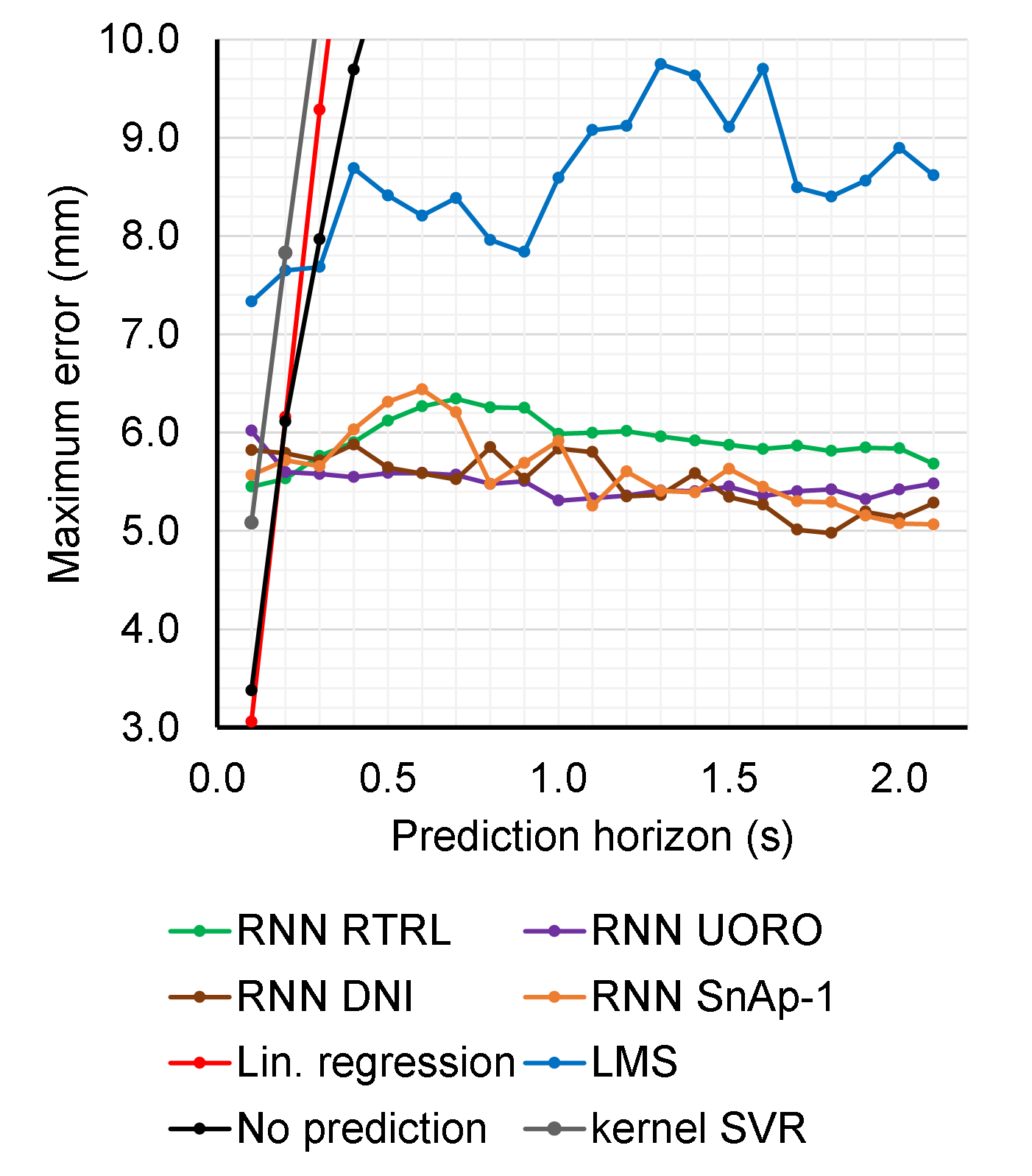} }%
    \subfloat[Maximum error as a function of $h$ \\ (irregular breathing, $f=10\text{Hz}$)]{\includegraphics[width=.31\textwidth]{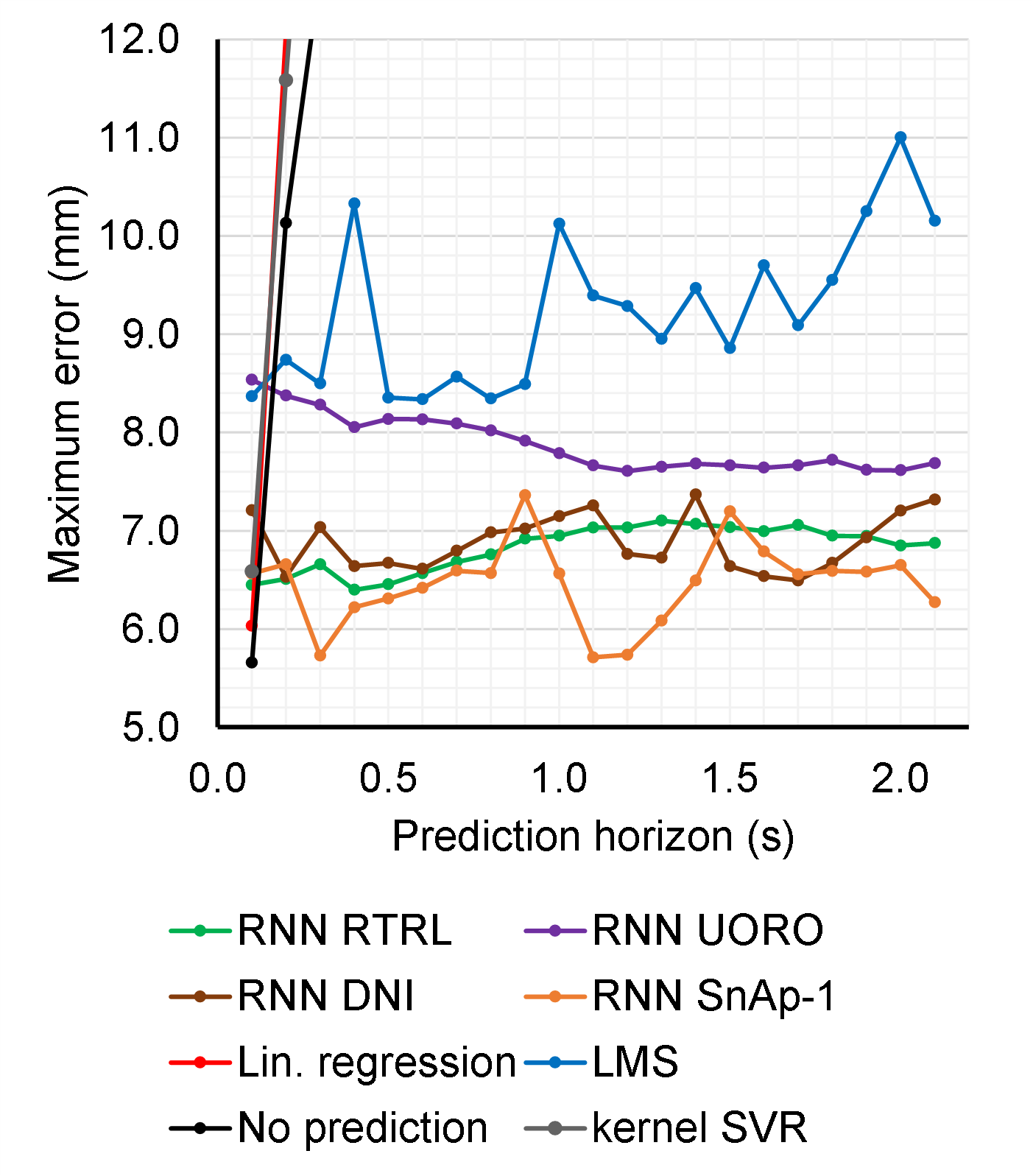} }%
    \caption{RMSE and maximum error of each algorithm as a function of $h$ at 3.33Hz and 10Hz. Each point represents the average error of the test set for a given horizon across the sequences corresponding to either regular or irregular breathing\protect\footnotemark, using the best hyperparameters for that horizon and record\protect\footnotemark.}%
    \label{fig:regular vs irregular breathing}%
    \vspace{-2cm} \ 
\end{figure*}
\addtocounter{footnote}{-1} 
\footnotetext{Same as footnote \ref{footnote: slow motion sequence removed}}
\addtocounter{footnote}{+1} 
\footnotetext{The error values corresponding to an RNN with fixed hidden layer weights were very high compared to the other methods for input sampling frequencies equal to 10Hz. Therefore, they were not plotted in the corresponding graphs to improve readability.}
\end{landscape}
}

\end{document}